\pdfoutput=1

\documentclass[10pt,twocolumn,twoside,journal]{IEEEtran}

\usepackage{graphicx}
\usepackage{amsmath,amssymb}    
\usepackage{cs}
\usepackage{mathsymb}
\usepackage{color}
\usepackage{subfigure}

\newtheorem{dfn}{Definition}   
\newtheorem{thm}{Theorem}      

\usepackage{amsmath}
\usepackage{url}
\usepackage{multirow}
\renewcommand{\Real}{{\mathbb R}}
\renewcommand{\Integer}{{\mathbb N}}
\renewcommand{\F}{{\mathcal F}}

\def\hbx{{\hat{\mathbf x}}}

\def\hbeta{{{\beta}}}   

\def\hw{{\widehat{w}}}

\def\dist{{ {\rm dist} }}
\def\sgn{{ {\rm sign} }}

\def\FRa{{${\rm FR}_{0.20}$}\xspace}
\def\FRb{{${\rm FR}_{0.25}$}\xspace}

\def\half{ {\frac{1}{2}} }

\def\AnnealedMS{{{\sc AnnealedMS}}}

\def\NORMA{{\sc Norma}}

\def \grad{{\nabla}}

\begin{document}

\title{Generalized Kernel-based Visual Tracking} 

\author{ 
         Chunhua~Shen,
         Junae~Kim,
         and
         Hanzi Wang
\thanks
{Manuscript received May 9, 2008; revised January 9, 2009 and April
16, 2009.
First published July X, 200X; current version published August X,
200X.
NICTA is funded by the Australian Government's Department of
Communications, Information Technology, and the Arts and the
Australian Research Council through {\em Backing Australia's Ability}
initiative and the ICT Research Center  of Excellence programs.
This paper was recommended
by Associate Editor D. Schonfeld.
}
\thanks
{
C. Shen is with NICTA, Canberra Research Laboratory, 
Locked Bag 8001, Canberra, ACT 2601, Australia
(e-mail: chunhua.shen@nicta.com.au).
}
\thanks
{
J. Kim is with the Research School of Information Science and Engineering,
Australian National University, Canberra, ACT 0200, Australia
(e-mail: jkim@rsise.anu.edu.au). 
}
\thanks
{
H. Wang is with the School of Computer Science, 
University of Adelaide, Adelaide, SA 5005, Australia
(e-mail: hanzi.wang@ieee.org).
}
\thanks
{Color versions of one or more of the figures
in this paper are available online
at http://ieeexplore.ieee.org.
}
\thanks
{Digital Object Identifier 10.1109/TCSVT.2010.XXXXXX}
\thanks
{
Copyright (c) 2009 IEEE. Personal use of this material is
permitted. However, permission to use this material
for any other purposes must be obtained from the IEEE by
sending an email to pubs-permissions@ieee.org.
}
}

 \markboth{IEEE TRANSACTIONS ON CIRCUITS AND SYSTEMS FOR VIDEO
 TECHNOLOGY,~VOL.~XX, NO.~XX,~MARCH~200X}
 {SHEN
 \MakeLowercase{\textit{et al.}}: GENERALIZED KERNEL-BASED VISUAL TRACKING}

\maketitle

%
%

%
%

\begin{abstract}

      Kernel-based mean shift (MS) trackers have proven to be a promising alternative to
      stochastic particle filtering trackers. Despite its popularity, MS trackers have two
      fundamental drawbacks: (1) The template model can only be built from a single image;
      (2) It is difficult to adaptively update the template model. 
      In this work we generalize the plain MS trackers and attempt to overcome these two
      limitations.  
      
      It is well known that modeling and maintaining a representation of a target object is
      an important component of a successful visual tracker.  
      However, little work has been done on building a robust template model for
      kernel-based MS tracking.  In contrast to building a template from a single frame, we
      train a robust object representation model from a large amount of data.        
      Tracking is viewed as a binary classification problem, and 
      a discriminative classification rule is learned to distinguish between 
      the object and background. 
      We adopt a support vector machine (SVM) for training.  The tracker is then
      implemented by maximizing the classification score. An iterative optimization scheme
      very similar to MS is derived for this purpose.  Compared with the plain MS
      tracker, it is now much easier to incorporate  on-line template adaptation to cope
      with inherent changes during the course of tracking.  To this end, a sophisticated
      on-line support vector machine is used.
       We demonstrate successful localization and tracking on various data sets.
	
%

\end{abstract}


\begin{IEEEkeywords}
        Kernel-based tracking, 
        mean shift,
        particle filter,
        support vector machine,
        global mode seeking.
\end{IEEEkeywords}

%
%

\section{Introduction}
\label{sec:Intro}

      Visual localization/tracking plays a central role for many
      applications like intelligent video surveillance, smart transportation
      monitoring systems \etc.  
      Localization and tracking  algorithms aim to find the most similar region 
      to the target in an image. 
      Recently, kernel-based tracking algorithms
	\cite{DVP2003KernelTrack,Fast07Shen,Qu2008Robust} have
	attracted much attention as an alternative to particle filtering
      trackers
	\cite{P02Color,Shen03Multi,Pan2008PF}. 
      One of the most crucial difficulties in robust tracking is the 
      construction of representation models (likelihood models in Bayesian 
      filtering trackers) that can accommodate illumination variations, 
      deformable appearance changes, partial occlusions, \etc. 
%
      Most current tracking algorithms use a single static template 
      image to construct a target representation based on density models.  
      For both kernel-based trackers and particle filtering trackers, 
      a popular method is to exploit color distributions in simple regions 
      (region-wise density models).  Generally semi-parametric kernel density 
      estimation techniques are adopted. However, it is difficult to update 
      this target model 
      \cite{DVP2003KernelTrack,Fast07Shen,P02Color,Wang2007Adaptive}, 
      and the target representation's fragility usually breaks these trackers 
      over a long image sequence.

	Considerable effort has been expended to ease these difficulties.
	We believe that the key to finding a solution is to find
	the right representation.
	In order to accommodate appearance changes, the representation
	model should be learned from as many training examples as
	possible.		
	Fundamentally two methods, namely on-line and off-line
	learning, can be used for the training procedure. 
	On-line learning means constantly updating the
	representation model during the course of tracking.
	\cite{Lim04Incremental} proposes an incremental eigenvector
	update strategy to adapt the target representation model. A
	linear probabilistic principal component analysis model is
	used. The main disadvantage of the eigen-model is that it is not
	generic and is usually only suitable for characterizing
	texture-rich objects.
	In \cite{Jepson01Robust} a wavelet model is updated using the
	expectation maximization (EM) algorithm. 
	A classification function is progressively learned using
	AdaBoost for visual detection and tracking in
	\cite{Javed05Online} and \cite{Avidan05Ensemble} respectively. 
	\cite{Han05Online} adopts pixel-wise Gaussian mixture models (GMMs)
	to represent the target model and sequentially update them. To
	date, however, less work has been reported on how to
	elegantly update {\em region-wise density} models in tracking.

	In contrast, classification\footnote{Object detection
	is typically a classification problem.}
	is a powerful bottom-up procedure: It is trained off-line and
	works on-line.  Due to the training being typically built on
	very large amounts of training data, its performance
	is fairly promising even without on-line updating of the
	classifier/detector.  
%
	Inspired by image classification tasks with color density
	features 
      and real-time detection,
	%
      %
      we learn off-line a density
	representation model from multiple training data.  By
	considering tracking as a binary classification problem, a
	discriminative classification rule is learned to distinguish
	between the tracked object and background patterns. 
	In this way a robust object representation model is obtained.
	This proposal provides a basis for considering the design of
	enhanced kernel-based trackers using robust kernel object
	representations.  A by-product of the training is the
	classification function, with which the tracking problem is
	cast into a binary classification problem. An object detector
	directly using the classification function is then available.
	Combining a detector into the tracker makes the tracker more
	robust and provides the capabilities of automatic initialization
	and recovery from momentary tracking failures.

	In theory, many classifiers can be used to achieve our goal.
	In this paper we show that the popular kernel based non-linear support
	vector machine (SVM) well fits the kernel-based tracking
	framework. Within this framework the traditional kernel object
	trackers proposed in
	\cite{DVP2003KernelTrack} and \cite{Elgammal2003Joint} can be
	expressed as special cases. Because we use probabilistic
	density features, the learning process is closely related to
	probabilistic kernels based SVMs
      \cite{Jaakkola98Exploiting,Tsuda02New,Moreno03Kullback,Jebara04Probability}.
      It is imperative to minimize computational costs for
	real-time applications such as tracking.
%
%
      A desirable property of the proposed algorithm
	is that the computational complexity is independent of the 
	number of support vectors. 
	Furthermore we empirically demonstrate that our algorithm
	requires fewer iterations to achieve convergence.

	Our approach differs from \cite{Avidan04Support}
	although both use the SVM classification score as the cost
	function.  In \cite{Avidan04Support}, Avidan builds a tracker
	along the line of standard optical flow tracking. 
	Only the homogeneous quadratic polynomial kernel (or kernels
	with a similar quadratic structure) can be used in order to
	derive a closed-form solution.
      This restriction prevents one using a more appropriate 
      kernel obtained by model selection.  
%
%
%
%
%
      An advantage of \cite{Avidan04Support} is that it can be used
      consistently with the optical flow tracking, albeit only gray pixel
      information can be used. 
	Moreover, the optimization procedure of our approach is
	inspired by the kernel-based object tracking paradigm
	\cite{DVP2003KernelTrack}. 
	Hence extended work such as \cite{Fast07Shen} is also
	applicable here, which enables us to find the global optimum.   
	If joint spatial-feature density is used to train an SVM,
	a fixed-point optimization method may also be derived that is similar to
	\cite{Elgammal2003Joint}.
	The classification function of the SVM trained for vehicle
	recognition is not smooth \wrt spatial
	mis-registration (see Fig.~1 in \cite{Williams05Sparse}). 
      %
      %
      We employ a spatial kernel to smooth the cost function when computing 
      the histogram feature.  In this way, gradient based optimization methods 
      can be used. 
	Using statistical learning theory, we
	devise an object tracker that is consistent with MS
	tracking.  The MS tracker is initially derived from
	kernel density estimation (KDE). Our work sheds some light on
	the connection between SVM and KDE\footnote{It is
	believed that statistical learning theory (SVM and many other
	kernel learning methods) can be interpreted in the framework
	of information theoretic learning
	\cite{Jenssen04Towards,Jenssen04Laplacian}.}.

      Another important part of our tracker is its on-line re-training in 
      parallel with tracking. 
      Continuous updating of the representation model can capture changes of 
      the target appearance/backgrounds.  
      Previous work such as 
      \cite{Jepson01Robust,Avidan05Ensemble,Lim04Incremental,Han05Online} 
      has demonstrated the importance of this 
      on-line update during the course of tracking. The incremental SVM 
      technique meets this end 
      \cite{Cauwenberghs00Incremental,Fung02Incremental,Kivinen04Online,Bordes05Fast},
      which efficiently 
      updates a trained SVM function whenever a sample is added to or removed 
      from the training set.  
      For our proposed tracking framework, the target model can be learned in
      either batch SVM training or on-line SVM learning. 
      We adopt a sophisticated on-line SVM learning 
      proposed in \cite{Kivinen04Online} for its efficiency and simplicity. 
      We address the crucial problem of adaptation, \ie, the on-line
      learning of discriminant appearance model while avoiding drift.

	The main contributions of our work are to solve MS trackers' two
      drawbacks: The template model can only be built from a single image; 
      and it is difficult to update the model.
      The solution is to
	extend the use of statistical learning algorithms for
	object localization and tracking. SVM has been used for
	tracking by means of spatial perturbation of the SVM
	\cite{Avidan04Support}. We exploit SVM for 
	tracking in a novel way (along the line of MS tracking). 
	The key ingredients of our approach are: 
	\begin{itemize}

	\item
            
            Probabilistic kernel based SVMs are trained and incorporated into 
            the framework of MS tracking.  By 
            carefully selecting the kernel, we show that no extra computation 
            is required compared with the conventional single-view MS 
            tracking.
       \item
                         
             An on-line SVM can be used to adaptively update the target model.
             We demonstrate the benefit of on-line target model update. 

      \item
            We show that the annealed MS algorithm proposed in  
            \cite{Fast07Shen} can be viewed as a special case of the 
            continuation method under an appropriate interpretation.  With the 
            new interpretation, annealed MS can be extended to more 
            general cases.  
            Extension and new discovers are discussed.       
	      An efficient localizer is built with global mode seeking
	      techniques. 

         \item

            Again, by exploiting the SVM binary classifier, it is able to 
            determine the scale of the target.  An improved annealed MS-like
            algorithm with a cascade architecture is developed.  It 
            enables a more systematic and easier design of the annealing 
            schedule, in contrast with {\em ad hoc} methods in previous work 
            \cite{Fast07Shen}.

      \end{itemize}

%

	 The remainder of the paper is organized as follows. 
       In \S\ref{sec:PRE}, the general theory of MS tracking and SVM is reviewed 
       for completeness.  
       Our proposed tracker is presented in \S\ref{sec:GKT}. Finally 
       experimental results are reported in \S\ref{sec:exp}. We conclude this 
       work in \S\ref{sec:Conclusion}.



%
%

\section{Preliminaries}
\label{sec:PRE}

    For self-completeness, we review mean shift tracking, support
    vector machine and its on-line learning version in this section.

\subsection{Mean Shift Tracking} 
\label{sec:MST}
    
    Mean shift (MS) tracking
    was firstly presented in \cite{DVP2003KernelTrack}.  
	In MS tracking, the object is represented by a square
    region which is cropped and normalized into a unit circle.
    By denoting $ \bq $
	as the color histogram of the target model, and $ \bp( \bc ) $ 
	as the target candidate color histogram with the center at 
	$ \bc $, the similarity function  between $ \bq $ and 
	$ \bp( \bc )  $
    is (when Bhattacharyya divergence
	\cite{DVP2003KernelTrack} is used),
    $$ 
               \dist ( \bq, \bp( \bc ) ) =  
               \sqrt{1 - \varrho( \bq, \bp) }.
    $$
    Here
    $ 
    \varrho( \bq, \bp) = 
    \sqrt { \bq } ^\T \sqrt{ \bp}
    $ is the dissimilarity measurement.
    Let $ \{ \bI_\ell \}_{\ell = 1 }^{n} $ be a region's pixel positions
	in image $ \bI $ with the center at 
	$ \bc $.
    In order to make the cost function smooth---otherwise gradient
    based MS optimization cannot be applied---a kernel with
    profile $ k( \cdot ) $ is employed to assign smaller weights to
    those pixels farther from the center, considering the fact that
    the peripheral pixels are less reliable. 
    An $ m $-bin color histogram is built for 
	an image patch located at $ \bc $, 
	$ \bq ( \bc ) = \{  q_u ( \bc ) \}_{ u = 1} ^ m $, where
\begin{equation}  
   	\label{EQ:hist1}	
   	q_u = \lambda \sum_{\ell = 1}^n 
			 k\Bigl( \Bigl\Vert
		\frac{
		\bc - \bI_\ell 
		}{h}
		\Bigr\Vert ^ 2  \Bigr)
		\delta (  \vartheta ( \bI_\ell ) - u ).
\end{equation}
	Here
	$ k( \cdot ) $ is the homogeneous spatial weighting kernel
	profile and 
	$ h $ is its bandwidth. 
    %
    %
	$ \delta( \cdot ) $ is the delta function and
	$ \lambda $ normalizes $ \bq $. The function
	$ \vartheta( \bI_\ell ) $ maps a feature of $ \bI_\ell $
	into a histogram bin $ u $.
	$ \bc $ is the kernel center; and for the target model
	usually $ \bc = 0 $. The representation of 
	candidate $ \bp $ takes the same form.

	Given an initial position $ \bc_0 $, the problem of
	localization/tracking is to estimate a best displacement 
	$\Delta\bc $ 
	such that the measurement $ \bp ( \bc_0 + \Delta\bc ) $
	at the new location best matches the target $ \bq $, \ie,
%
%
$$
   \Delta\bc^\star = \argmin\nolimits_{ \Delta\bc }
   { \dist ( \bq, \bp ( \bc_0
   + \Delta\bc )    )   }.
$$

	By Taylor expanding $ \dist( \bq, \bp (\bc) ) $ at 
	the start position $ \bc_0 $ and keeping only the linear
	item (first-order Taylor approximation), the above optimization
    problem can be resolved by an iterative procedure:
   
    \begin{equation}
    \label{EQ:MSIteration0}	
	\bc ^ { [ \tau + 1 ] } 
		=
        \frac{  \sum_{\ell=1}^n {  \bI_\ell {\widetilde w}_\ell  g
        (
		\Vert  \frac{\bc ^ { [ \tau ] }
			  - \bI_\ell}{h}
		\Vert^2          
        )
		}   
		}	
		{ 
        \sum_{\ell=1}^n {  
        {\widetilde w}_\ell  g  
        (
		\Vert  \frac{\bc ^ { [ \tau ] } 
		          - \bI_\ell}{h}
		\Vert^2 
		)	
		}   
		}, 
     \end{equation}
     where 
     $ g( \cdot ) = - k'( \cdot )$ and the superscript 
     $ \tau = 0,1,2\dots $, indexes the iteration step.
     The weights $ {\widetilde w}_\ell $ are calculated as: 
     $
      {\widetilde w}_\ell
      = \sum_{u=1}^{m}
      {
      \sqrt{\frac{ q_u }{ p_u( \bc_0 )  } }
      \delta (  \vartheta ( \bI_\ell ) - u ).
      }
     $
     See \cite{DVP2003KernelTrack} for details.


\subsection{Support Vector Machines} 
\label{sec:SVM}

     We limit our explanation of the support
     vector machine classifiers algorithm to an overview.
    
    Large margin classifiers have demonstrated their
    advantages in many vision tasks.
	SVM is one of the popular large margin
	classifiers \cite{Vapnik95Nature} which
	has a very promising generalization capacity.
    
     The linear SVM is the best understood and simplest to apply.
     However, linear separability is a rather strict condition.
     Kernels are combined into margins for relaxing this
     restriction. 
	SVM is extended to deal with linearly non-separable
	problems by mapping the training data from the input space
	into a high-dimensional, possibly infinite-dimensional,
	feature space, \ie, $ \Phi(\cdot):  \cX \rightarrow \F$.
	Using the kernel trick, the map $ \Phi( \cdot ) $ is not
	necessarily known explicitly.
     Like  other kernel methods, SVM constructs a
     symmetric and positive definite kernel matrix (Gram matrix)
     which represents the similarities between all training datum
     points. 
    Given $ N $ training data 
	$ \{ ( \bx_i, y_i ) \}_{i=1}^N $,
	the kernel matrix is written as:
	$
	     K_{ij} \equiv K( \bx_i, \bx_j )
		    = \left< \Phi (\bx_i), \Phi (\bx_j) \right>,
		    i,j=1\cdots N
	$. 
	When $ K_{ij} $ is large, the labels of
	$ \bx_i $ and $ \bx_j $, $ y_i $ and $ y_j $,
	are expected to be the same. Here, 
	$ y_i, y_j \in \{+1, -1 \} $.
	The decision rule is
	given by $ \sgn \left( f( \bx ) \right) $ with
	\begin{equation}
	\label{EQ:svm1}	
		f( \bx ) = \sum_{i=1}^{N_S} 
		\hbeta_i K( \hbx_i, \bx ) + b	
	\end{equation}
	where $ \hbx_i \in \cX $, $ i = 1 \cdots N_S  $, are support
	vectors, $ N_S $ is the number of support vectors,  
	$ \hbeta_i $ is the weight associated with $ \hbx_i $, and $ b
	$ is the bias. 
    
    The training process of SVM then determines
	the parameters $ \{ \hbx_i, \hbeta_i, b, N_S \} $ by solving
	the optimization problem
	\begin{align} 
	\label{EQ:svm2}
		\minimize_{ {\boldsymbol \xi},\bw,b}   \quad 
		&  
		\half  \| \bw \|_r^r + C \sum\nolimits_{i=1}^N
		\xi_i  ,  \\
	      \st \quad                     &  
		y_i (  \bw ^\T \Phi (  \bx_i )  + b ) \geq 1 - \xi_i,
			\quad \forall i,                 \notag   \\
			& \xi_i \geq 0, \quad \forall i, \notag 	
      \end{align}
      where $ {\boldsymbol \xi} = \{ \xi_i \}_{i=1}^N  $ is the slack variable 
      set and the regularization parameter $ C $ determines the trade-off 
      between SVM's generalization capability and training error.  
      $ r =1 , 2 $ corresponds to $ 1 $-norm and $2$-norm SVM respectively. 
      The solution takes the form 
      $ \bw = \sum_{i=1}^N y_i \alpha_i \Phi ( \bx_i )  $.  Here, 
      $ \alpha_i \geq 0 $ and most of them are $ 0 $, 
      yielding sparseness.  
      The optimization \eqref{EQ:svm2} can be efficiently solved by linear 
      programming ($1$-norm SVM) or quadratic programming ($2$-norm SVM) in
      its dual. Refer to \cite{Vapnik95Nature} for details.


\subsection{On-line Learning with Kernels}
\label{sec:onlineSVM}

   A simple on-line kernel-based algorithm, termed \NORMA, has been
   proposed for a variety of standard machine learning tasks in
   \cite{Kivinen04Online}.  The algorithm is computationally cheap at
   each update step. We have implemented {\NORMA} here for on-line SVM
   learning. See Fig.~$1$ in \cite{Kivinen04Online} for the backbone
   of the algorithm. We omit the details due to space constraint.
   
   As mentioned, visual tracking is naturally a time-varying problem.
   An on-line learning method allows updating the model during the
   course of tracking.


\section{Generalized Kernel-based Tracking}
\label{sec:GKT}

   The standard kernel-based MS tracker is generalized by maximizing a
   sophisticated cost function defined by SVM.

\subsection{Probability Product Kernels}

	Measuring the similarity between images and image patches is
	of central importance in computer vision. In SVMs, the kernel
	$ K( \cdot, \cdot ) $ plays this role. Most commonly used
	kernels such as Gaussian and polynomial kernels are not
	defined on the space of probability distributions. Recently
	various probabilistic kernels have been introduced, including 
	the Fisher kernel \cite{Jaakkola98Exploiting}, TOP \cite{Tsuda02New},
	Kullback-Leibler kernel \cite{Moreno03Kullback} and
	probability product kernels (PPK) \cite{Jebara04Probability},
	to combine generative models into discriminative classifiers. 
	A probabilistic kernel is defined by first fitting a
	probabilistic model $ \bp ( \bx_i ) $ to each training vector 
	$ \bx_i $. The kernel is then a measure of similarity between
	probability distributions.
      PPK is an example \cite{Jebara04Probability}, 
      with kernel given by
      \begin{equation}
      \label{eq:PPK1}
		K_{\rho}^\star ( \bq( \bx ) , \bp( \bx ) ) 
		 = \int_\cX  \bq( \bx ) ^ \rho \bp( \bx ) ^ \rho 
		 \, d\bx
 	\end{equation}
	where $ \rho $ is a constant. When $ \rho = \frac{1}{2} $,
	PPK reduces to a special case, termed the Bhattacharyya
	kernel:
	\begin{equation}
		K_\half^\star ( \bq( \bx ) , \bp( \bx ) ) 
		 = \int_\cX \sqrt{ \bq( \bx ) } \sqrt{ \bp( \bx ) }
		 \, d\bx.
		 \label{EQ:BhatKernel1}	 
 	\end{equation}
	In the case of discrete histograms, 
	\ie, $ \bq ( \bx ) = [ q_1\cdots q_m ]^\T $
	and  $ \bp ( \bx ) = [ p_1\cdots p_m ]^\T $,
	\eqref{EQ:BhatKernel1} becomes
	\begin{equation}
		K_\half^\star ( \bq( \bx ) , \bp( \bx ) ) 
		= \sqrt { \bq ( \bx ) } ^\T \sqrt{ \bp( \bx )}
		 = \sum_{u=1}^{m} \sqrt{ q_u p_u }.
	\label{EQ:BhatKernel2}	 
	\end{equation}
	When $ \rho =1 $, 
	$  K_{1}^\star( \cdot, \cdot ) $ computes the expectation of
	one distribution over the other, and hence is termed the expected
	likelihood kernel \cite{Jebara04Probability}.
	In \cite{Yang05Efficient} its corresponding statistical
	affinity is used as similarity measurement for tracking.
      The Bhattacharyya kernel is adopted in this work
      due to:
      \begin{itemize}
          \item
               The standard MS tracker \cite{DVP2003KernelTrack} 
               uses the Bhattacharyya 
               distance.
               It is clearer to show the connection between the proposed
               tracker and the standard  MS  tracker by using
               Bhattacharyya kernel. 
          \item
               It has been empirically shown, at least for image
               classification, that the generalization capability of
               expected likelihood kernel $ K_1^\star (\cdot, \cdot) $ is 
               weaker than the Bhattacharyya kernel.
               Meanwhile, non-linear probabilistic kernels including 
               Bhattacharyya kernel, Kullback-Leibler kernel, R\'{e}nyi kernel  
               \etc perform similarly \cite{Chan04Family}.
               Moreover, Bhattacharyya kernel is simple and has no kernel
               parameter to tune.                 
      \end{itemize}

     The PPK has an interesting characteristic that the mapping function
     $\Phi( \cdot ) $ is explicitly known: 
     $
          \Phi( \bq ( \bx ) ) = \bq(\bx) ^\rho
     $.
     This is equivalent to directly setting 
     $   \bx = \bq(\bx)^\rho  $ and the kernel 
     $ K^\star_\rho ( \bx_i, \bx_j ) = \bx_i ^\top \bx_j  $.
     Consequently for discrete PPK based SVMs, in the test
	phase the computational complexity is independent of the
	number of support vectors.
	This is easily verified. The
	decision function is 
	\begin{align*}
	f( \bx ) 
		&= \sum_{i=1}^{N_S}{ 
			\hbeta_i \left[  \bq( \bx_i )^\rho \right]^\T 
			\bp( \bx )^\rho + b  
	 }  \\ 
	    &= 
		\left[  \sum_{i=1}^{N_S}{ 
		\hbeta_i   \bq( \bx_i )^\rho } 
		\right]^\T 
		\bp( \bx )^\rho + b.  
	\end{align*} 
	The first term in the bracket can be calculated beforehand. 
	For example, for histogram based image classification like
	\cite{Chapelle99SVM}, given a test image $ \bx $,
	the histogram vector $ \bp( \bx ) $ is immediately available.
	In fact we can interpret discrete PPK based SVMs as {\em linear}
	SVMs in which the input vectors are $ \bq( \bx_i )^\rho
	$---the features {\em non-linearly}\footnote{
      When $\rho=1$, it is linear.          
      The non-linear probabilistic kernels induce a transformed feature space 
      (as the Bhattacharyya kernel does) to smooth density such that they 
      significantly improve classification over the linear kernel 
      \cite{Chan04Family}.} extracted from image densities.      
      Again, one might argue that, since the Bhattacharyya kernel is very 
      similar to the linear SVM, it might not have the same power in 
      modelling complex classification boundaries as the traditional 
      non-linear kernels like the Gaussian or polynomial kernel.  The 
      experiments in \cite{Chan04Family} indicate that the classification 
      performance of a probabilistic kernel which consists an exponential 
      calculation is not clearly better: exponential kernels like the 
      Kullback-Leibler 
      kernel and R\'enyi kernel performs similarly as Bhattacharyya kernel on 
      various datasets for image classification.
      Moreover our main purpose is to learn a representation model for
      visual tracking. Unlike other image classification tasks---in
      which high generalization accuracy is demanded---for visual
      tracking achieving very high accuracy might not be necessary 
      and may not translate to a significant increase in tracking
      performance.

      Note that PPKs are less compelling when the input data are vectors 
      with no further structure. However, even the Gaussian kernel is a 
      special case of PPK ($\rho = 1 $ in Equation~\eqref{eq:PPK1} and 
      $\bp(\bx)$ is a single Gaussian fit to $\bx_i$ by maximum likelihood)
      \cite{Jebara04Probability}.   

    By contrast,  the reduced set method is applied in
    \cite{Avidan04Support} to reduce the number of support vectors for
    speeding up the classification phase.
    Applications which favour fast computation in the testing phase,
	such as large scale image retrieval, might also benefit from
	this discrete PPK's property.

\subsection{Decision Score Maximization}
\label{sec:DSM}

	It is well known that the magnitude of the SVM score 
	$ | f( \bx ) | $ measures the {\em confidence} in the
	prediction. 
	The proposed tracking is based on the assumption that the
	local maximum of the SVM score corresponds to the target
	location we seek, starting from an initial guess close to the
	target.
	
	If the local maximum is positive, the tracker accepts the
	candidate.  Otherwise an exhaustive search or localization
	process will start. The tracked position at time $ t $ is the
	initial guess of the next frame $ t+1 $ and so forth. 
	We now show how the local maximum of the decision score is
	determined.

    As in \cite{DVP2003KernelTrack}, a histogram
    representation of the image region can be computed as
    Equation~\eqref{EQ:hist1}.


	With Equations~\eqref{EQ:svm1}, \eqref{EQ:BhatKernel2} and 
	\eqref{EQ:hist1}, we have\footnote{ 
	We have used $  \bx $ to represent the image region.
	We also use the image center $ \bc $ to represent the 
	image region $\bx$.
	For clarity we define notation
	$q_{i,u} \equiv q_u( \hbx_i ) $.	
	}
	\begin{equation} 
	\label{EQ:svm3}	
		f( \bc ) = \sum_{i=1}^{N_S}{
		\hbeta_i 
		\sum_{u=1}^m{\sqrt{ q_{i,u}  p_u( \bc ) }}
	   } +b.
   	\end{equation}  
	We assume the search for the new target location starts from 
	a near position $\bc_0 $, then a 
	Taylor expansion of the kernel around 
	$ p_u ( \bc_0 ) $ is applied, similar to \cite{DVP2003KernelTrack}. 
	After some manipulations and putting those terms independent of 
	$ \bc $ together, denoted by $ \Delta $,
	\eqref{EQ:svm3} becomes 
	\begin{align}  
		f( \bc ) &=
          \frac{1}{2}
          \sum_{i=1}^{N_S} {
          \hbeta_i \sum_{ u = 1}^{ m }{ p_u ( \bc ) \sqrt
          { \frac{q_{i,u}}{p_u(\bc_0)} }
          }
          } + \Delta        \notag
         \\   &=
		\frac{\lambda}{2} 
		\sum_{i=1}^{N_S} { 
		\hbeta_i \sum_{\ell = 1}^{n}{w_{i,\ell} 
		 k\Bigl( \Bigl\Vert
		\frac{
		\bc - \bI_\ell 
		}{h}
		\Bigr\Vert ^ 2  \Bigr)
		}
		} + \Delta			\notag  
		\\
			&=
		\frac{\lambda}{2}
		\sum_{\ell = 1}^{n}{ \hw_\ell }
 		k\Bigl( \Bigl\Vert
		\frac{
		\bc - \bI_\ell 
		}{h}
		\Bigr\Vert ^ 2  \Bigr)
		+ \Delta		\label{EQ:svm5}
	\end{align} 	
	where 
	\begin{equation}  
		w_{i,\ell} = \sum_{ u = 1}^m { \sqrt 
		{
		\frac{ q_{i,u} }
		{  p_u ( \bc_0 )} 
		}
		\delta ( \vartheta( \bI_\ell )- u ) }
	\label{EQ:svmWeight1}	
	\end{equation}   
	and
	\begin{equation}
	\hw_\ell  
		= \sum_{i=1}^{N_S}{ \hbeta_i w_{i,\ell}}  
		= \sum_{u=1}^m{ \frac
		{ \left[ \sum_{i=1}^{N_S} { \hbeta_i \sqrt{q_{i,u}}}
		\right] }
		{ \sqrt{p_u( \bc_0 )}  } \delta 
		( { \vartheta(
		\bI_\ell -u )}   )}.
					\label{EQ:SVTWeight2}	
	\end{equation}
    Here \eqref{EQ:svm5} is obtained by swapping the order of
    summation.  The first term of $ f( \bc ) $ is the weighted kernel
    density estimate with kernel profile $ k( \cdot ) $ at $ \bc $.
    It is clear now that our cost function $ f ( \bc ) $ has an
    identical format as the standard MS tracker.
    
    Can we simply set
    $ \nabla_\bc f( \bc ) = 0 $
	which leads to a fixed-point iteration procedure 
    to {\em maximize} $ f( \bc ) $ as the standard MS does?
    If it works, the  optimization would be similar to 
    \eqref{EQ:MSIteration0}.
    
     
     Unfortunately, $ \nabla_\bc f( \bc ) = 0 $ cannot guarantee 
     a local maximum convergence. That means, the fixed point iteration 
     \eqref{EQ:MSIteration0} can converge to a local minimum. 
     We know that only when all the weights $ \hw_\ell $ are positive, 
     \eqref{EQ:MSIteration0} converges to a local maximum---as the
     standard MS does. See Appendix for the theoretical analysis.
     
     However, in our case,
     a negative support vector's weight $ \hbeta_i $ is negative,
     which means some of the weights computed by
     \eqref{EQ:SVTWeight2} could be negative. The traditional
     MS algorithm requires that the sample weights must be
     non-negative. 	
     \cite{Collins03Blob} has discussed the issue
     on MS with negative weights and a {\em heuristic} modification is 
     given to make MS able to deal with samples with negative weights. 
     According to \cite{Collins03Blob},  the modified MS is
     \begin{equation}
        \label{EQ:MSIteration}	
        \bc ^ { [ \tau + 1 ] } 
        =
        \frac{  \sum_{\ell=1}^n 
        {  
        \bI_\ell \hw_\ell  g
        (
        \Vert  \frac{\bc ^ { [ \tau ] }
			  - \bI_\ell}{h}
		\Vert^2                                  
                       )
		}   
		}	
            {  
            \sum_{\ell=1}^n 
            { 
		\vert  
            \hw_\ell 
            g   
        (
		\Vert  \frac{\bc ^ { [ \tau ] } 
		          - \bI_\ell}{h}
		\Vert^2 
		)   
		\vert	
		}   
		}.
	\end{equation}
      Here $ \vert \cdot \vert $ is the absolute value operation. 
      Alas this heuristic solution is problematic. 
      Note that no theoretical analysis is given in 
      \cite{Collins03Blob}.
      We  show that the methods in 
      \cite{Collins03Blob} cannot guarantee converging 
      to a local maximum mode.  
      See Appendix for details.

      The above problem may be avoided by using $1$-class
      SVMs \cite{Scholkopf01Estimating}
      in which $ \hw_\ell $ is strictly positive.  
      However the discriminative power of SVM is also
      eliminated due to its unsupervised nature.

      In this work, we use a Quasi-Newton gradient descent
      algorithm for maximizing $ f(\bc) $ in \eqref{EQ:svm5}. 
      In particular, the L-BFGS algorithm \cite{Liu89LBFGS} is adopted
      for implementing the Quasi-Newton algorithm.
      We provide callbacks for calculating the value of the SVM
      classification function $ f(\bc) $ and its gradient.  Typically,
      only few iterations of the optimization procedure
      are performed at each frame.
      It has been shown that Quasi-Newton can be a better alternative
      to MS optimization for visual tracking \cite{Liu2004Real} in
      terms of accuracy. 
      Quasi-Newton was also used in \cite{Guskov06Kernel} for
      kernel-based template alignment.   
      Besides, in \cite{Fast07Shen} the authors have
      shown that Quasi-Newton converges around twice faster than the
      standard MS does for data  clustering.

      The essence behind the proposed SVM score maximization strategy is
      intuitive. The cost function \eqref{EQ:svm3} favors both the
      dissimilarity to negative training data (\eg, background) and the
      similarity to positive training data. Compared to the standard
      MS tracking, our strategy provides the capability to
      utilize a large amount of training data. The terms with positive $
      \hbeta $ in the cost function play the role to attract the target
      candidate while the negative terms repel the candidate.  In 
      \cite{Zhao04Tracking,Zhao08Segmentation} Zhao \etal 
      have extended MS tracking
      by introducing a background term to the cost
      function, \ie, $  f( \bc ) = \lambda_f K^\star_{\half}
      (\bq, \bp( \bc ) ) - \lambda_b K^\star_{\half} 
      (\bb( \bc ), \bp( \bc )  )$. $ \bb( \cdot ) $ is the
      background color histogram in the corresponding region.
      It also linearly combines both positive and negative terms
      into tracking and better performance has been observed.
      It is simple and no training procedure is needed. Nevertheless
      it lacks an elegant means to exploit available training
      data and the weighting parameters $ \lambda_f $ and $
      \lambda_b $ need to be tuned manually\footnote{Zhao \etal
      \cite{Zhao04Tracking,Zhao08Segmentation} 
      did not correctly treat MS iteration with
      negative weights either because they 
      have used Collins' modified MS
      (Equation~\eqref{EQ:MSIteration}).  
      }.

      The original MS tracker's analysis relies on kernel properties
      \cite{DVP2003KernelTrack}. We argue that the main purpose of the kernel 
      weighting scheme is to smooth the cost function such that iterative
      methods are applicable. Kernel properties then derive an
      efficient MS optimization. As observed by many other authors 
      \cite{Liu2004Real,Dewan2006Toward}, the kernels used as weighting
      kernel density estimation \cite{Wand1995Kernel,Comaniciu2002Mean}.     
      We can simply treat the feature distribution as a weighted histogram to 
      smooth the cost function and, at the same time, to account for the 
      non-rigidity of tracked targets.

	Note that (1) the optimization reduces to the standard MS
	tracking if $ N_S = 1 $; (2) Other probability kernels like 
	$K_1^\star( \cdot, \cdot )$ are also applicable here. The only
	difference is that $ w_{ i,\ell} $ in \eqref{EQ:svmWeight1}
	will be in other forms.
		
	
	In previous contents we have shown that in the testing phase
	discrete PPK's support vectors do not introduce extra computation. Again,
	for our tracking strategy, no computation overhead is
	introduced compared with the traditional MS tracking in
	\cite{DVP2003KernelTrack}. This can be seen from Equation
	\eqref{EQ:SVTWeight2}. The summation in \eqref{EQ:SVTWeight2}
	(the bracketed term) can be computed off-line. 
	The only extra computation resides in the training
	phase: the proposed tracking algorithm has the {\em same}
	computation complexity as the standard MS tracker.
	It is also straightforward to extend this tracking framework
	to spatial-feature space \cite{Elgammal2003Joint} which has
	proved more robust.

%
%
%
%
%
%
%
%
%
\begin{figure*}[t!]
   \centering
   \includegraphics[width=.271\textwidth]{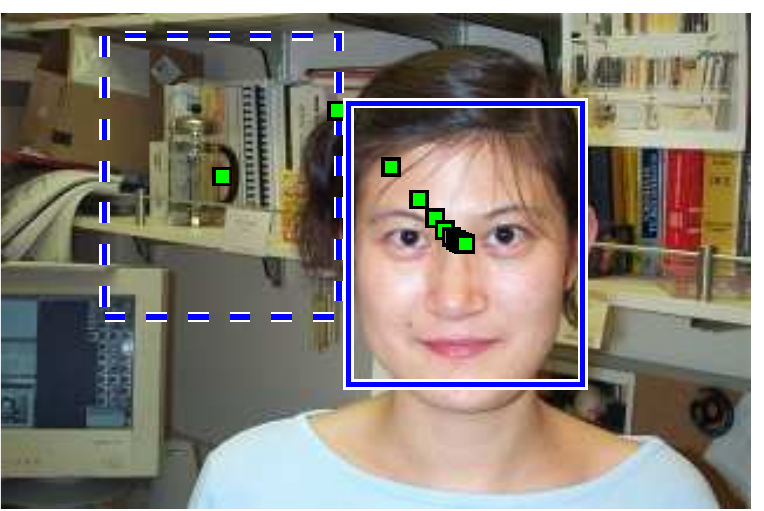}
   \includegraphics[width=.271\textwidth]{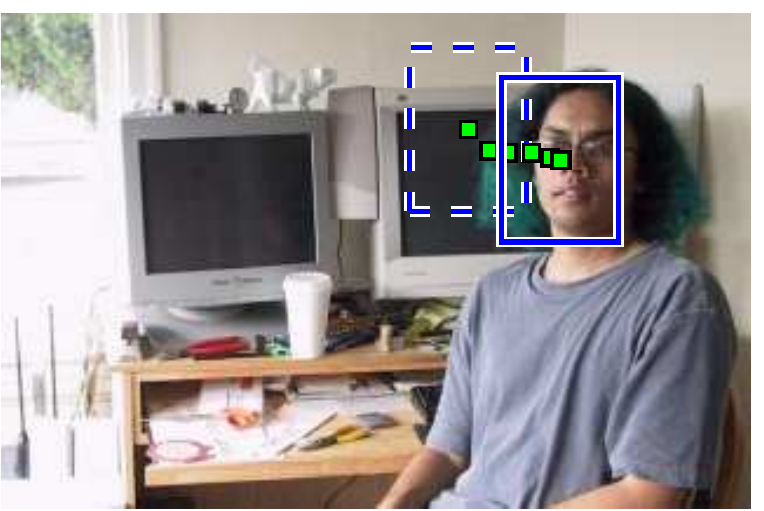}
   \includegraphics[width=.271\textwidth]{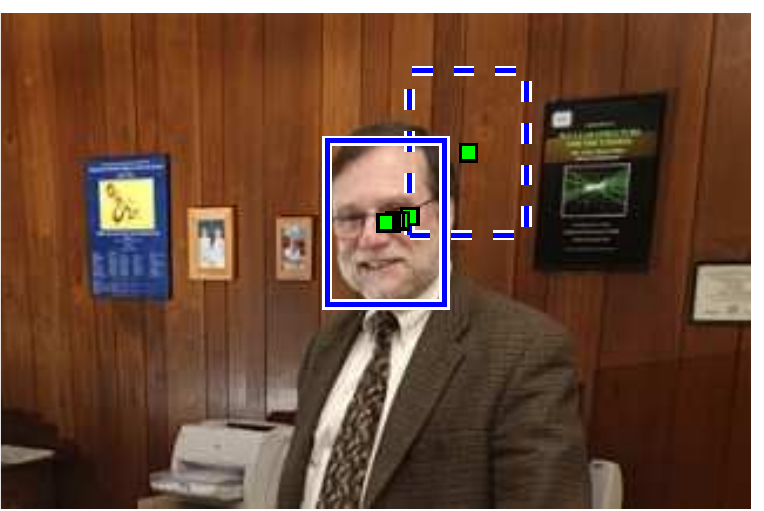} \\
   \includegraphics[width=.271\textwidth]{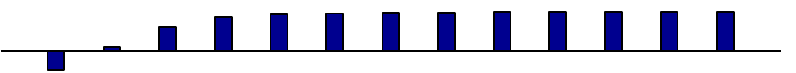}
   \includegraphics[width=.271\textwidth]{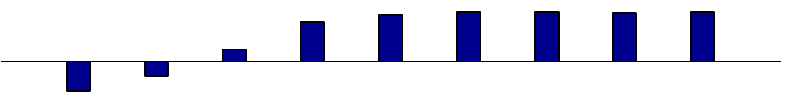}
   \includegraphics[width=.271\textwidth]{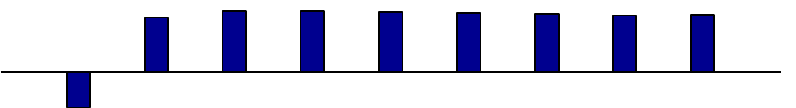}
   \caption{Examples 
   of the initial position (dashed line)
   and the final convergence position (solid line).
   Squared dots show the optimization convergence trajectory.
   The image size in all tests is $ 320 \times 211 $.
   The object size is $ 60 \times 50 $ for the first
   example and $ 35 \times 25 $ for the other two.  
   The bars under every test image indicate the SVM
   score at each gradient-ascent iteration. 
   The SVM score change is: ({left}) initial: $ -1.41 $,
   final: $ 2.77 $;  ({middle}) initial: $ -0.86 $,
   final: $ 1.41 $; ({right}) initial: $ -1.04 $, final:
   $ 1.62 $.
   }
   \label{fig:TrackingExp1}
\end{figure*}
%
%


\subsection{Global Optimum Seeking}

      A technique is proposed in \cite{Fast07Shen}, 
      dubbed annealed mean shift 
      ({\AnnealedMS}), to reliably find the global {\em density} mode.  
      {\AnnealedMS} is motivated by the observation that the number of modes 
      of a kernel density estimator with a Gaussian kernel is monotonically 
      non-increasing \wrt the bandwidth of the kernel.

      Here we re-interpret this global optimization and show that it
      is essentially a special case of the {\em continuation} approach
      \cite{Wu1996Effective}. With the new interpretation, it is clear
      now that this technique is applicable to a broader types of cost
      functions, not necessary to a density function.   

      The continuation method is one of the unconstrained global
      optimization techniques which shares similarities with
      deterministic annealing.   A series of gradually deformed but
      smoothed cost functions are successively optimized, where the
      solution obtained in the previous step serves as an initial
      point in the current step. This way the convergence information
      is conveyed. With sufficient smoothing, the first cost function
      will be concave/convex such that the global optimum can be
      found.  The algorithm iterates until it traces the solution back
      to the original cost function.             
      We now recall some basic concepts of the continuation method.        
      \begin{dfn}[\cite{Wu1996Effective}] 
            Given a non-linear function $f$, the transformation
            $ \left< f \right>_h $ for $f$ is defined such that 
            $\forall \, \bx$,
            \begin{equation}
            \label{EQ:deformation}
                  \left< f \right>_h (\bx) = C_h 
                  \int{ f(\bx') 
                  k \Bigl( \Bigl\Vert \dfrac{\bx - \bx'}{h} \Bigr\Vert^2 \Bigr) }
                  d\bx', 
            \end{equation}
            where $k(\cdot)$ is a smoothing function; usually the Gaussian is used.  
            $ h $ is a positive scalar which controls the degree of smoothing.  
            $C_h$ is a normalization constant such that 
            $ C_h {\displaystyle \int{ 
                  k \Bigl( \Bigl\Vert \dfrac{\bx }{h} \Bigr\Vert^2 \Bigr) 
                  }}d\bx =1. $
      \end{dfn}
      Note the similarity between the smoothing function $k(\cdot)$
      and the definition of the kernel in KDE.  From
      \eqref{EQ:deformation}, the defined transformation is actually
      the convolution of the cost function with $ k(\cdot) $. In the
      frequency domain, the frequency response of $\left< f \right>_h
      $ equals the product of the frequency responses of $ f $ and $k
      $.     
      Being a smoothing filter, the effect of $ k(\cdot) $ is to
      remove high frequency components of the original function.
      Therefore one of the requirements for $k(\cdot)$ is its
      frequency response must be a low-pass frequency filter. 
      We know that popular kernels like Gaussian or Epanechnikov
      kernel are low-pass frequency filters.   This is one of the
      principle justifications for using Gaussian or Epanechnikov to
      smooth a function.
      When $ h $ is increased,  $ \left< f \right>_h $ becomes
      smoother and for $ h = 0 $, the function is the original
      function.

      \begin{thm}
         The annealed version of mean shift introduced in \cite{Fast07Shen}
         for global mode seeking is a special case of the general
         continuation method defined in Equation \eqref{EQ:deformation}.
      \end{thm}
      \begin{proof}
            Let the original function $ f(\bx') $ take the form of a
            Dirac delta comb (\aka impulse train in signal processing), 
            \ie, $ f(\bx' ) = \sum_i{ \delta( \bx' - \hbx_i )  }$,
            where  $ \hbx_i $ is known.
            With the fundamental property that 
            $
            {
               \displaystyle 
               \int
               F( \bx) \delta( \bx - \hbx ) 
            }
               d\bx = F( \hbx) 
            $ for any function $ F(\cdot)$, we have
            $\left< f \right>_h (\bx) = C_h
            {\displaystyle
            \sum_i{
                  k \Bigl( \Bigl\Vert \dfrac{\bx - \hbx_i}{h}
                  \Bigr\Vert^2 \Bigr).
               }
            }
            $
            This is exactly same as a KDE. This discovers that
            {\AnnealedMS}  is a special case of the continuation
            method.
            When $ f(\bx') =   \sum_i { w_i \delta( \bx' - \hbx_i ) } $
            with $ w_i \in \Real $ ($ w_i $ can be negative), the
            above analysis still holds and this case corresponds to
            the SVM score maximization in \S\ref{sec:DSM}.
      \end{proof}
      It is not a trivial problem to determine the optimal scale of
      the spatial kernel bandwidth, \ie, the size of the target, for
      kernel-based tracking.  A line search method is introduced in
      \cite{Collins03Blob}.  For 
      \AnnealedMS, an important
      open issue is how to design the annealing schedule. Armed with
      an SVM classifier, it is  possible to determine the
      object's scale.  
      If only the color feature is used, due to its
      lack of spatial information and insensitive to scale change, 
      it is difficult to estimate a fine scale
      of the target. By combining other features, better
      estimates are expected.  
      As we will see in the experiments, reasonable results can be
      obtained with only color. It is natural to combine 
      {\AnnealedMS} into a cascade structure, like the cascade detector
      of \cite{Viola04Robust}.  We start MS search from a
      large bandwidth $ h_0 $.  After convergence, an extra
      verification is applied to decide whether to terminate the
      search. If $ \sgn( f( \bI_0 ) ) = -1$, it means $ h_0 $ is too
      large. Then we need to reduce the bandwidth to $ h_1 $ and
      start MS  with the initial location $ \bI_0 $. This
      procedure is repeated until $ \sgn( f( \bI_m ) ) = +1$, $ m
      \in \{0,\cdots,M \} $.  $ h_m $ and $ \bI_m $ are the final
      scale and position.  
      %
      %
      Little extra computation is
      needed because only a decision verification is introduced at
      each stage.


%
%

\section{Experiments}
\label{sec:exp}

	In this section we implement a localizer and tracker and
	discuss related	issues. 
	Experimental results on various data sets are shown.
%
%

\begin{figure}[t!]
\centering
\fbox
{
\includegraphics[width=0.2083\textwidth]{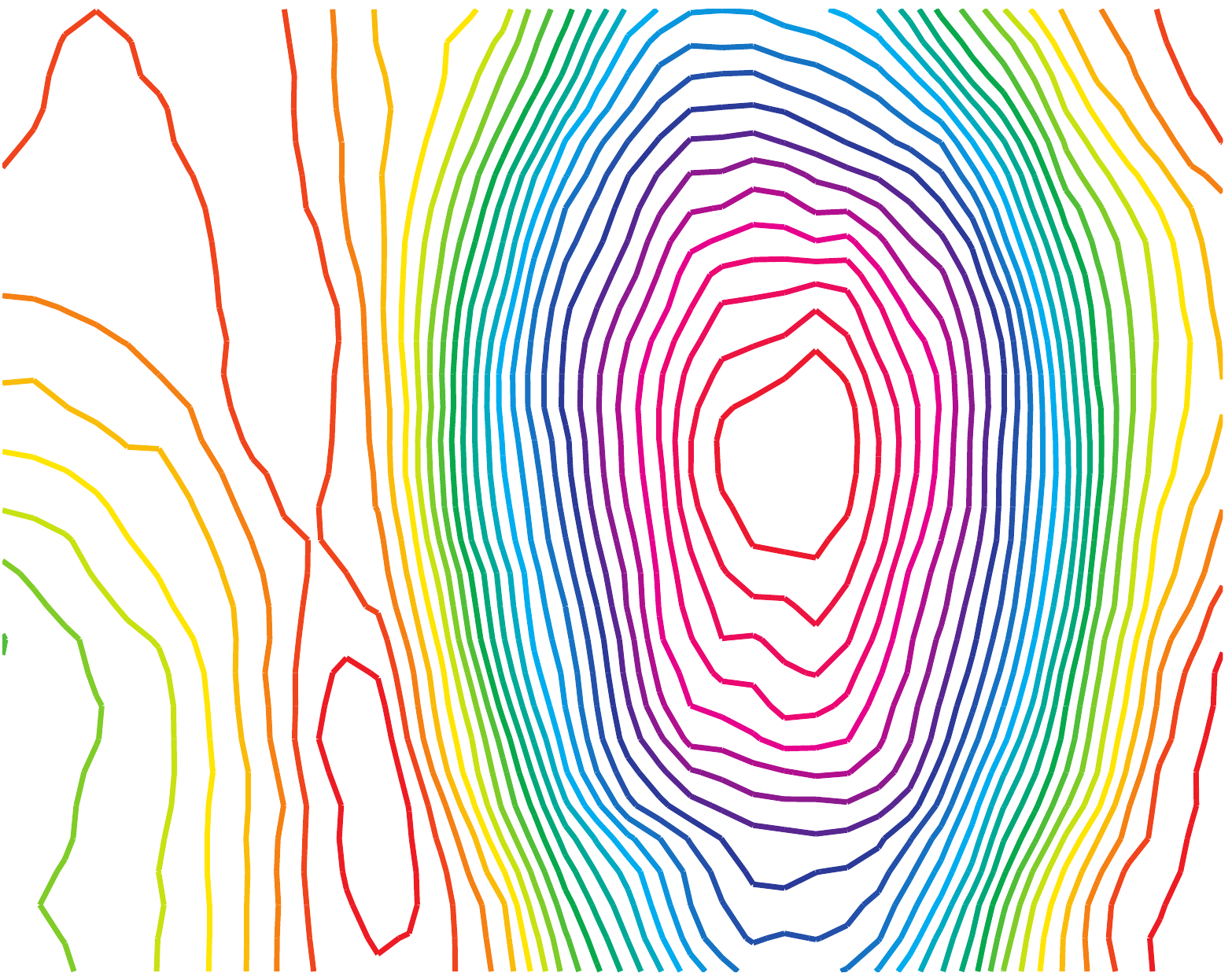}
}
\fbox
{
\includegraphics[width=0.2083\textwidth]{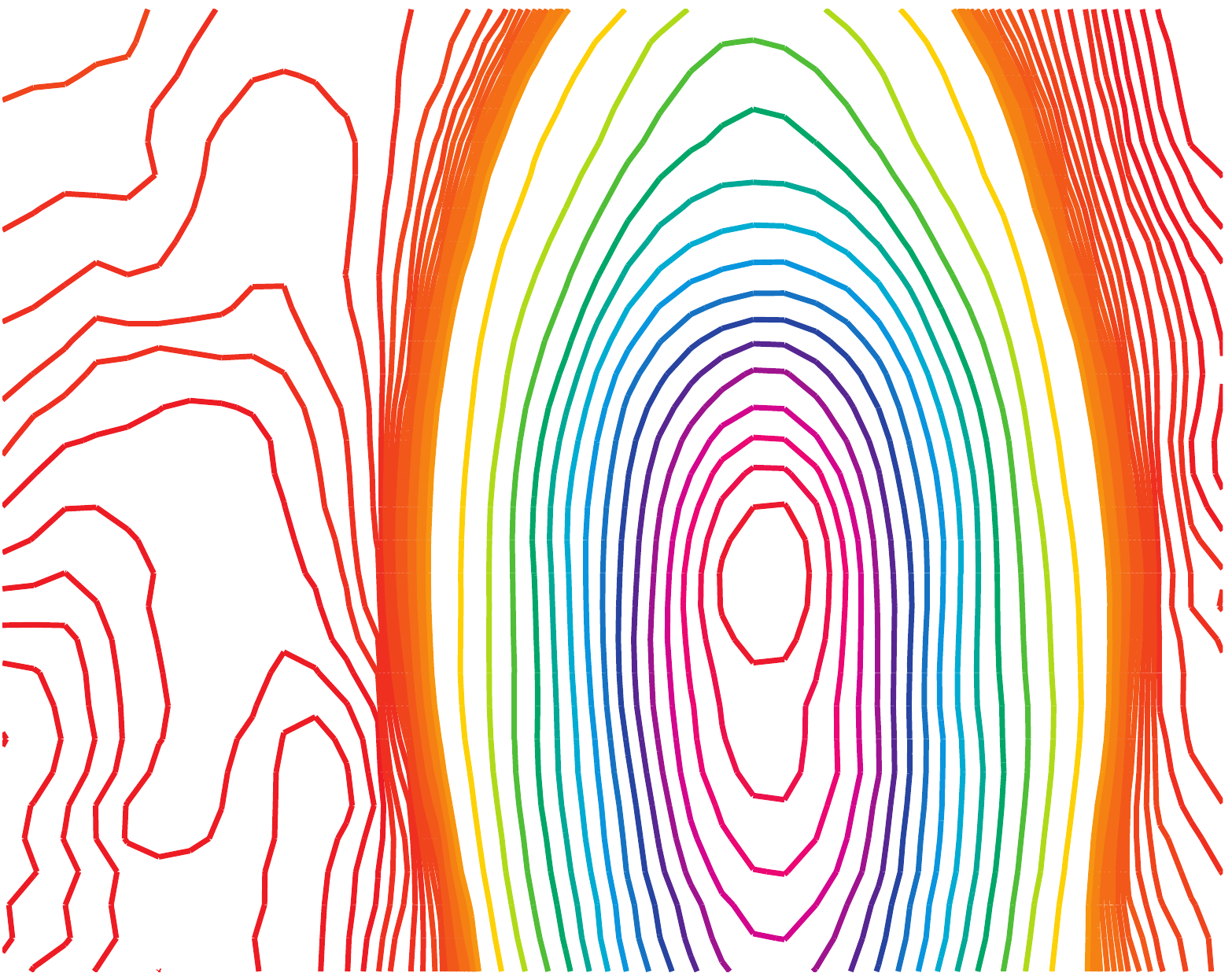}  
}
  \caption{A close look at the cost function of the first
  example in Fig.~\ref{fig:TrackingExp1}:
  (left) SVM score; (right) Bhattacharyya distance of 
  standard mean shift.
  Note that for the standard mean shift, the target model is extracted
  from the same test image; while for SVM, the target model is learned
  from a large number of training images that do not contain the test
  image.
  }
  \label{fig:1}
\end{figure}


     \begin{figure*}[h]
         \centering
         \fbox{
         \includegraphics[width=.3\textwidth]{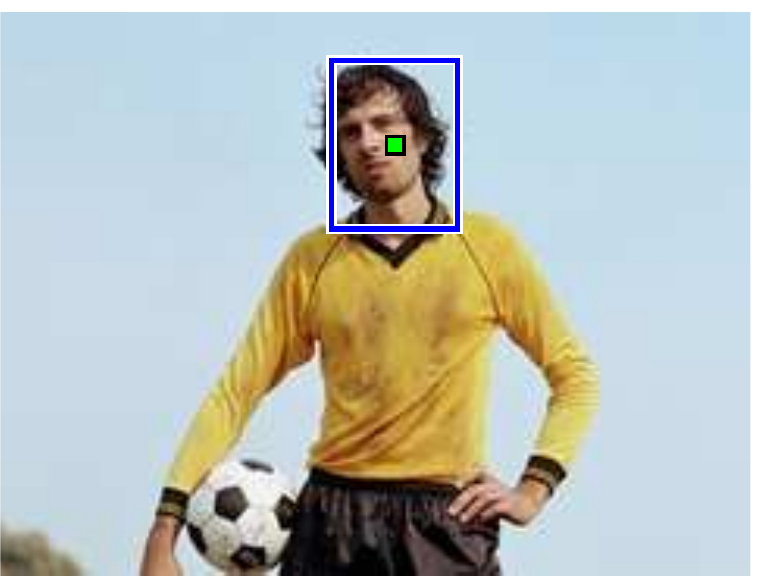}
         }
         \fbox
         {
         \includegraphics[width=.3\textwidth]{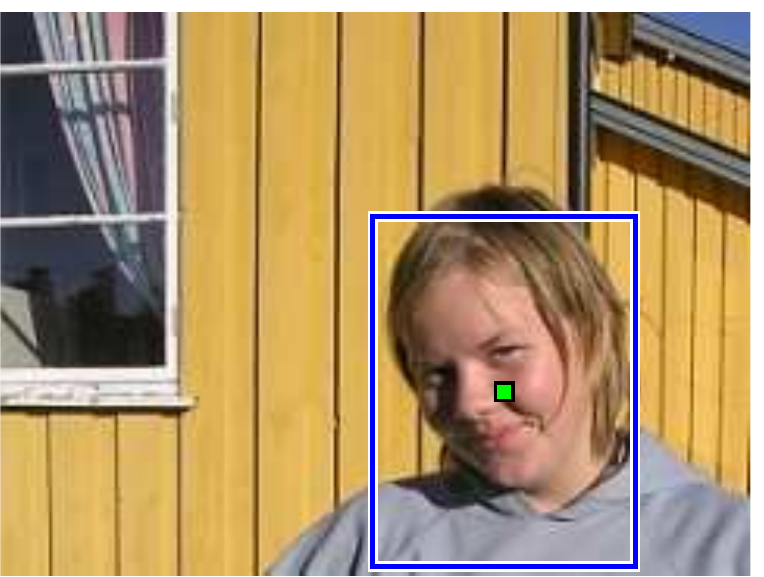}
         }
         \fbox
         {
         \includegraphics[width=.3\textwidth]{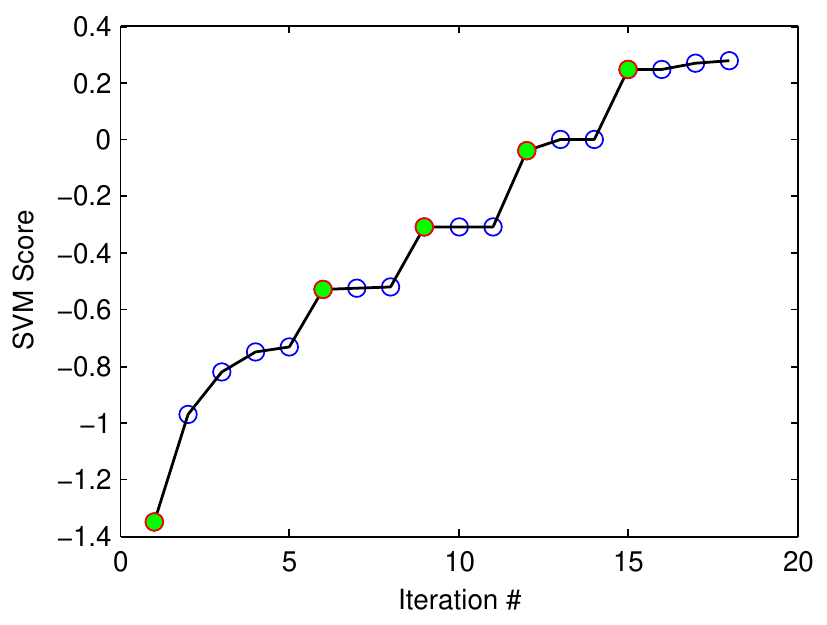}
         }
     \caption{Face localization. The final decision is marked with
     a rectangle. 
     The image size in all tests is $ 240 \times
     180 $. 
     In the first test (left), the proposed cascade
     localizer works very well. For the second one (middle), the detected
     scale of the target is slightly big, but acceptable.
     The SVM scores for the first example are also plotted (right). The
     first iteration at each bandwidth is marked with a solid
     circle.
     }
     \label{fig:LocalisationExp1}
     \end{figure*}

%
%

\subsection{Localization}

	For the first experiment, we have trained a face representation model.
	$ 404 $  faces cropped from CalTech-$ 101 $ 
	are used as positive
	raw images, and $ 1400 $  negative images are randomly cropped from
	images which do not contain faces. The image size is reduced
	to $ 42 \times 56 $ pixels.
	%
	%
    Kernel-weighted RGB colour histograms, consisting
	of $ 16 \times 16 \times 16 $ bins,
	are extracted for classification.
	By default we use a soft SVM trained with LIBSVM 
	(slightly modified to use customized kernels).
%
%
    Test accuracy on the
	training data is  $99.5\%~(1795/1804) $; and
	$91.7\%~(2752/3000) $ on a test data set which contains
	totally $ 3000 $ negative data.  
	Note that our main purpose is not to train a powerful face
	detector; rather, we want to obtain an appearance model that is more robust 
	than the single-view appearance model. 
%
%
	We now test how well the algorithm maximizes the SVM score.
	First, we feed the algorithm a rough initial guess
    %
    and run MS. See Fig.~\ref{fig:TrackingExp1}
	for details.

    The first example in Fig.~\ref{fig:TrackingExp1} comes
	from the training data set. The initial SVM score is
	negative. In this case, a single step is required to switch to a
	positive score---it moves closely to the target after
	one iteration. 
    We plot the corresponding cost function in Fig.~\ref{fig:1}.
    By comparison, the cost function of the standard MS 
	is also plotted (the target
	template is cropped from the same image). We can clearly see the
    difference. 
%
    The other two test images are from outside
	of the training data set. Despite the significant face color
	difference and variation in illumination, our SVM localizer
	works well in both tests. To compare the robustness, 
	we use the first face as a template to track the second face
    in Fig.~\ref{fig:TrackingExp1}, the standard MS  tracker
    fails to converge to the true position.

	We now apply the global maximum seeking algorithm to object
	localization. In \cite{Fast07Shen}, it has been shown that it is
	possible to locate a target no matter from which initial
	position the MS tracker starts. Here we use the learned
	classification rule to determine when to stop searching.
	We start the annealed continuation procedure with the initial
	bandwidth $ h_0 = ( 42 , 56 ) $. Then the bandwidth pyramid
	works with the rule $ h_{m+1} = \frac{h_m}{1.25} $,
	$ m \in \{ 0,\cdots,M \} $. 
    $ M $ is the maximum number of iterations.
    We stop the search when for some $ m $
    the SVM score is positive upon convergence. 
    The image center is set to be the
	initial position of the search for these $ 2 $ tests.
	We present the results in Fig.~\ref{fig:LocalisationExp1}.

	In the first test, our proposed algorithm works well: It
	successfully finds the face location, and also the final
	bandwidth well fits the target.
	Fig.~\ref{fig:LocalisationExp1} (right) shows how the SVM score
	evolves. 
	It can be seen that every bandwidth change significantly
	increases the score. 
    If the target size
	is large and there is a significant overlap between the target
	and a search region at a coarse bandwidth, $ h_m $, the overlap
	can make the cascade search stop {\em prematurely} (see
	the second test in Fig.~\ref{fig:LocalisationExp1}).
	Again this problem is mainly caused by the color feature's weak
	discriminative power. A remedy is to include more features.
	However, for certain applications where the scale-size is not
    critically important, our localization results
	have been usable.
    Furthermore, better results could be achieved
	when we train a model for a specific object (\eg, train an
	appearance model for a specific person) with a single color
	feature. 

\subsection{Tracking}

   Effectiveness of the proposed generalized kernel-based tracker is
   tested on a number of video sequences. We have compared with two
   popular color histogram based methods: 
   the standard MS tracker \cite{DVP2003KernelTrack}
   and particle filters \cite{P02Color}.

    Unlike the first experiment, we do not train an {\em off-line} SVM
    model for tracking.  
    It is not easy to have a large amount of training data for a
    general object, therefore 
    in the tracking
	experiment, an on-line SVM described in \S\ref{sec:onlineSVM}
    is used for training. The user crops several negative data
    and positive data for initial training. During the course of
    tracking the on-line SVM updates its model by regarding the
    tracked region as a positive example and randomly selecting a few
    sub-regions (background area) around the target as negative examples.
    A $ 16 \times 16 \times 16 $-binned color histogram is used
    for both the generalized kernel tracker and standard MS tracker.
    For the particle filter, with $1000$ or $800$
    particles, the tracker fails at the first a few frames. So we
    have used 
    $1500$ particles. 

%



\begin{figure}[ht!]
\centering
\includegraphics[width=0.117\textwidth]{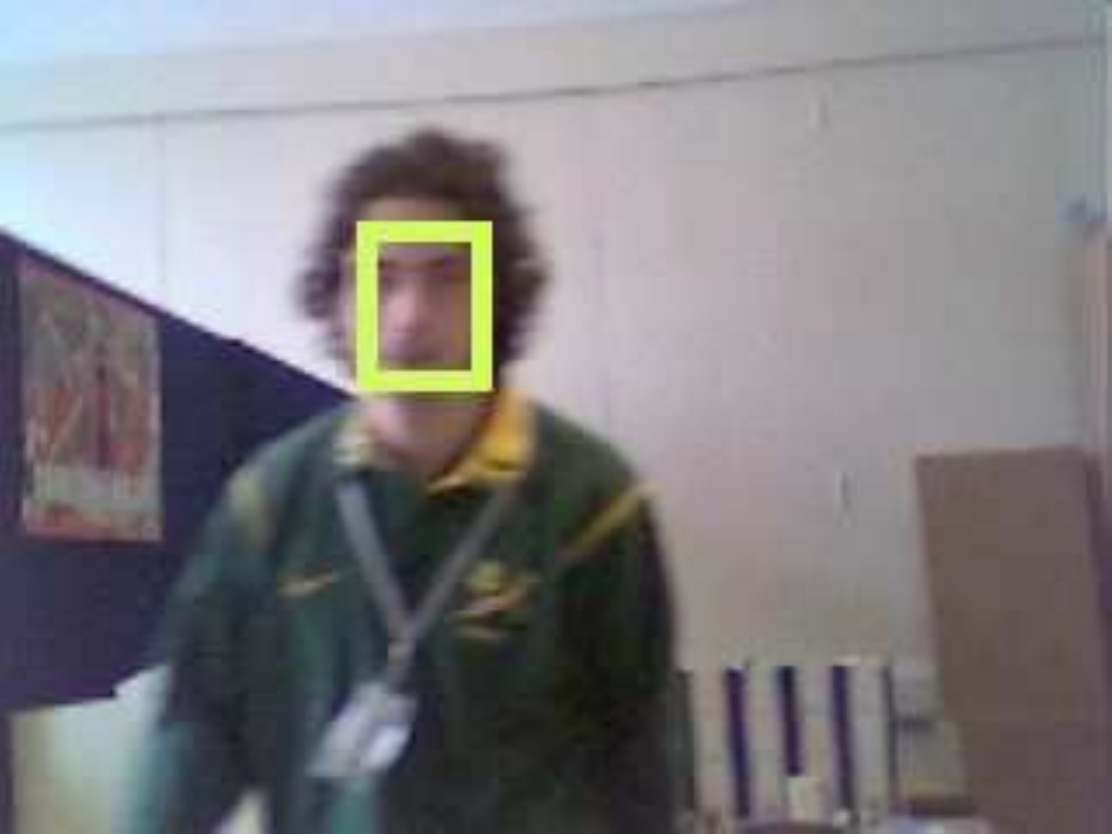}
\includegraphics[width=0.117\textwidth]{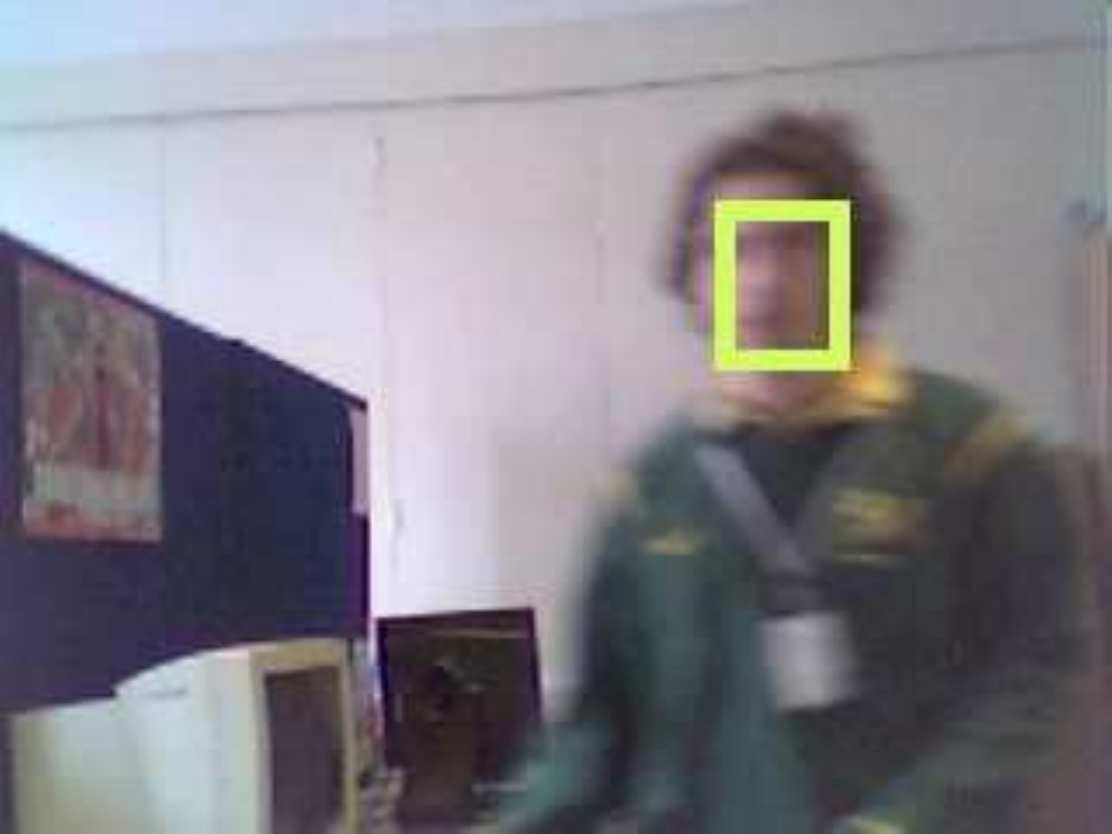}
\includegraphics[width=0.117\textwidth]{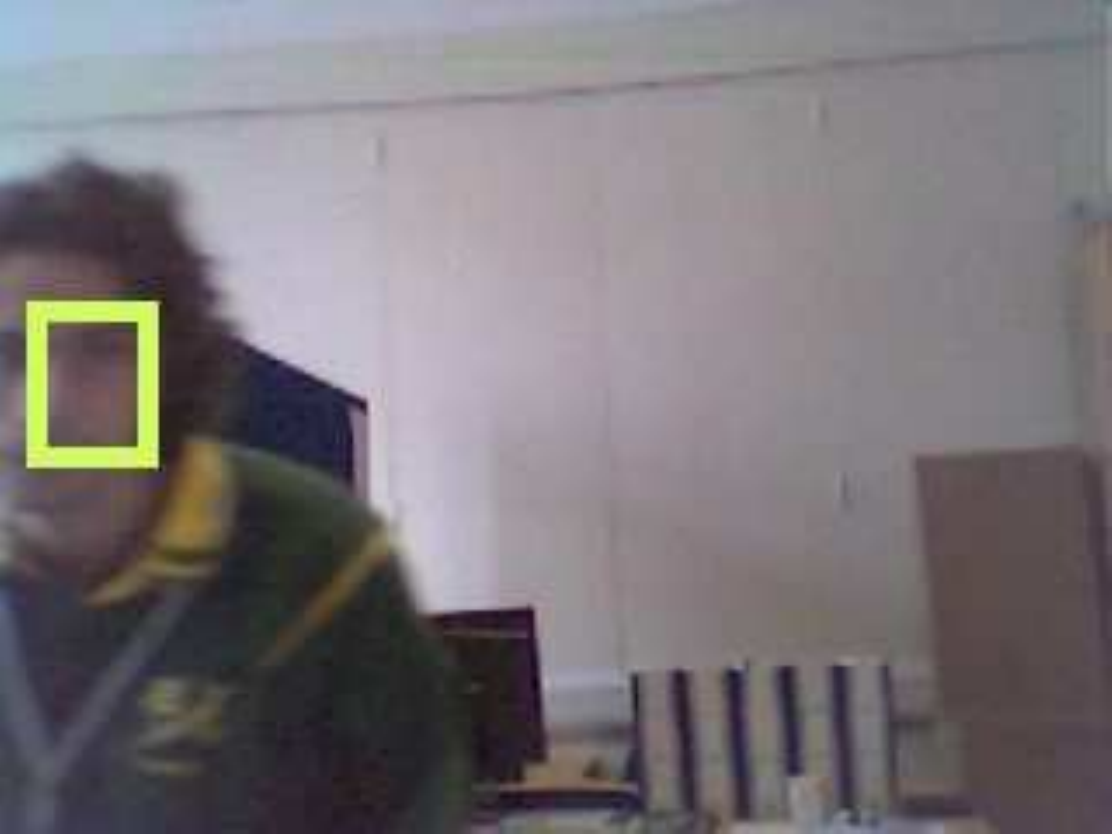}
\includegraphics[width=0.117\textwidth]{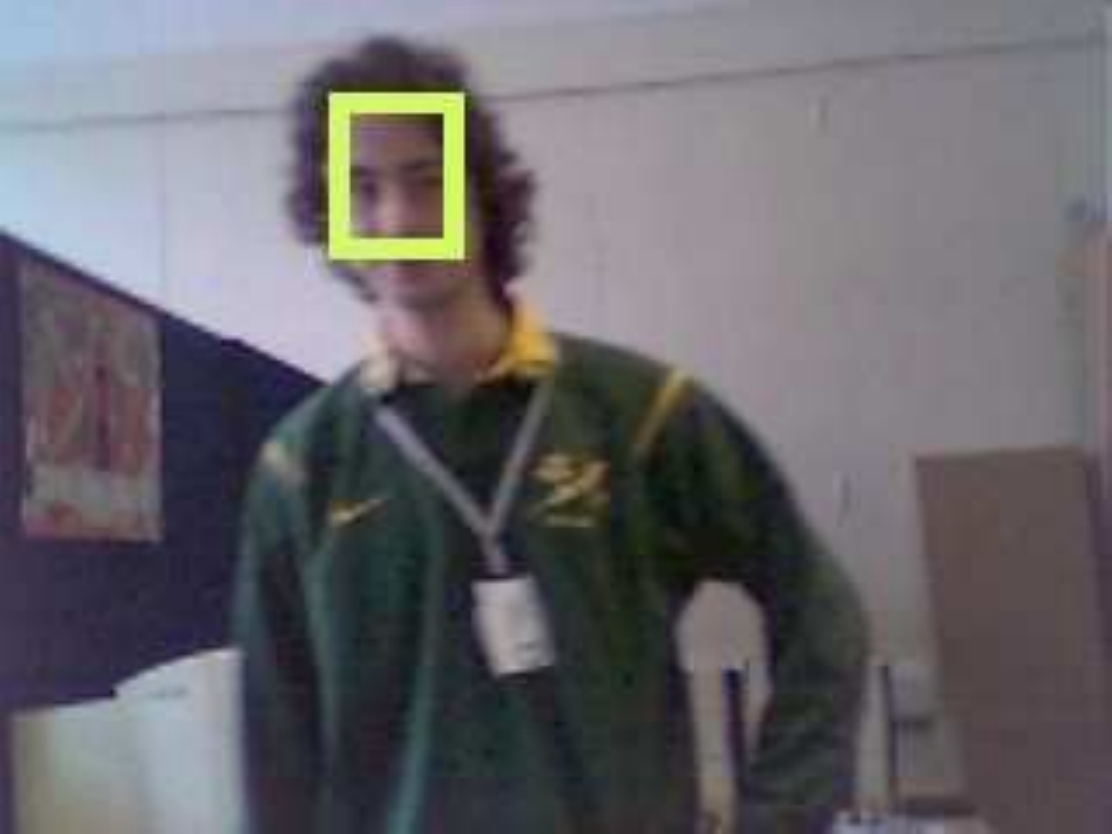}
\\
\includegraphics[width=0.117\textwidth]{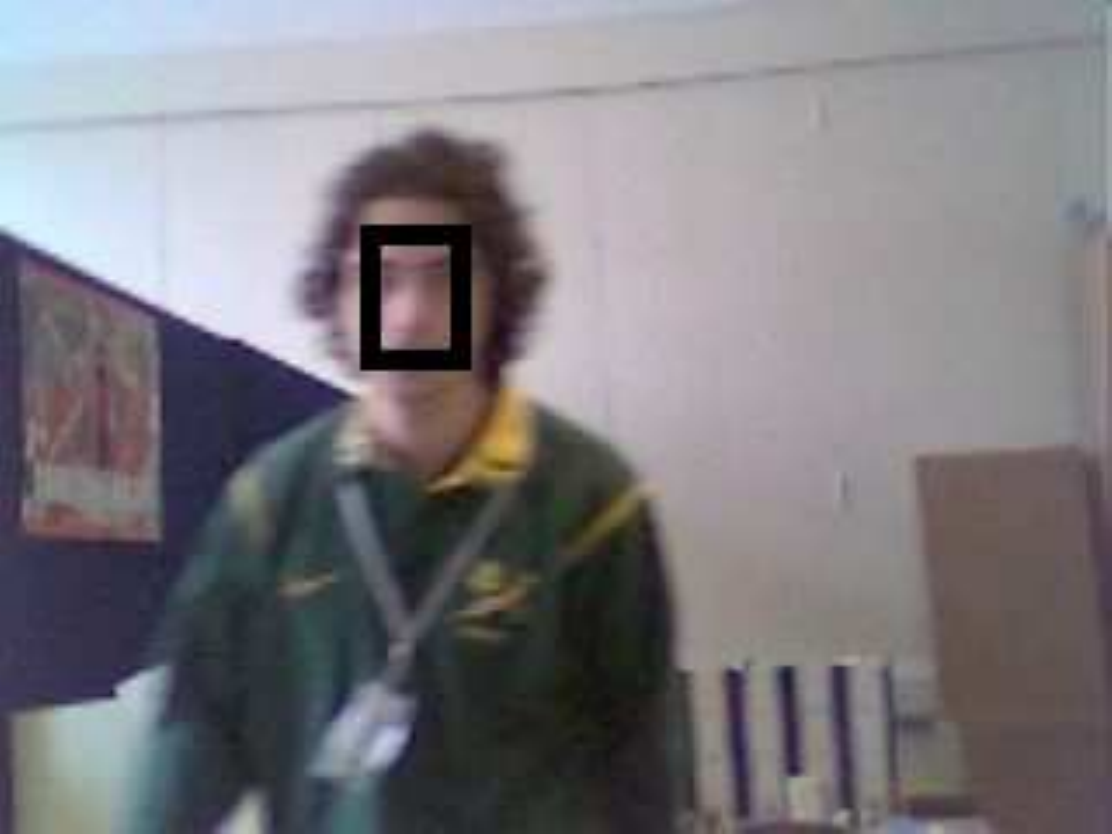}
\includegraphics[width=0.117\textwidth]{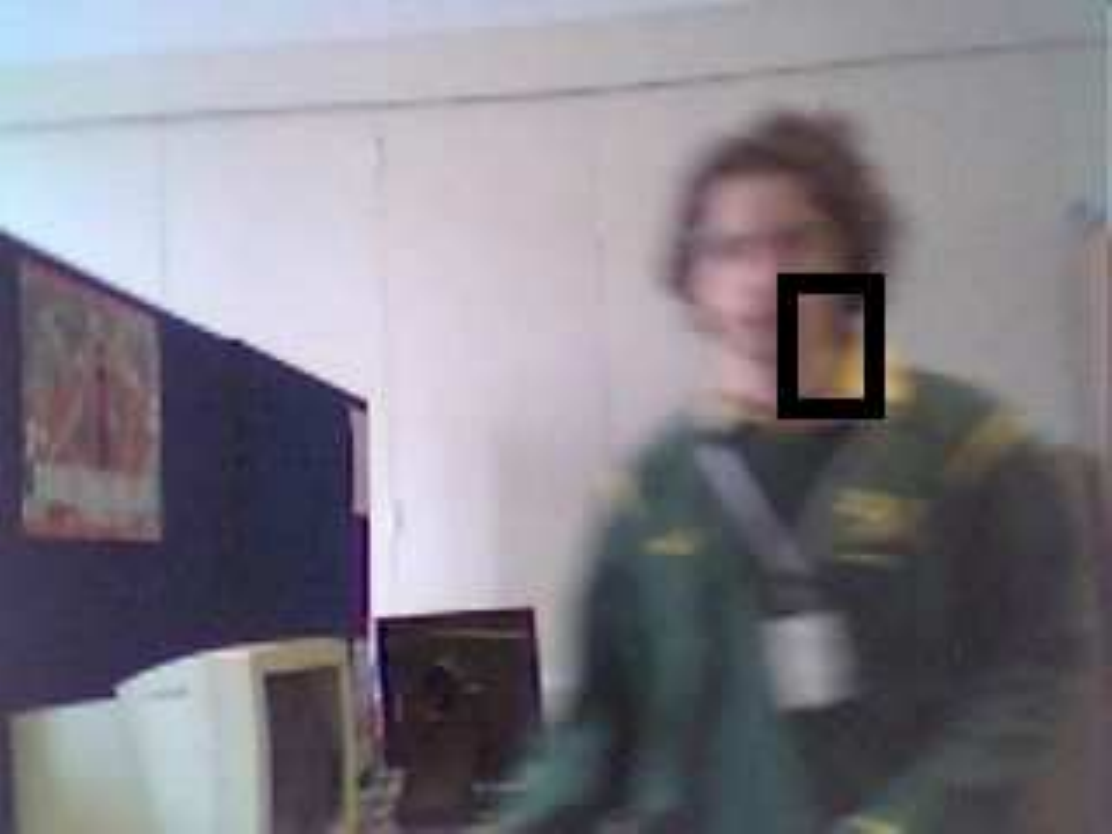}
\includegraphics[width=0.117\textwidth]{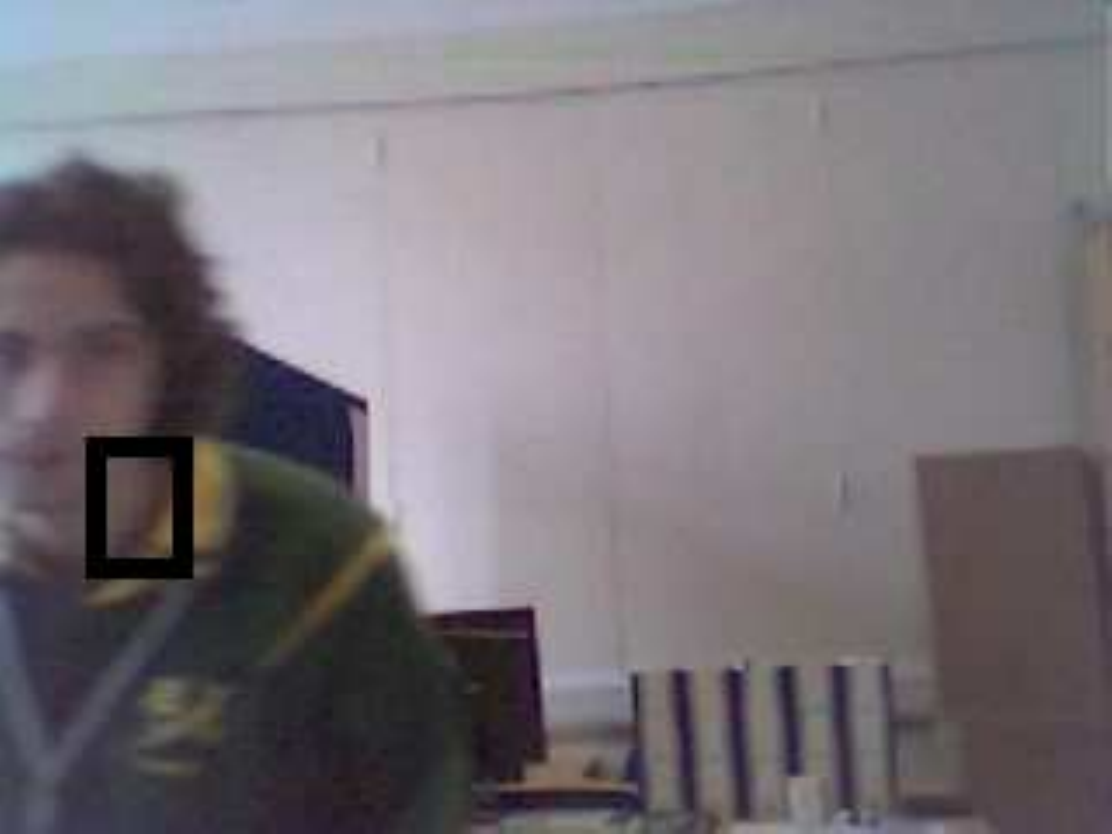}
\includegraphics[width=0.117\textwidth]{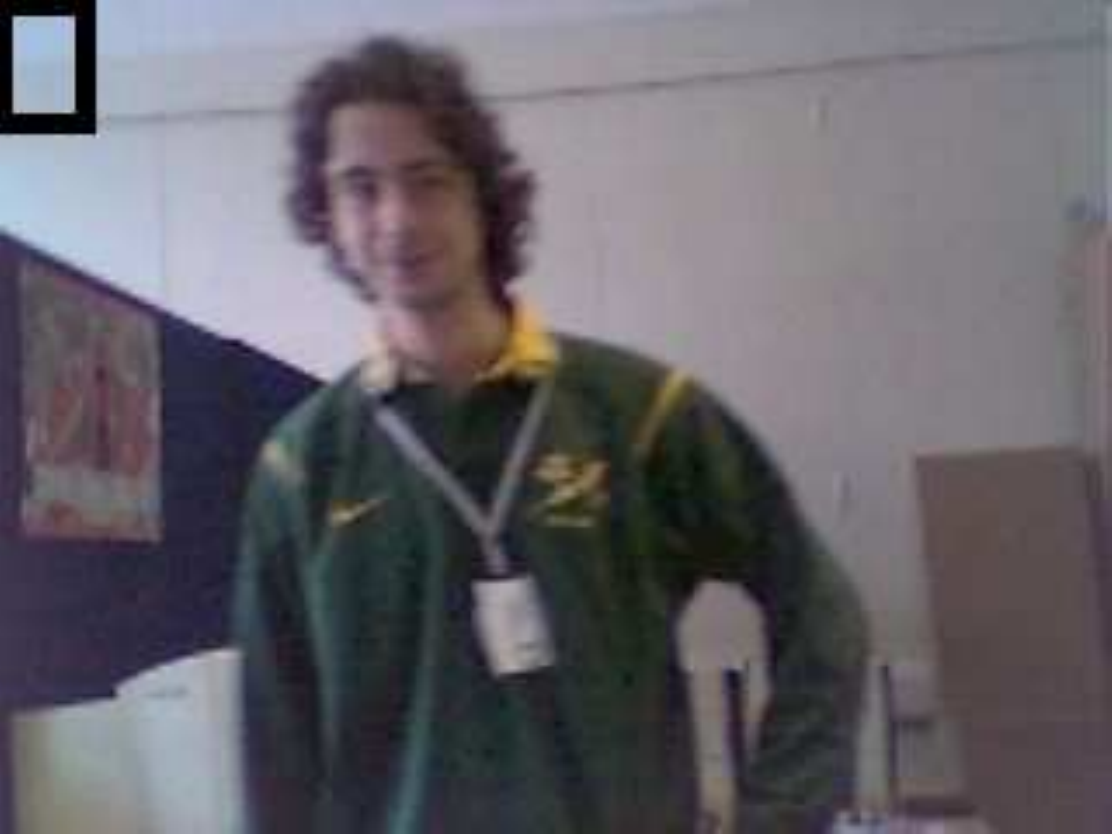}
\\
\includegraphics[width=0.117\textwidth]{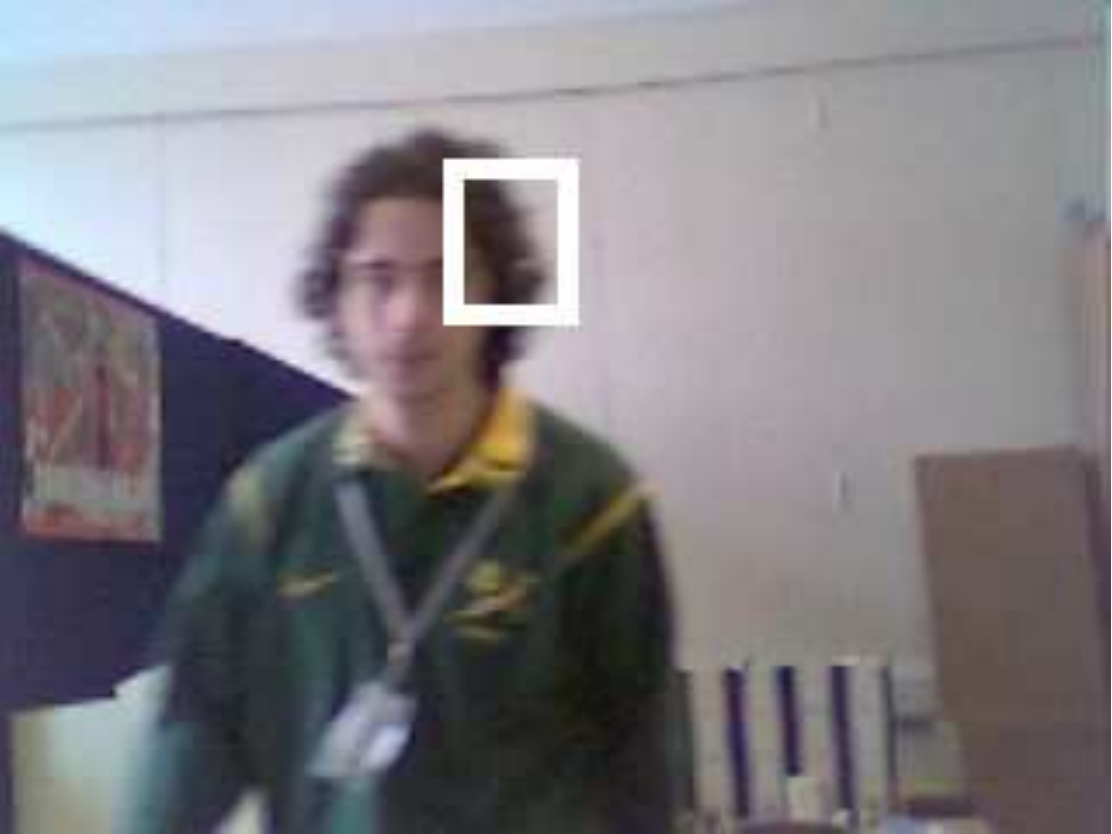}
\includegraphics[width=0.117\textwidth]{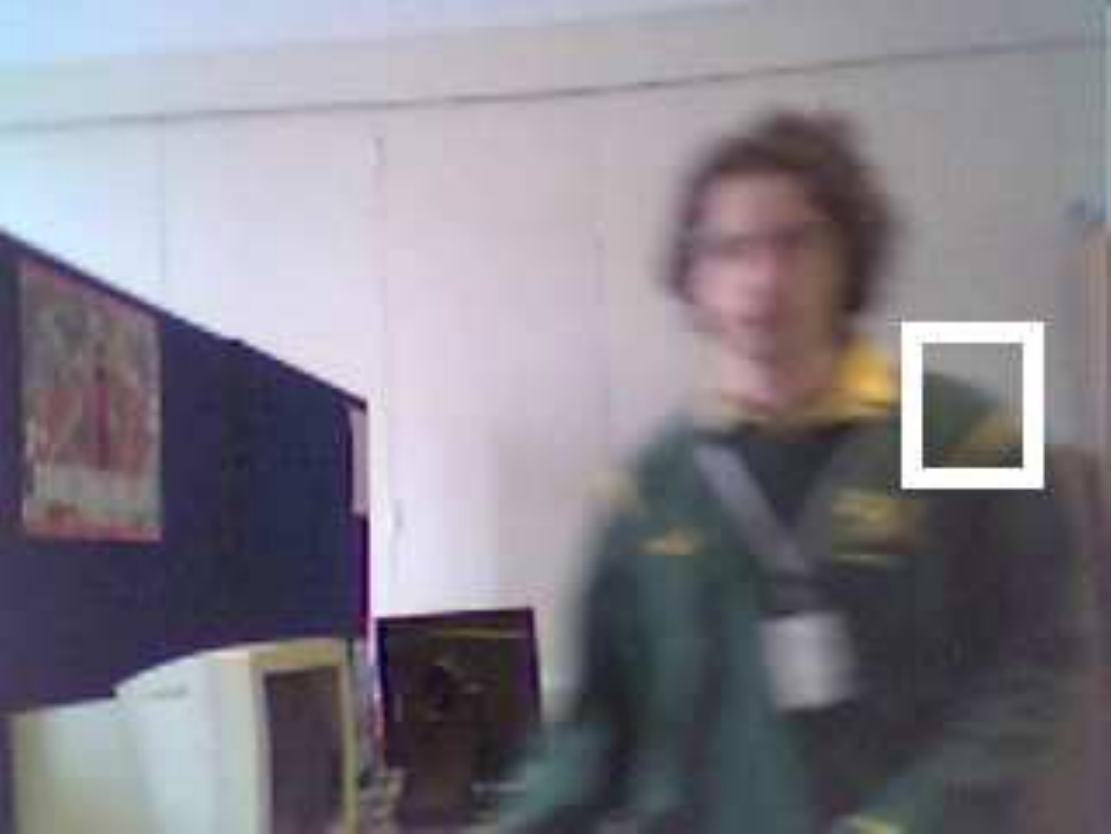}
\includegraphics[width=0.117\textwidth]{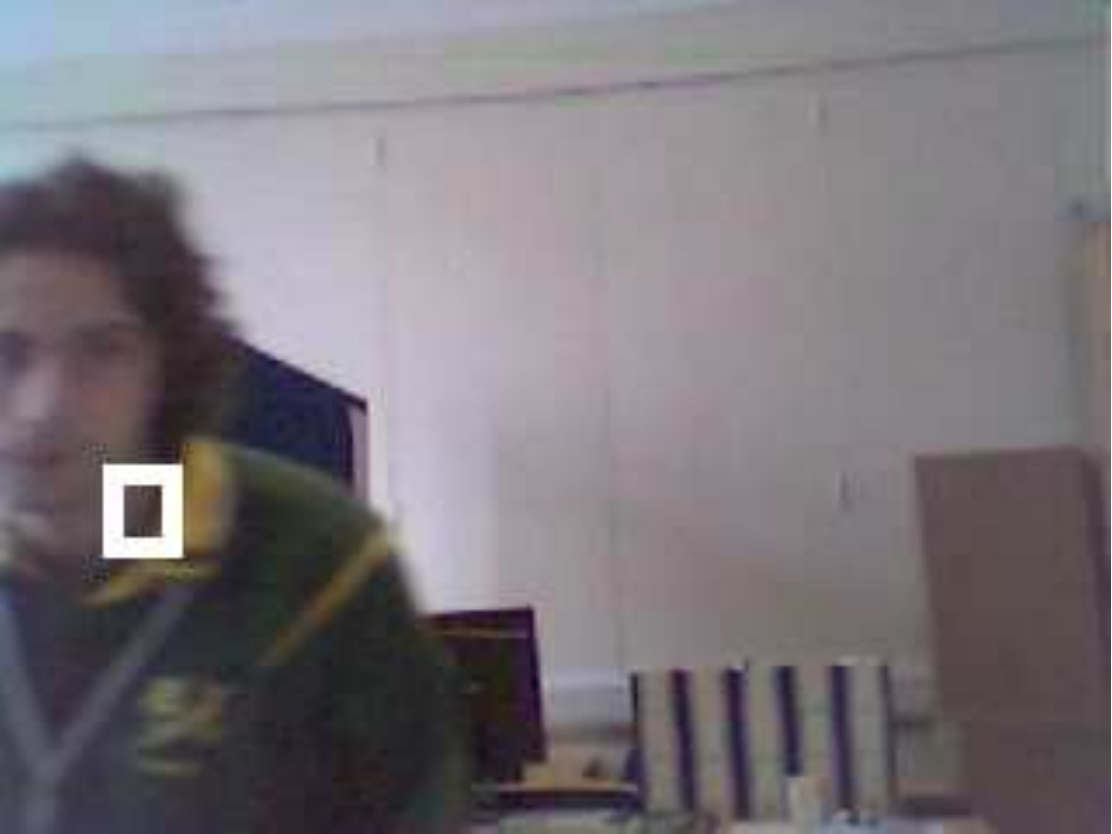}
\includegraphics[width=0.117\textwidth]{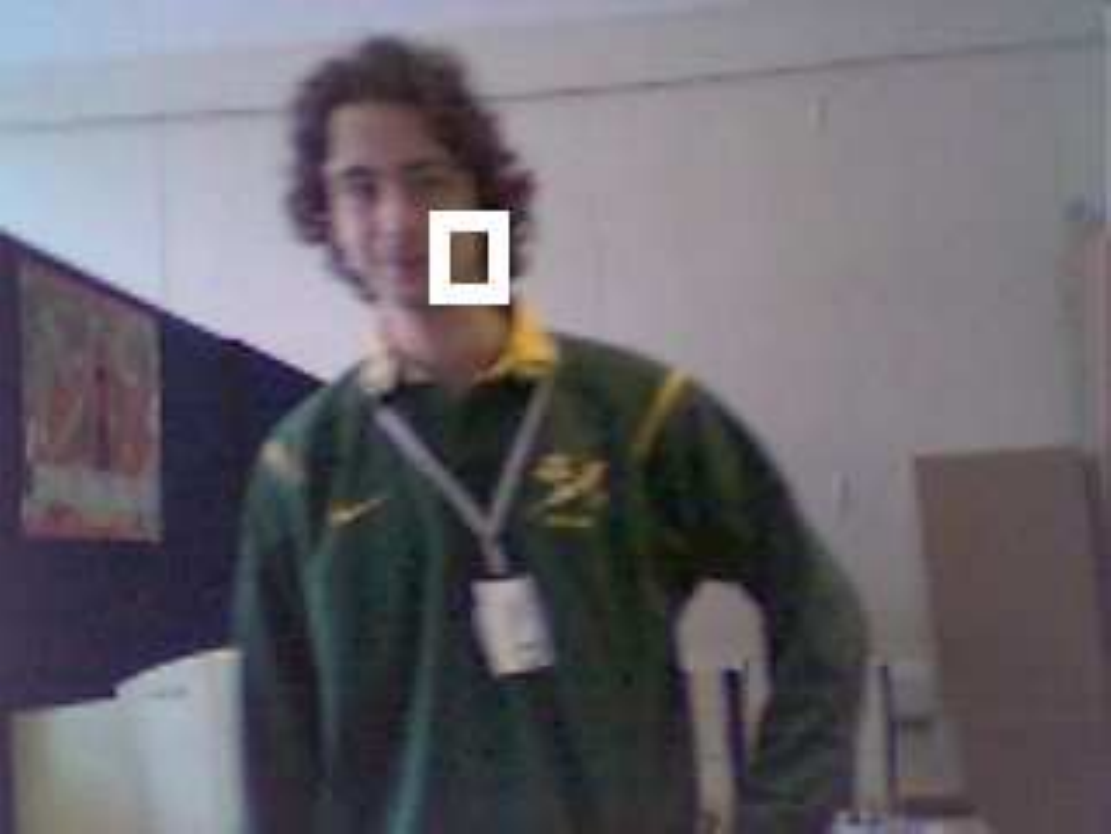}
\caption{ {\tt Face} sequence 1. Tracking results of the proposed tracker (top row);
  standard mean shift tracker (middle) and particle filtering (bottom row).
  Frames 26, 56, 318, 432 are shown. The video size is $ 320 \times 240 $ 
  and the frame rate is $10$ frames per second (FPS).
  }
  \label{fig:tracking1}
\end{figure}



\begin{figure}[!t]
\centering
\includegraphics[width=0.117\textwidth]{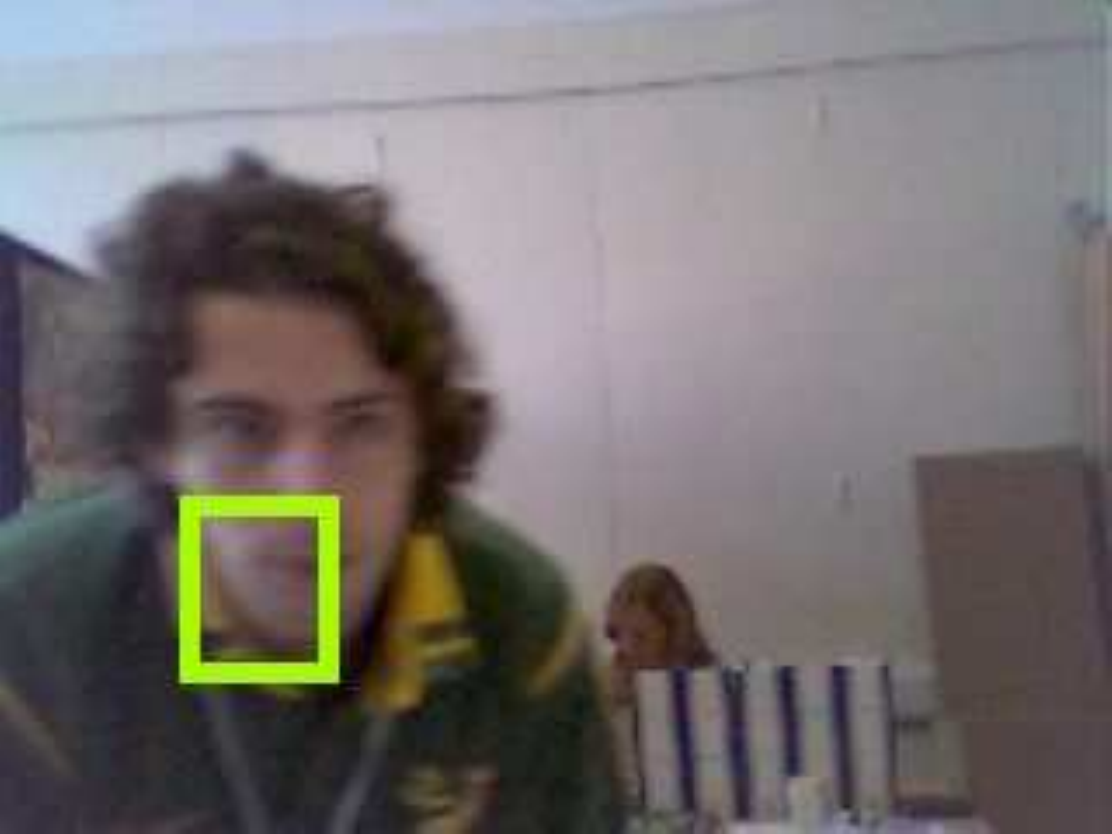}
\includegraphics[width=0.117\textwidth]{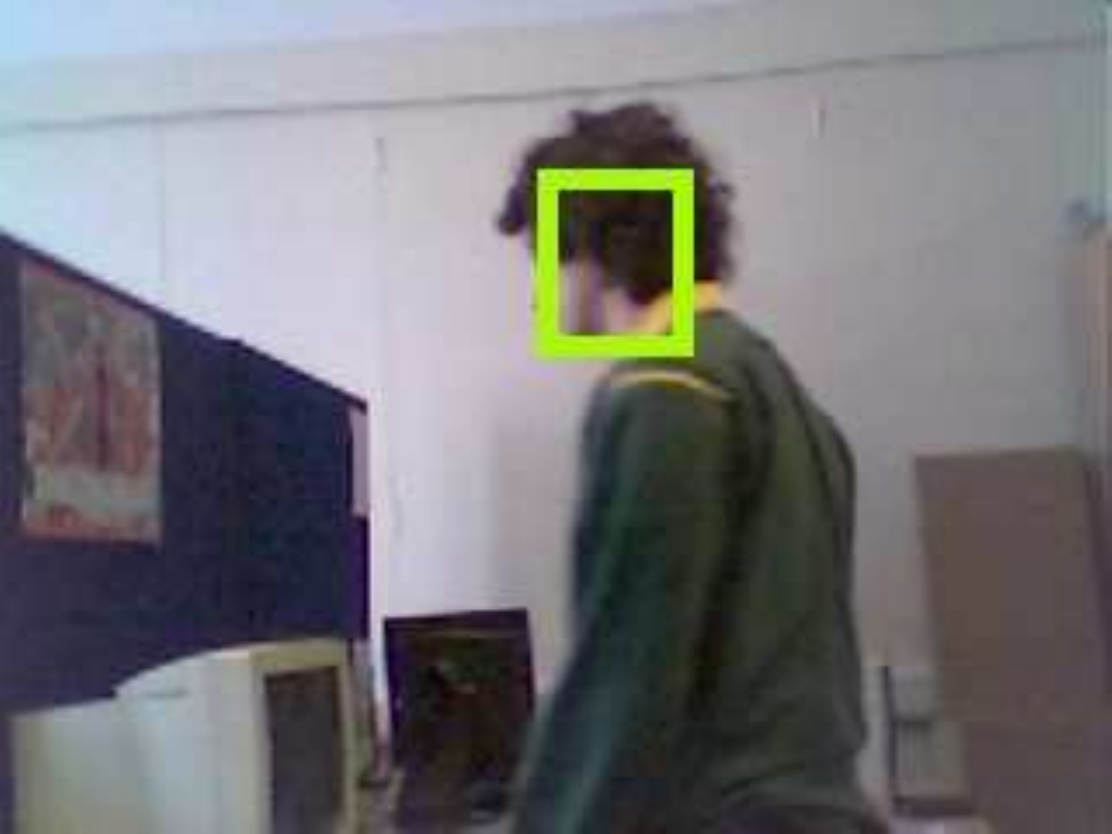}
\includegraphics[width=0.117\textwidth]{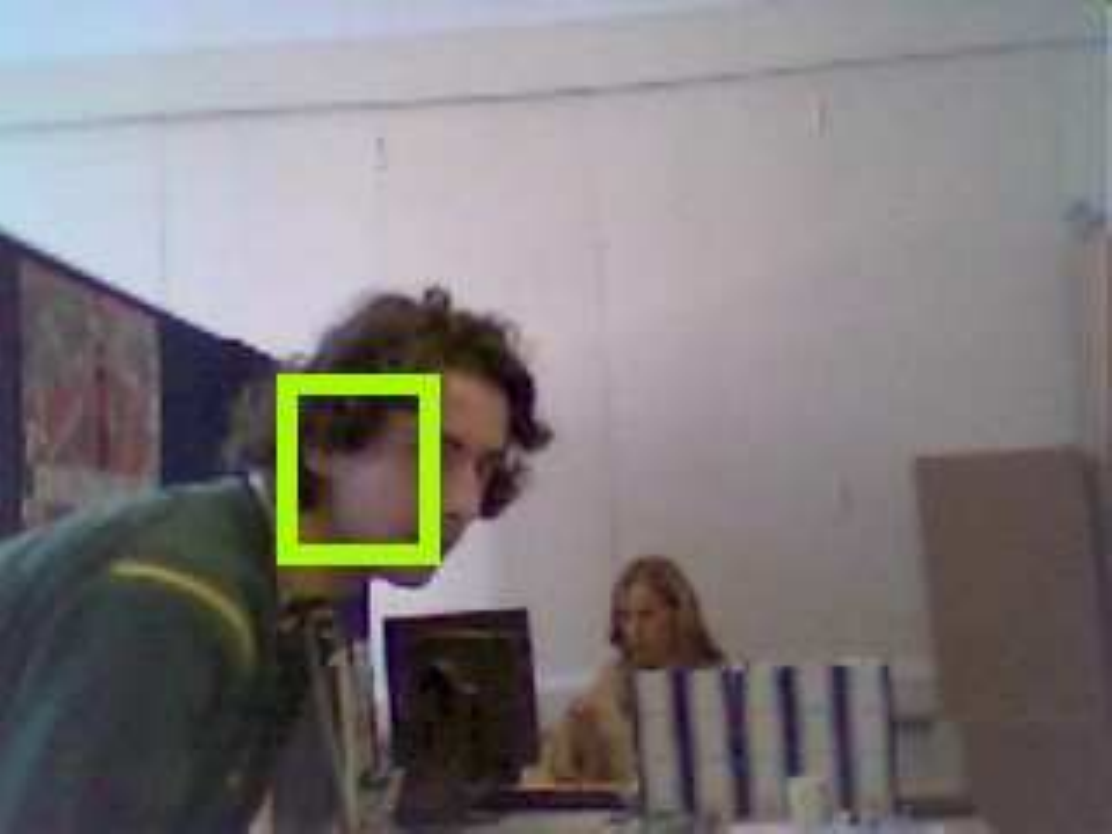}
\includegraphics[width=0.117\textwidth]{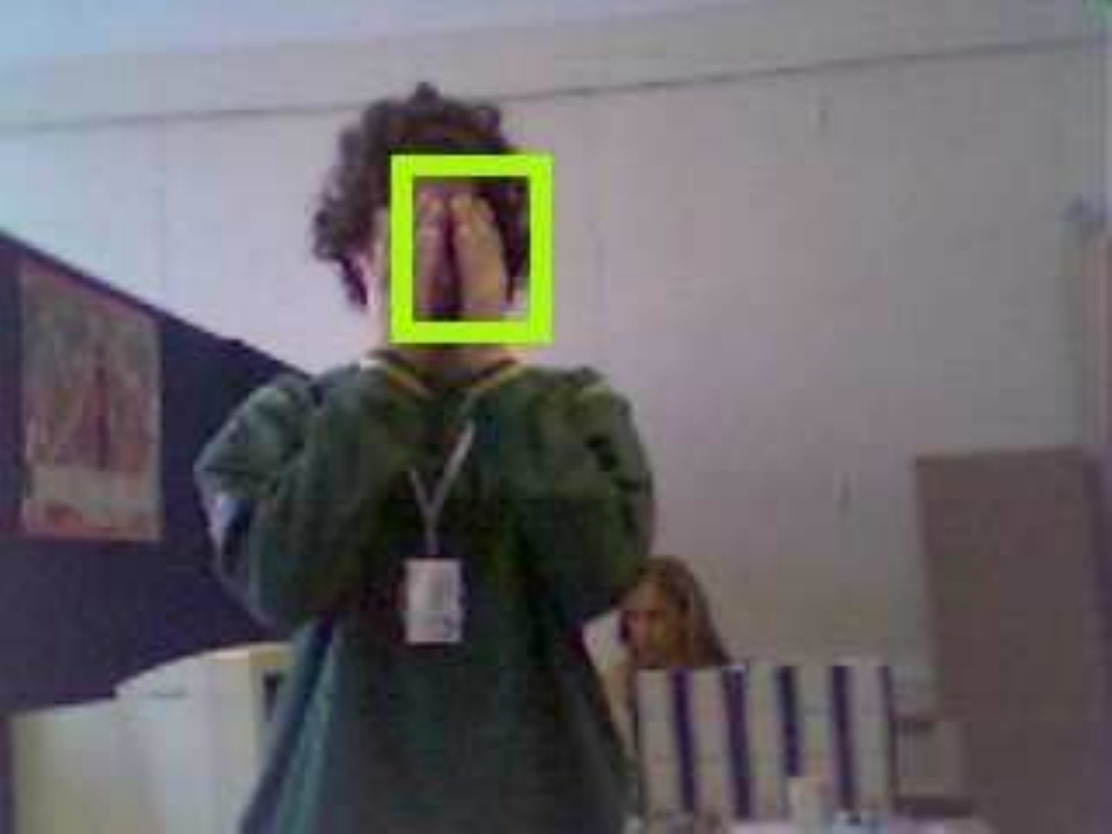}
\\
\includegraphics[width=0.117\textwidth]{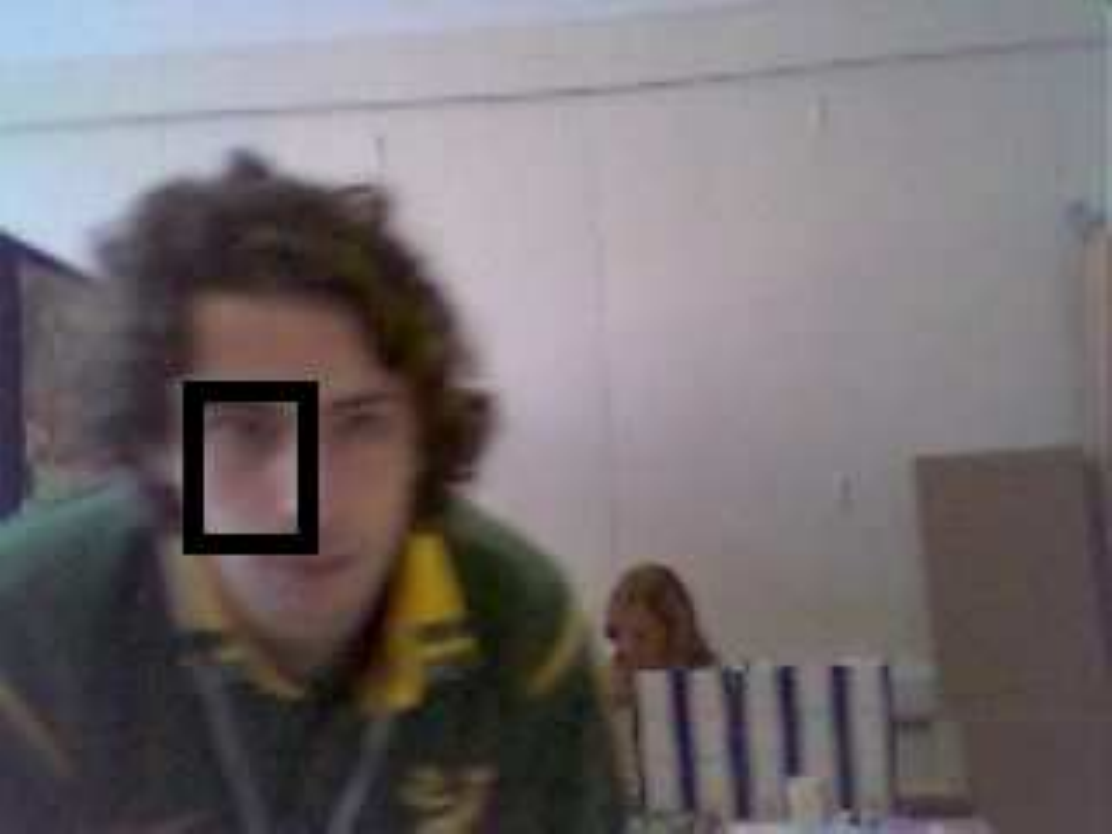}
\includegraphics[width=0.117\textwidth]{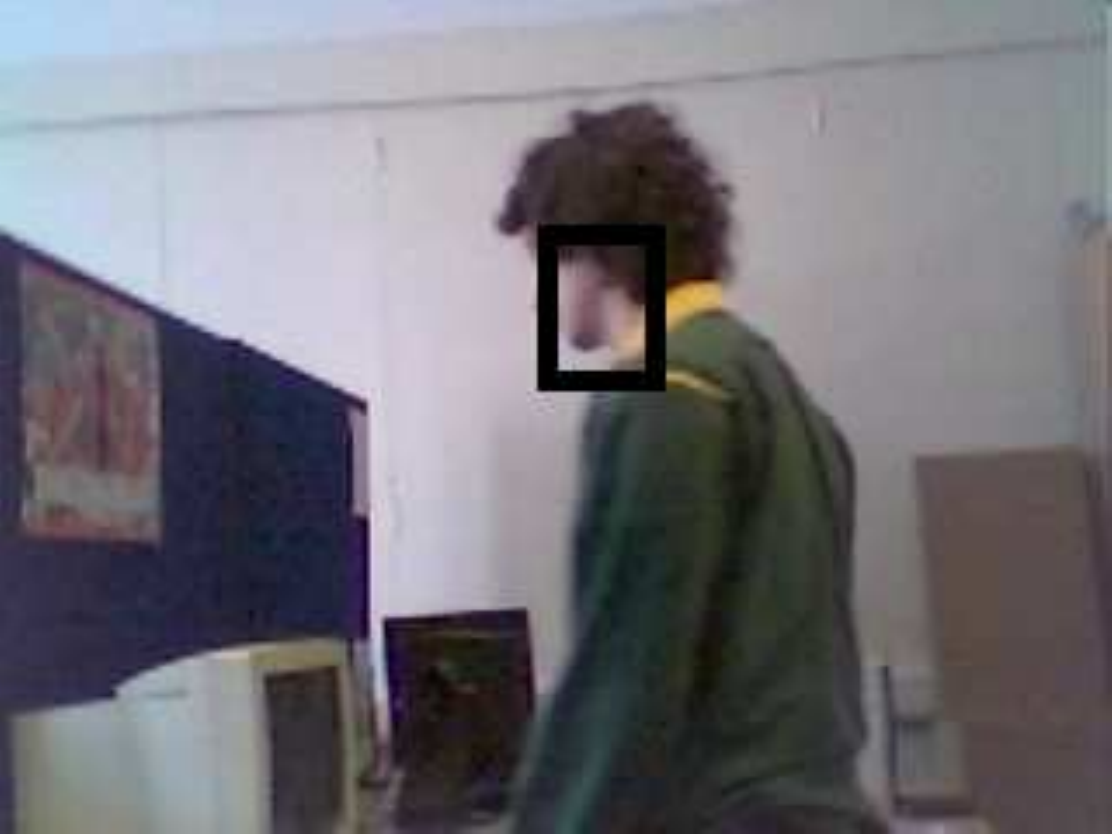}
\includegraphics[width=0.117\textwidth]{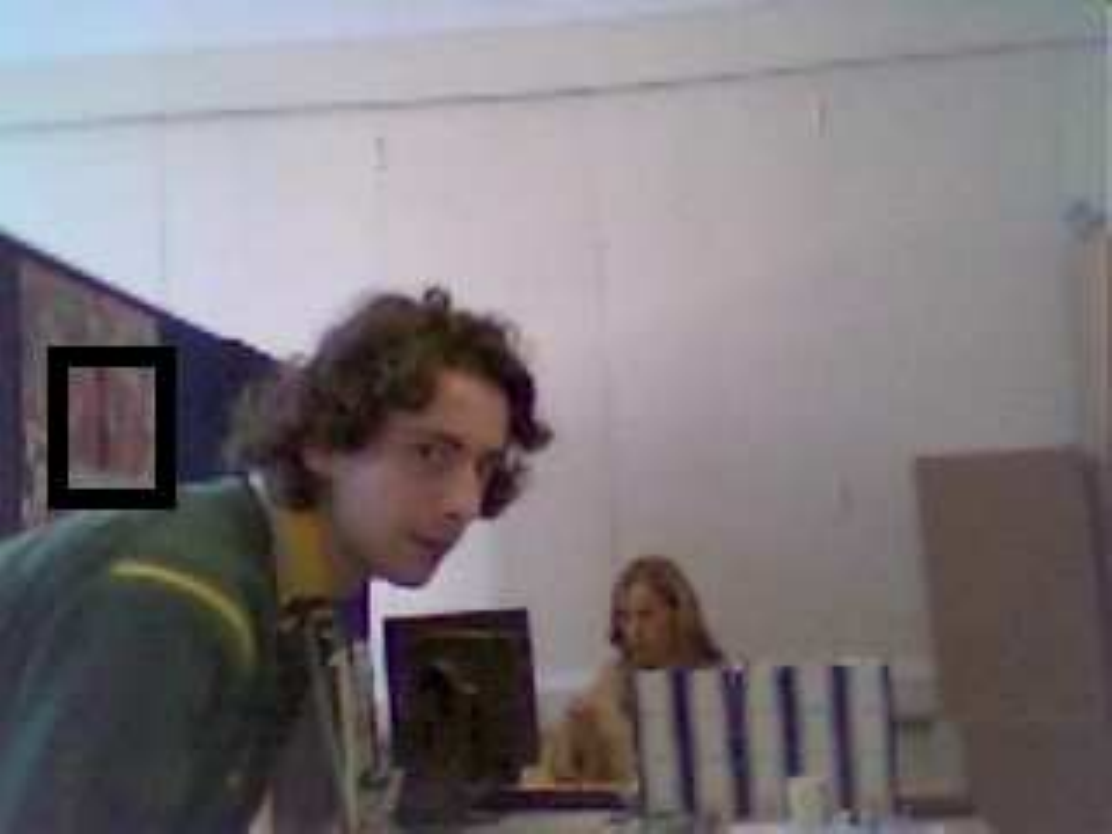}
\includegraphics[width=0.117\textwidth]{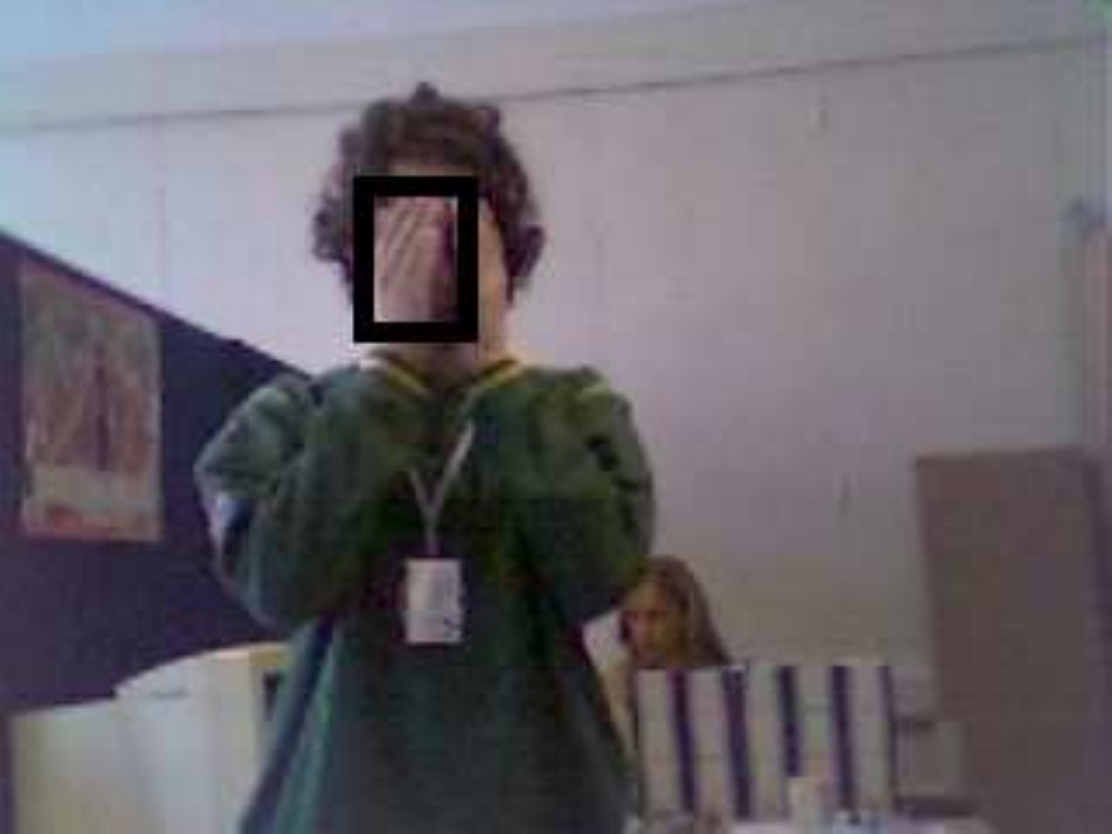}
\\
\includegraphics[width=0.117\textwidth]{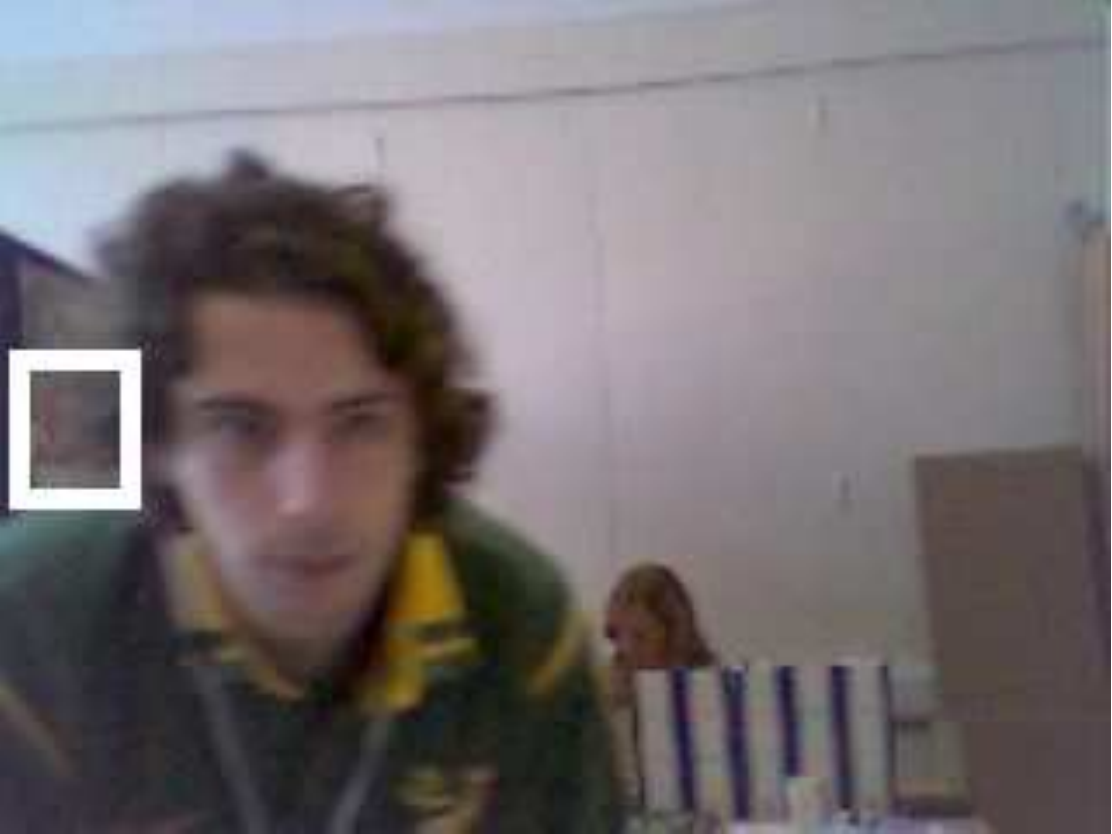}
\includegraphics[width=0.117\textwidth]{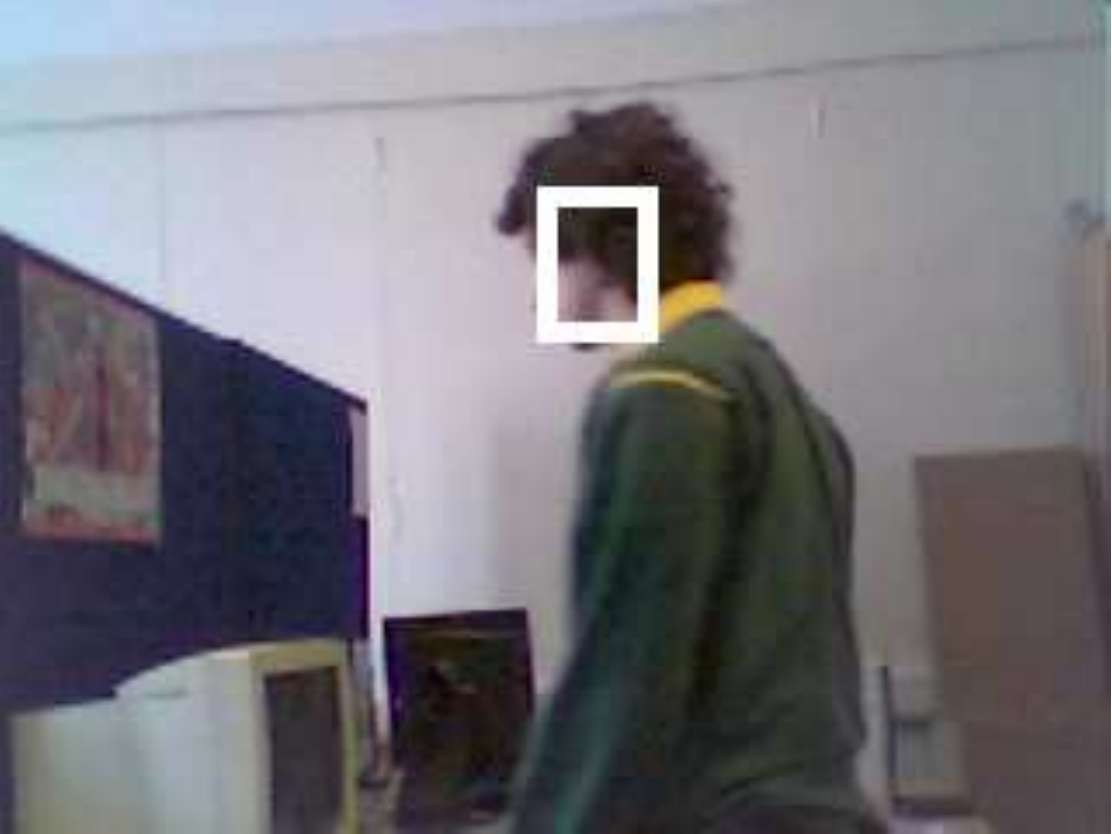}
\includegraphics[width=0.117\textwidth]{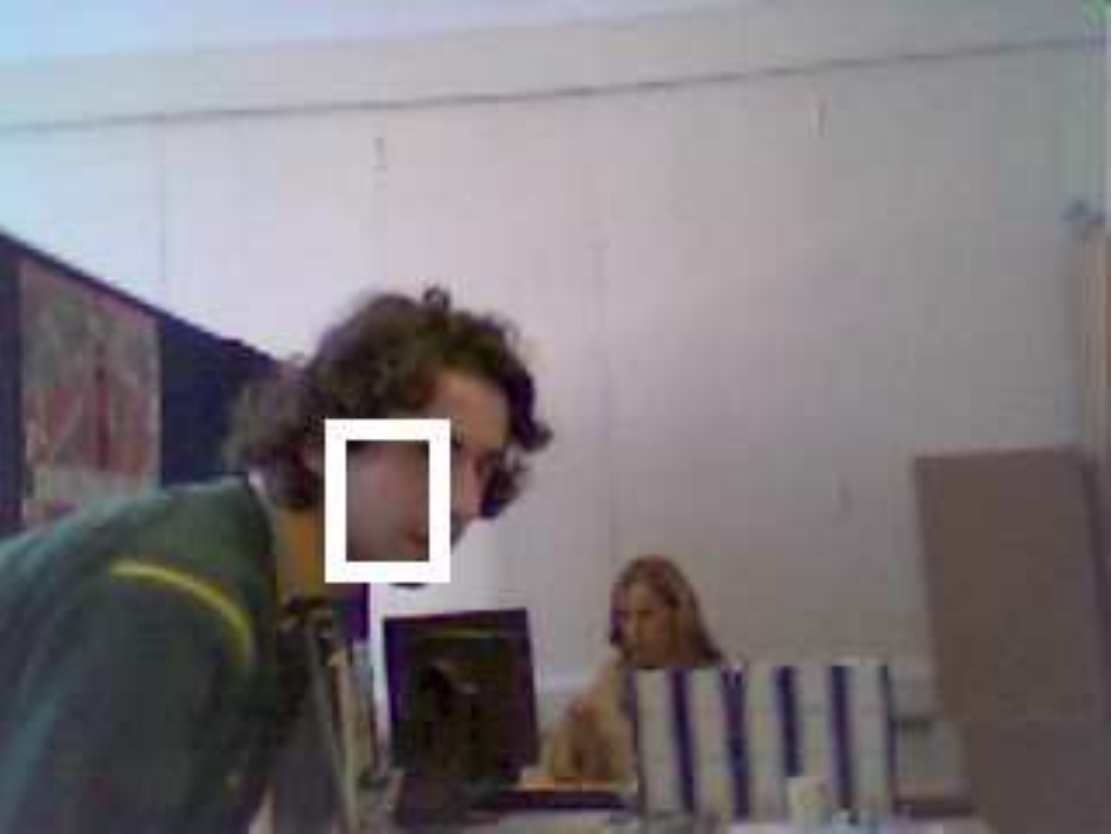}
\includegraphics[width=0.117\textwidth]{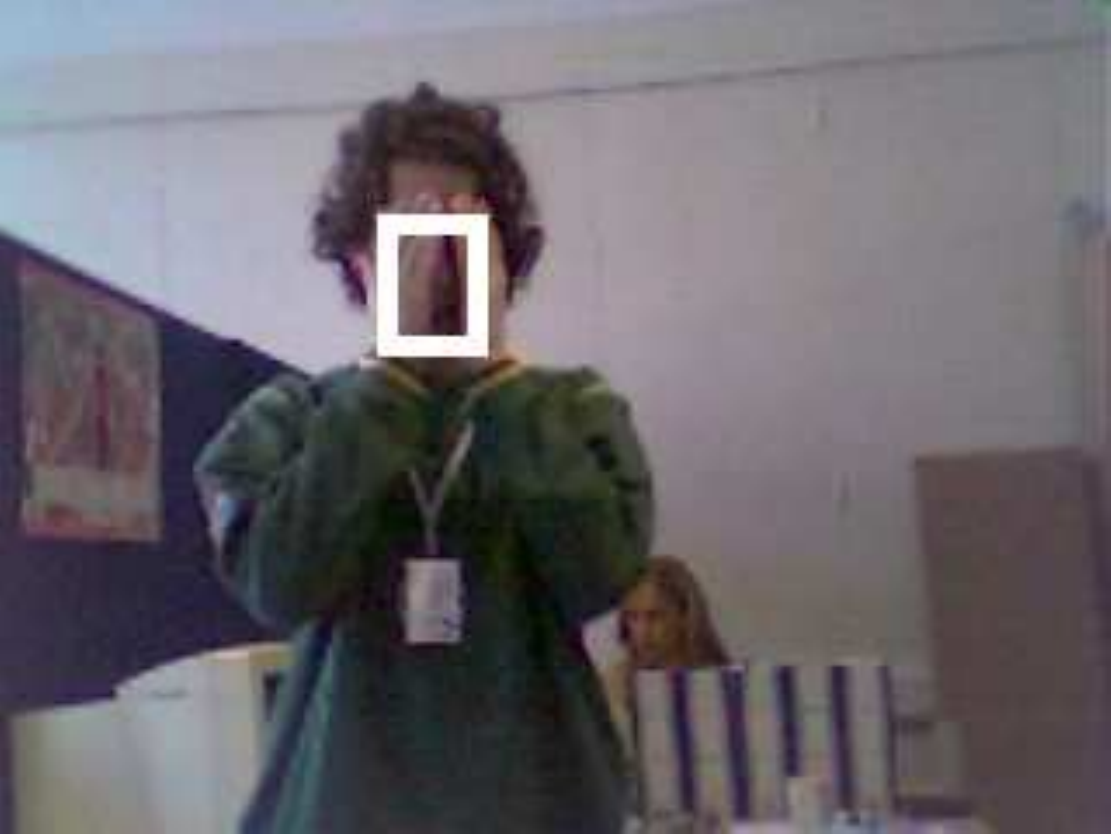}
\caption{ {\tt Face} sequence 2. Tracking results of the proposed tracker (top row);
  standard mean shift tracker (middle) and particle filtering (bottom row).
  Frames 86, 135, 204, 512 are shown. 
  The video size is $ 320 \times 240 $ 
  and the frame rate is $10$ FPS.
  }
  \label{fig:tracking2}
\end{figure}


\begin{figure*}[]
\centering
\includegraphics[width=0.16\textwidth]{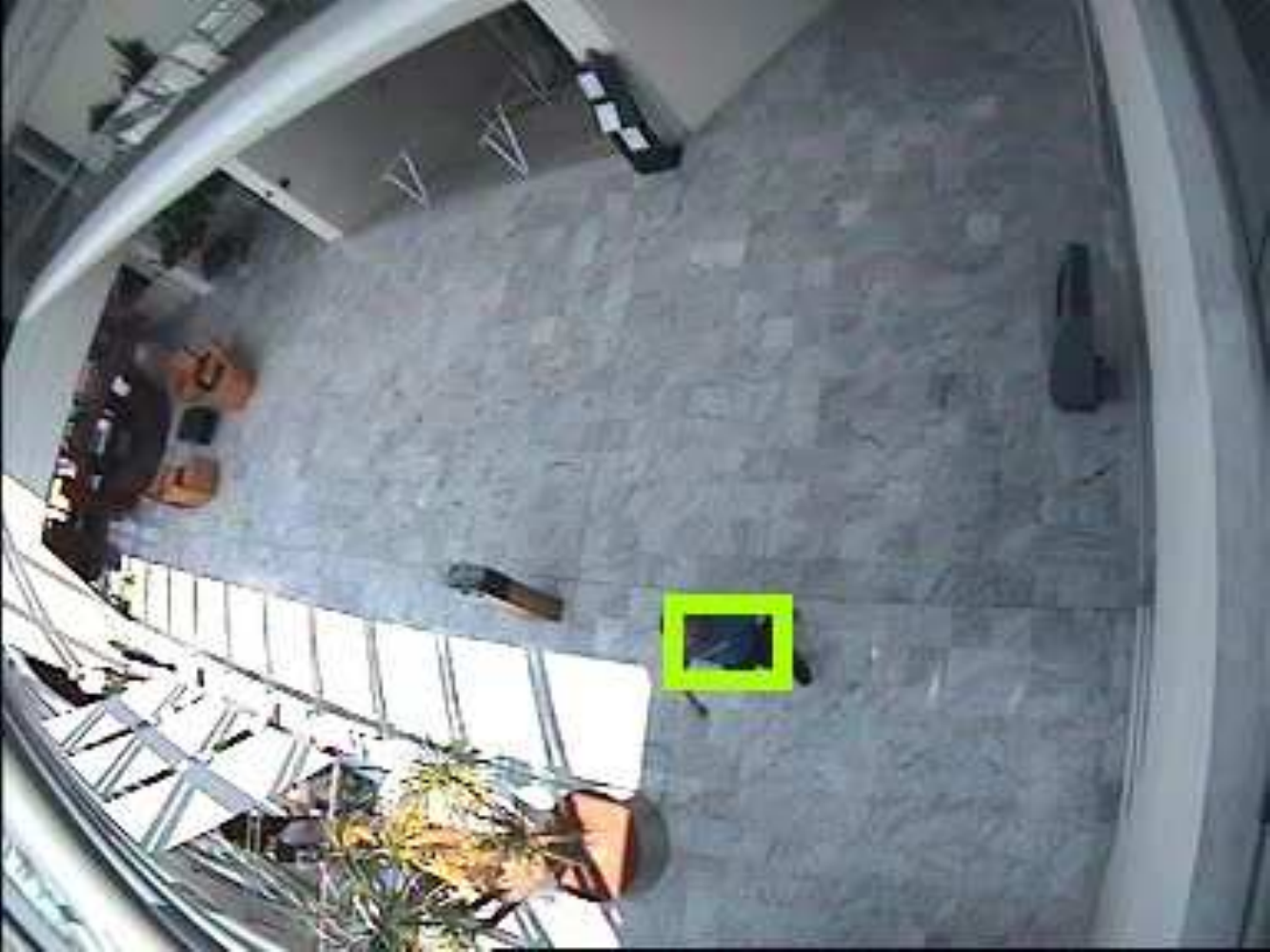}
\includegraphics[width=0.16\textwidth]{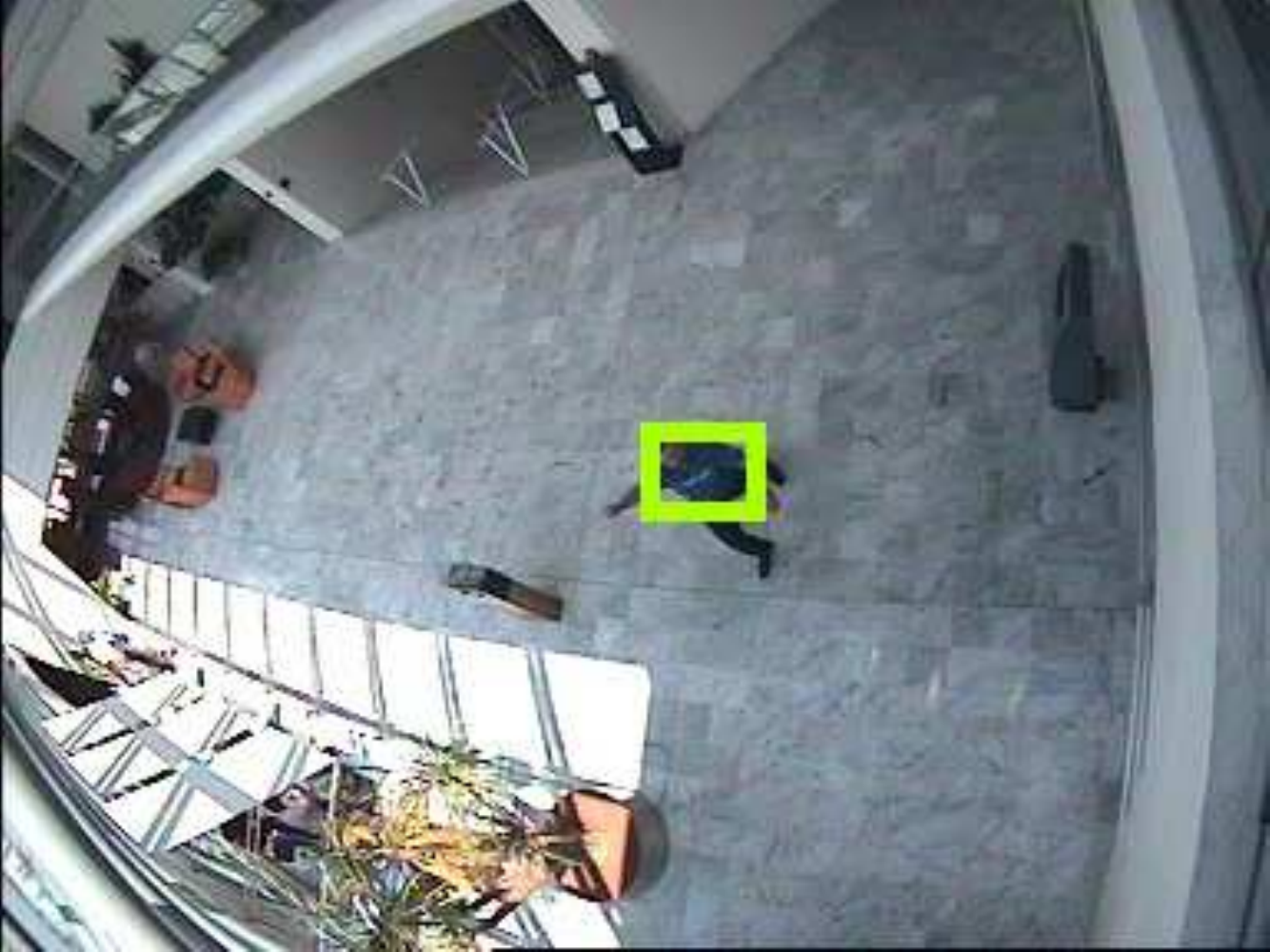}
\includegraphics[width=0.16\textwidth]{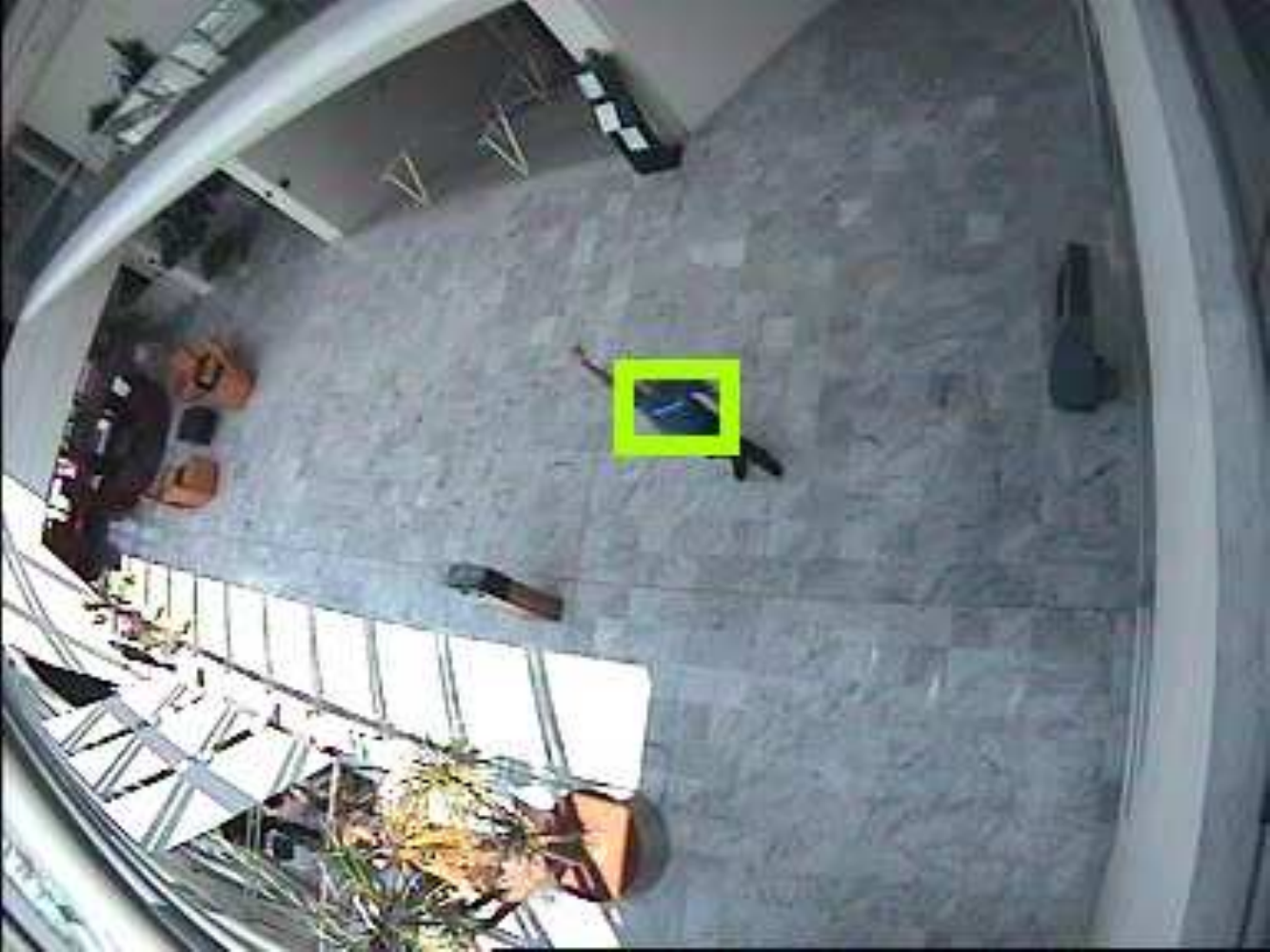}
\includegraphics[width=0.16\textwidth]{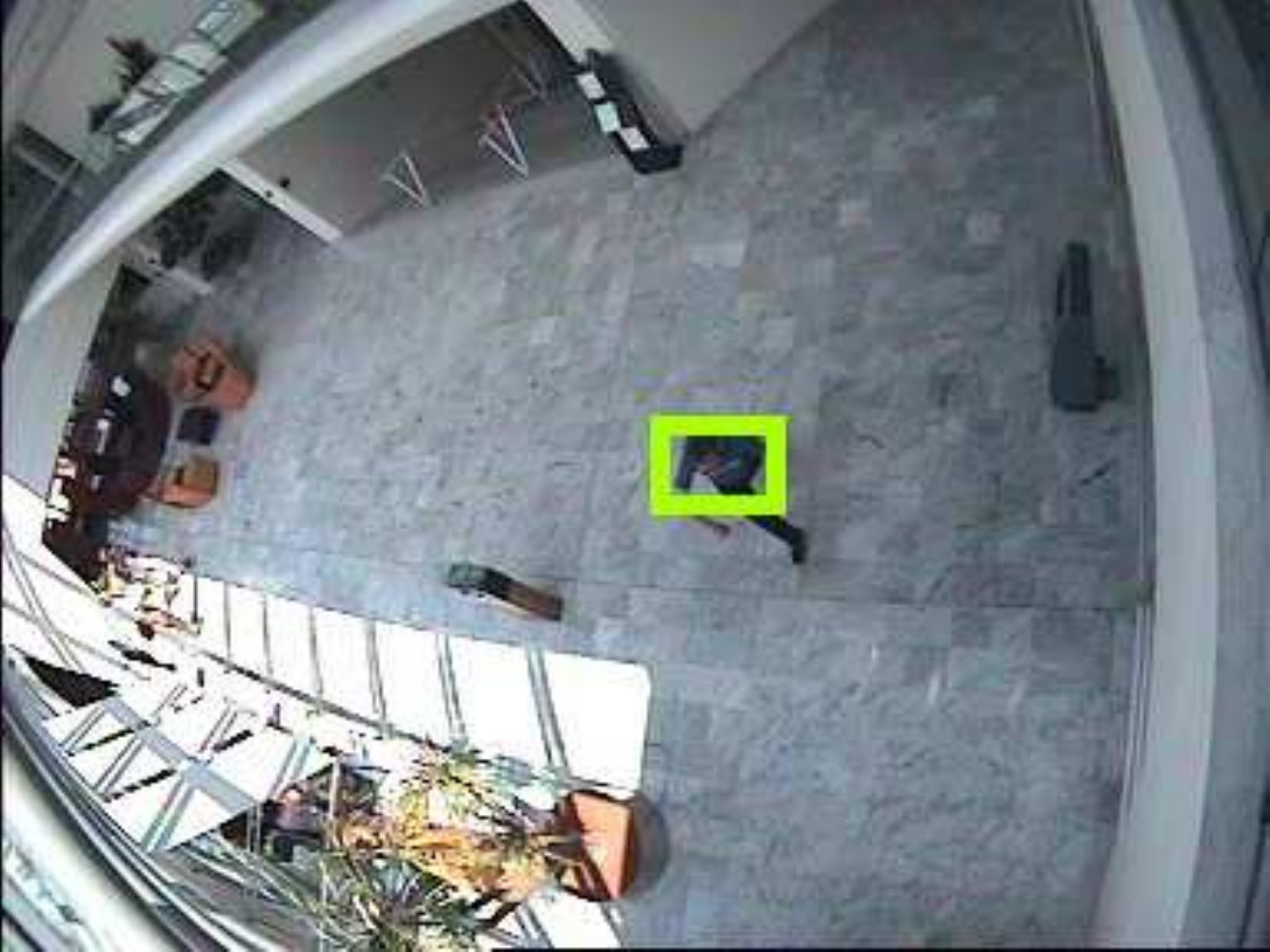}
\includegraphics[width=0.16\textwidth]{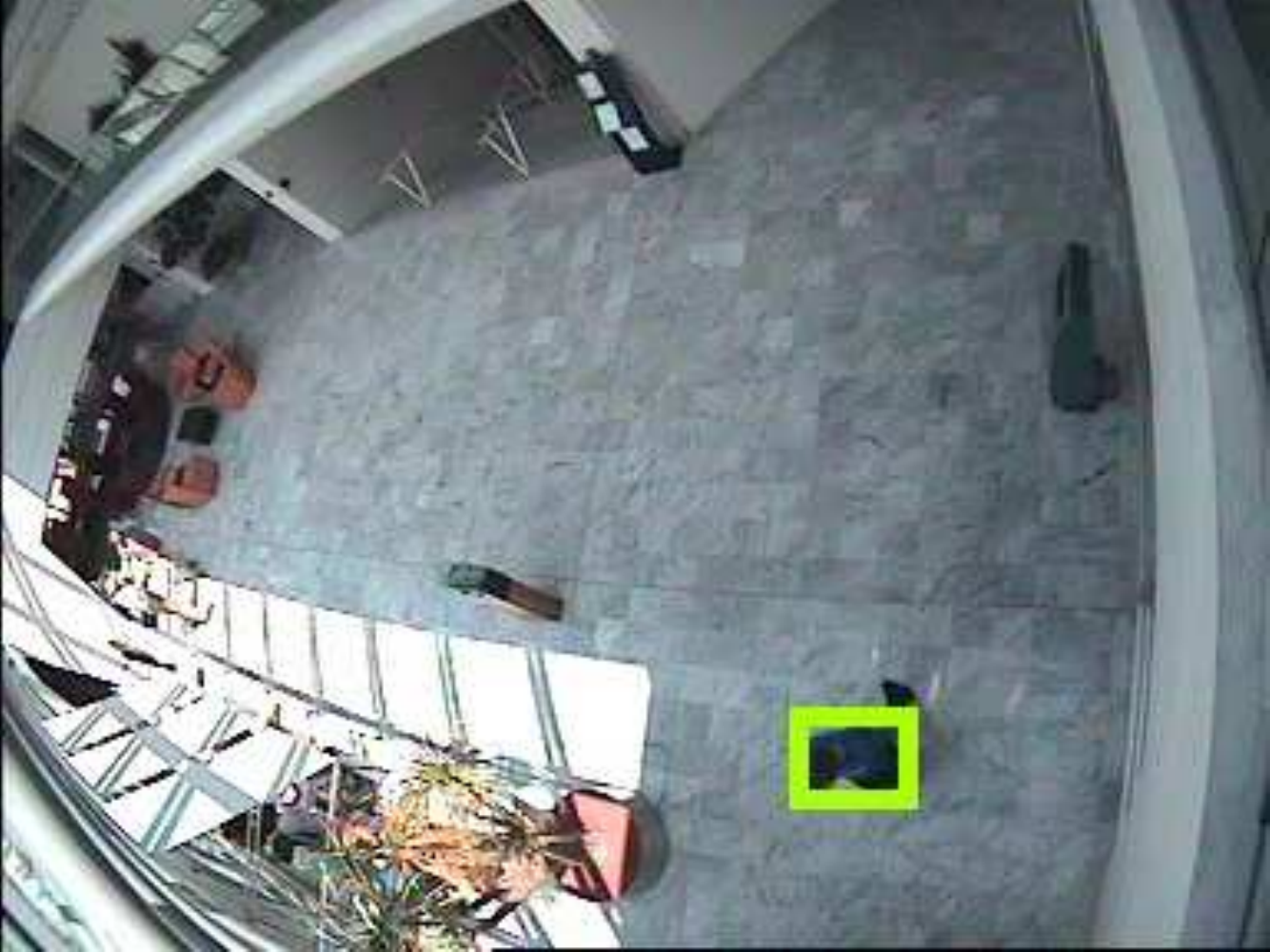}
\includegraphics[width=0.16\textwidth]{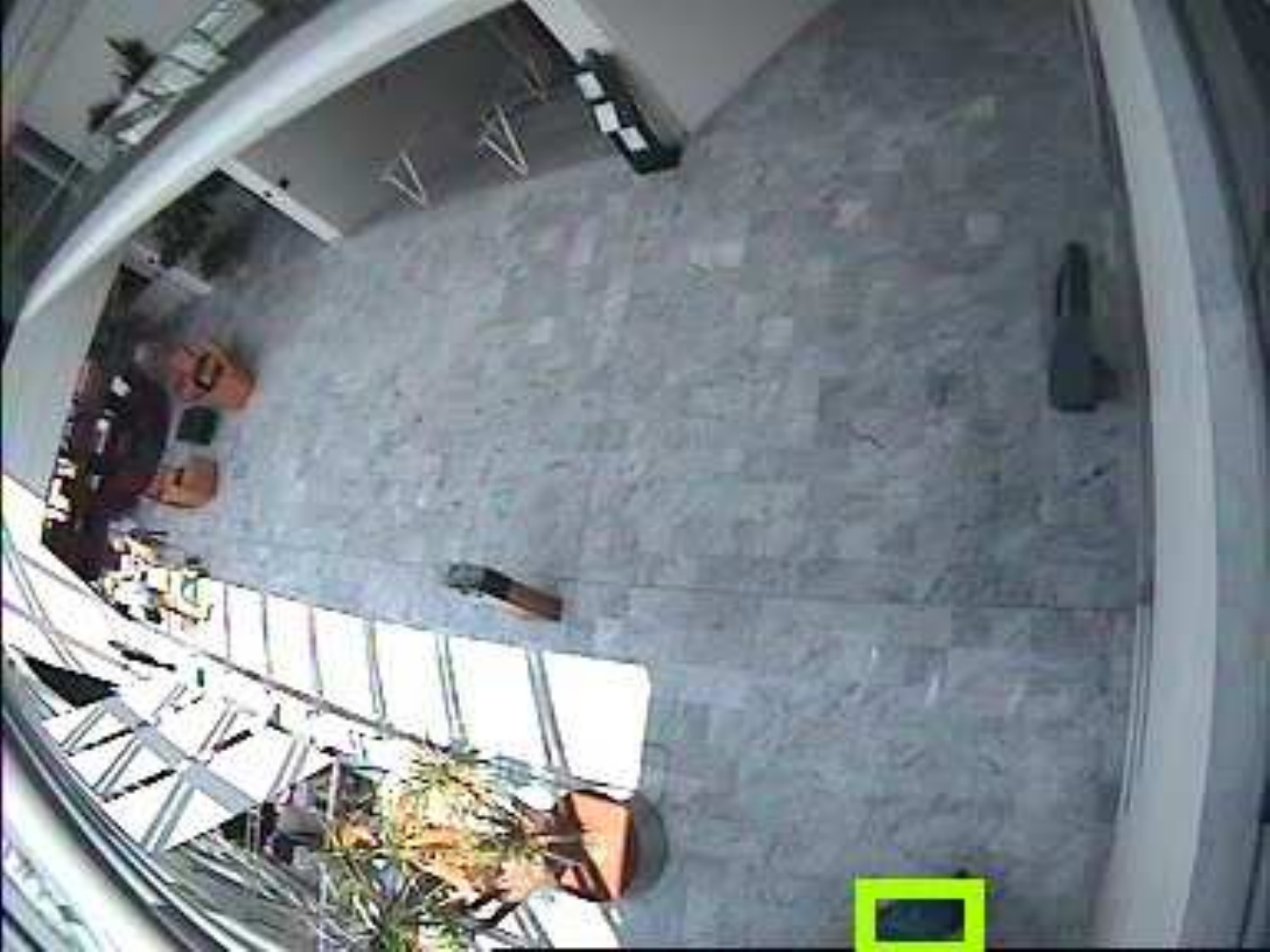}
\caption{ {\tt Walker} sequence 1. Tracking results of the proposed generalized
  kernel tracker.
  Frames 20, 40, 60, 90, 115, 130 are shown.
  The video size is $ 384 \times 288 $ and the frame rate is 
  $ 25$ FPS.
  }
  \label{fig:tracking3}
\end{figure*}


\begin{figure*}[]
\centering
\includegraphics[width=0.16\textwidth]{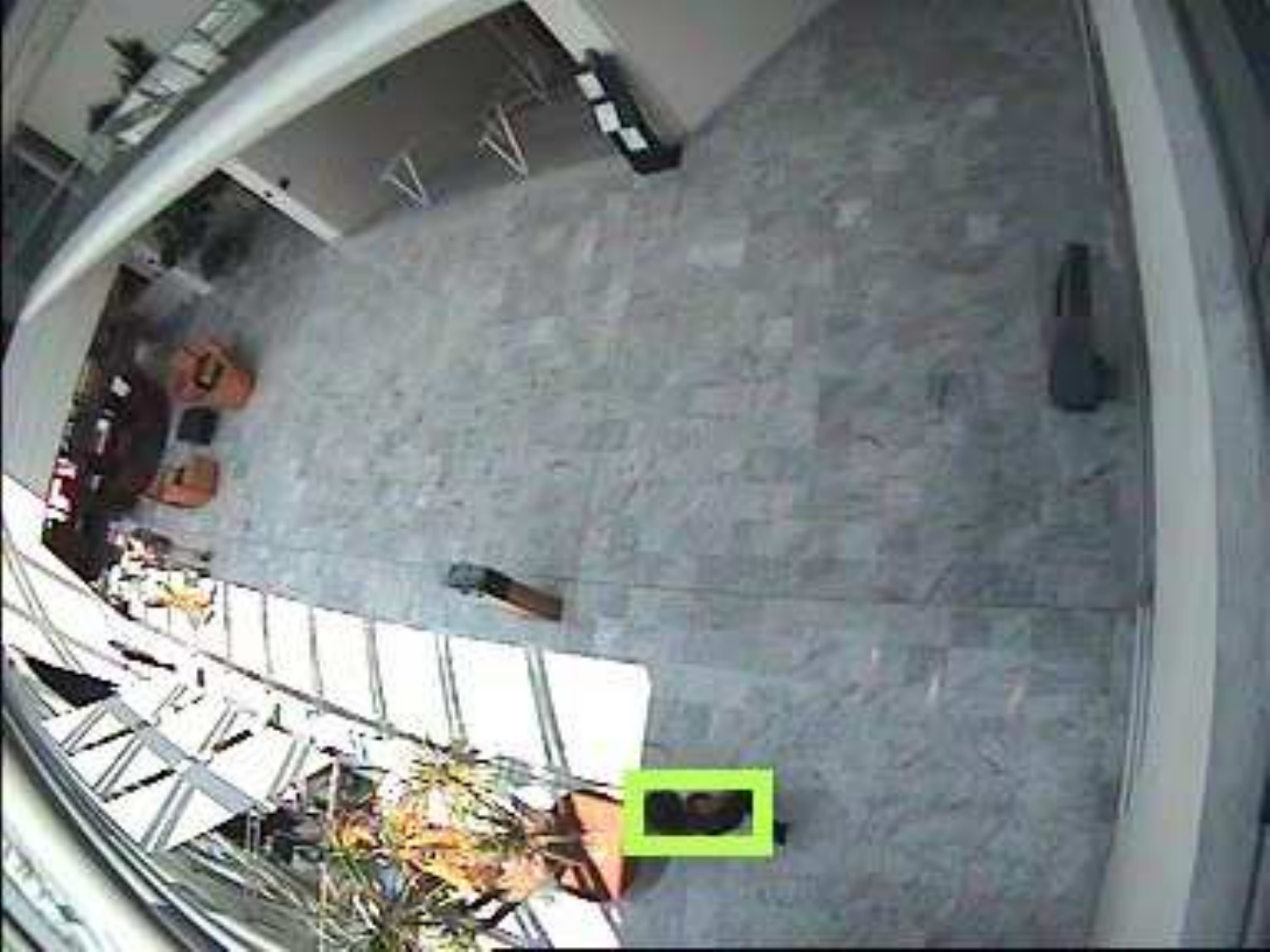}
\includegraphics[width=0.16\textwidth]{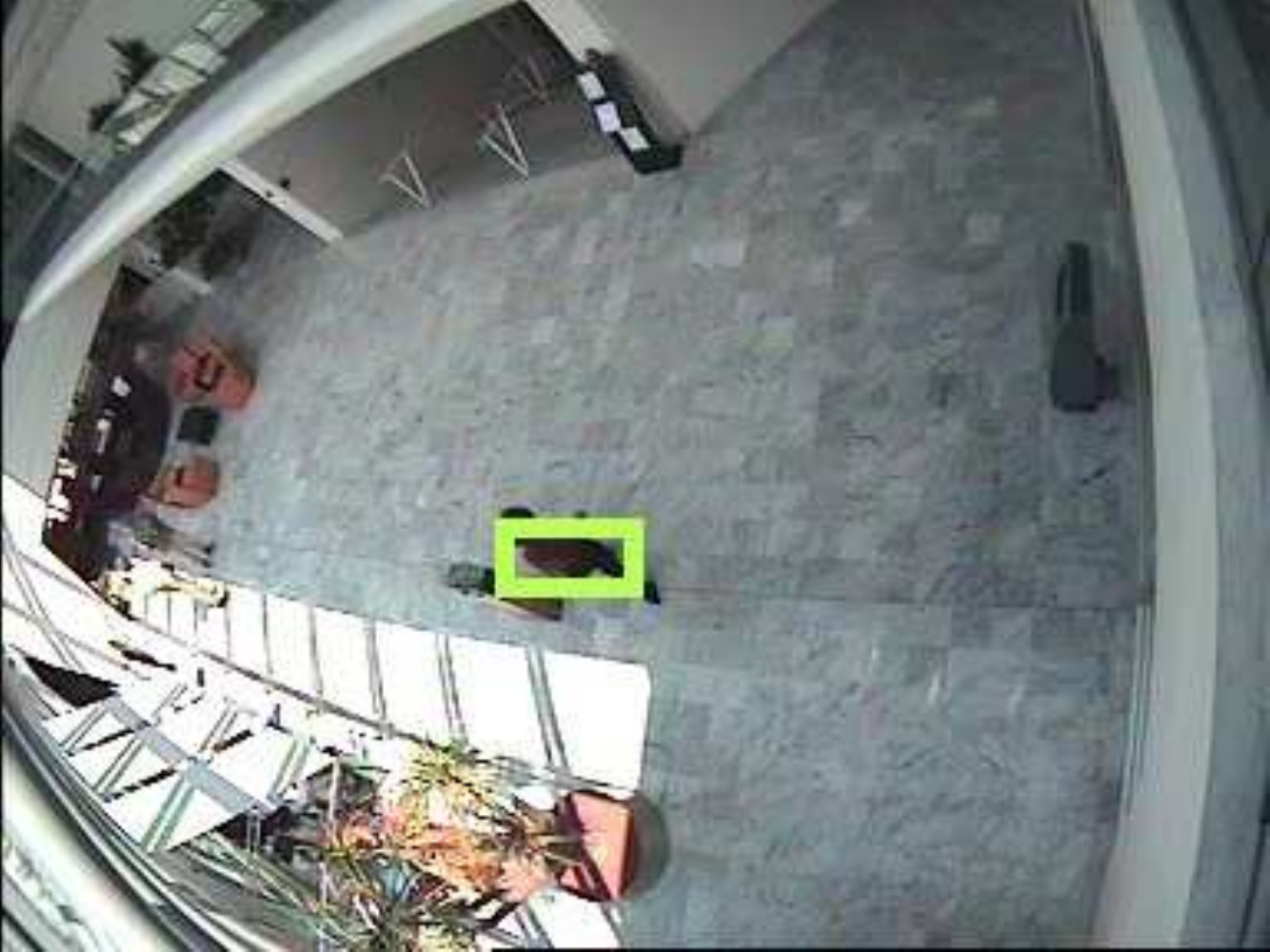}
\includegraphics[width=0.16\textwidth]{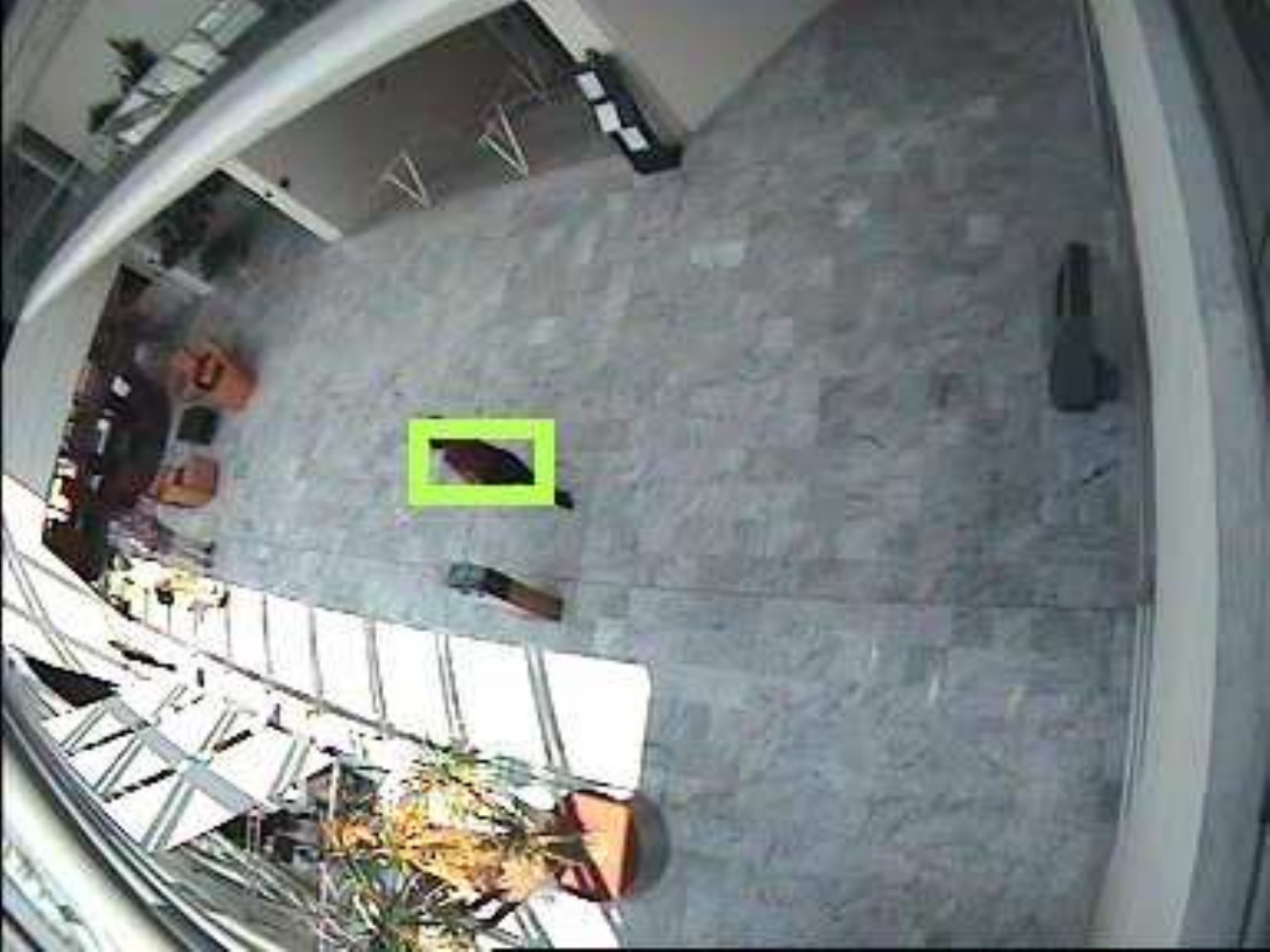}
\includegraphics[width=0.16\textwidth]{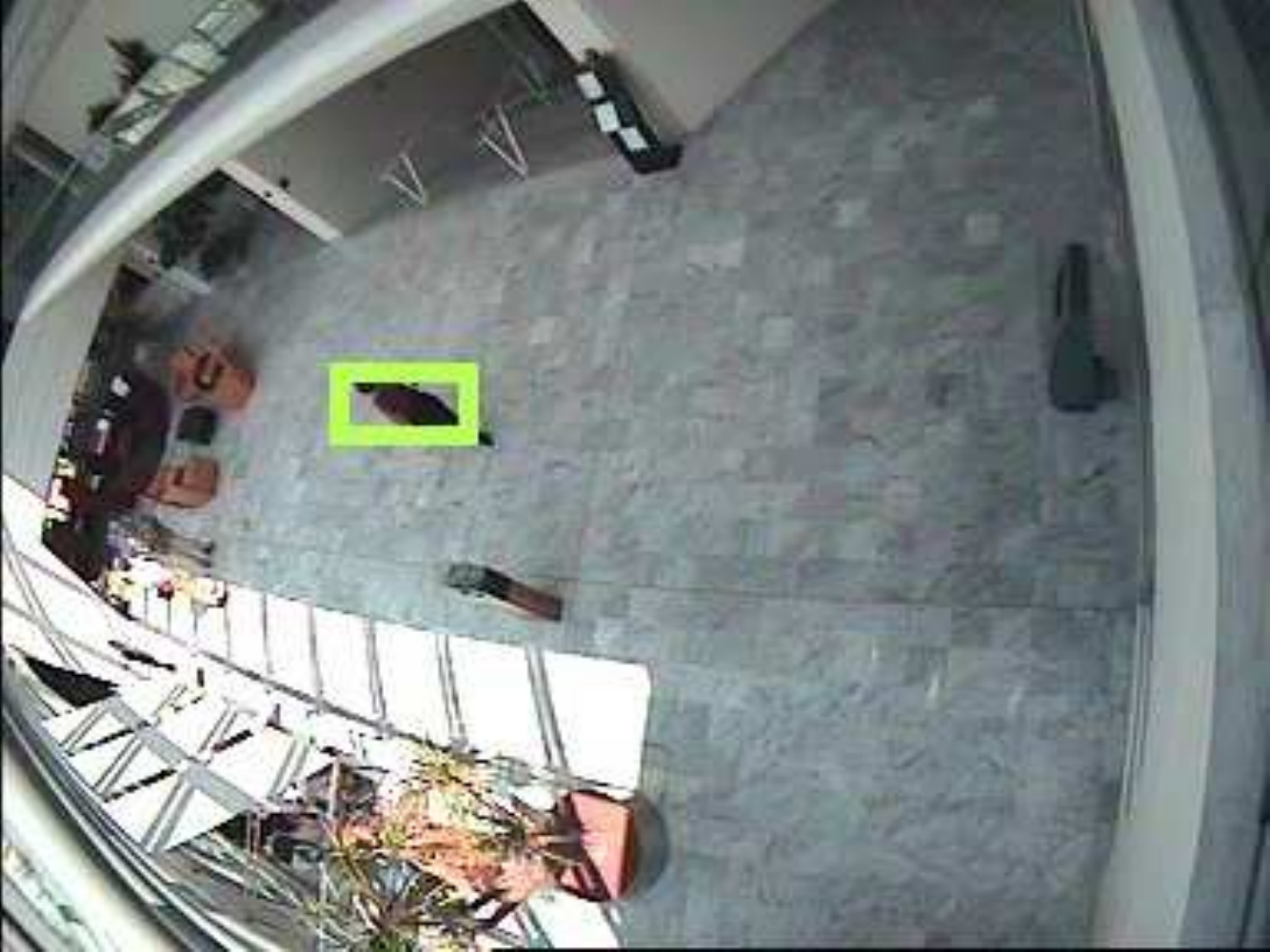}
\includegraphics[width=0.16\textwidth]{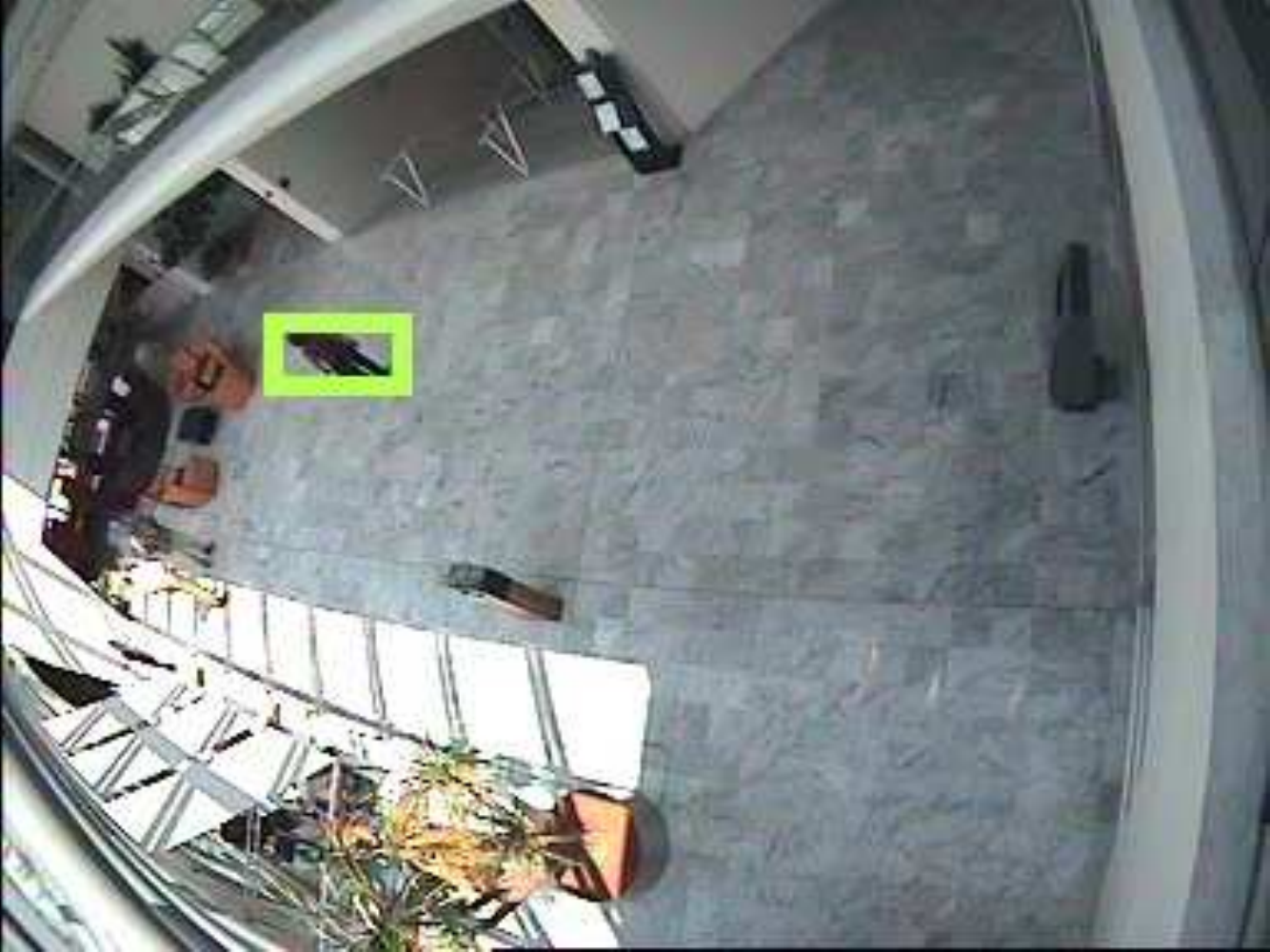}
\includegraphics[width=0.16\textwidth]{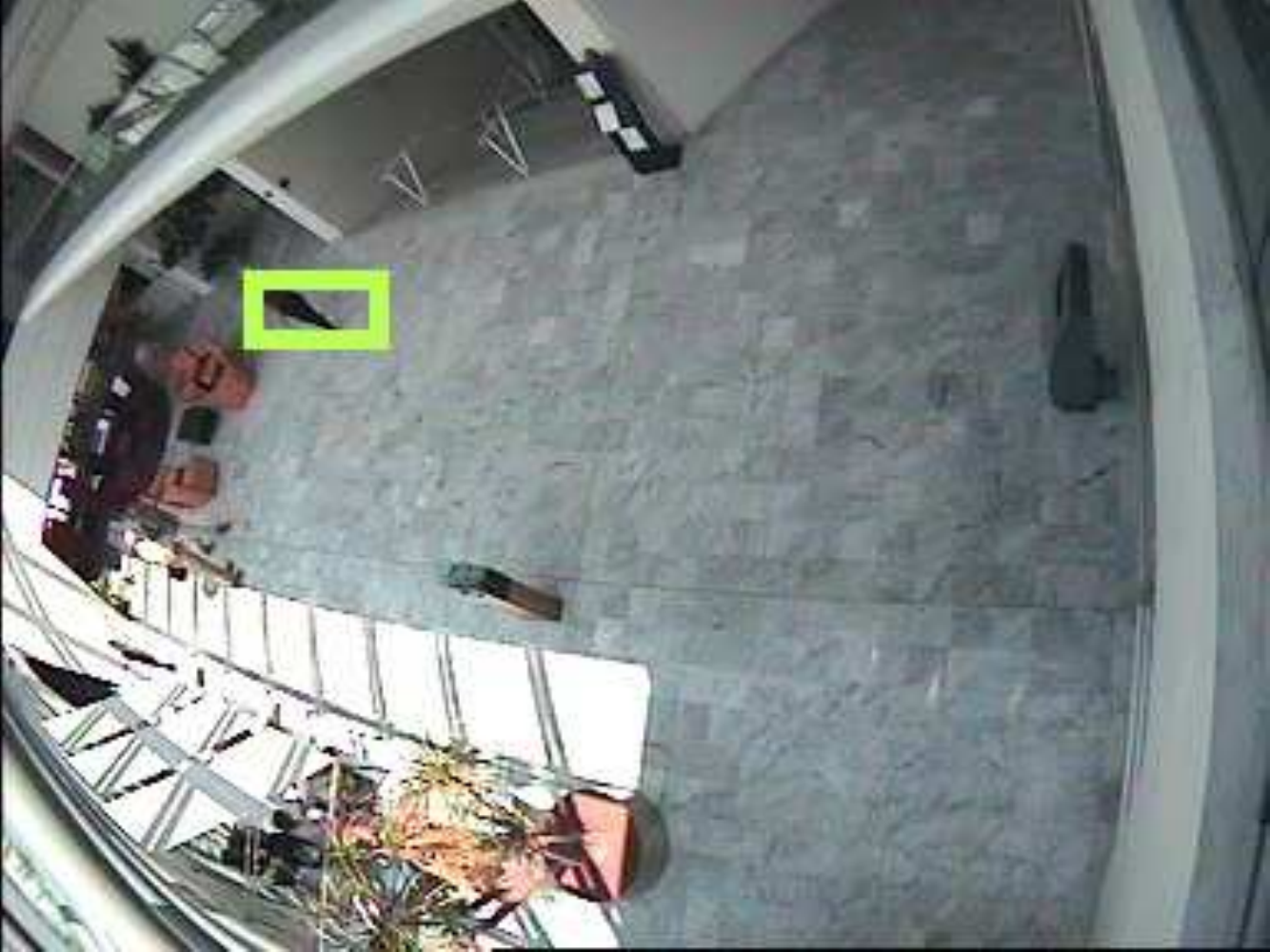}
\caption{{\tt Walker} sequence 2. Tracking results of the proposed generalized
  kernel tracker.
  Frames 10, 55, 80, 105, 140, 183 are shown.
  The video size and frame rate are same as Fig.~\ref{fig:tracking3}.
  }
  \label{fig:tracking4}
\end{figure*}


\begin{figure*}[]
\centering
\includegraphics[width=0.16\textwidth]{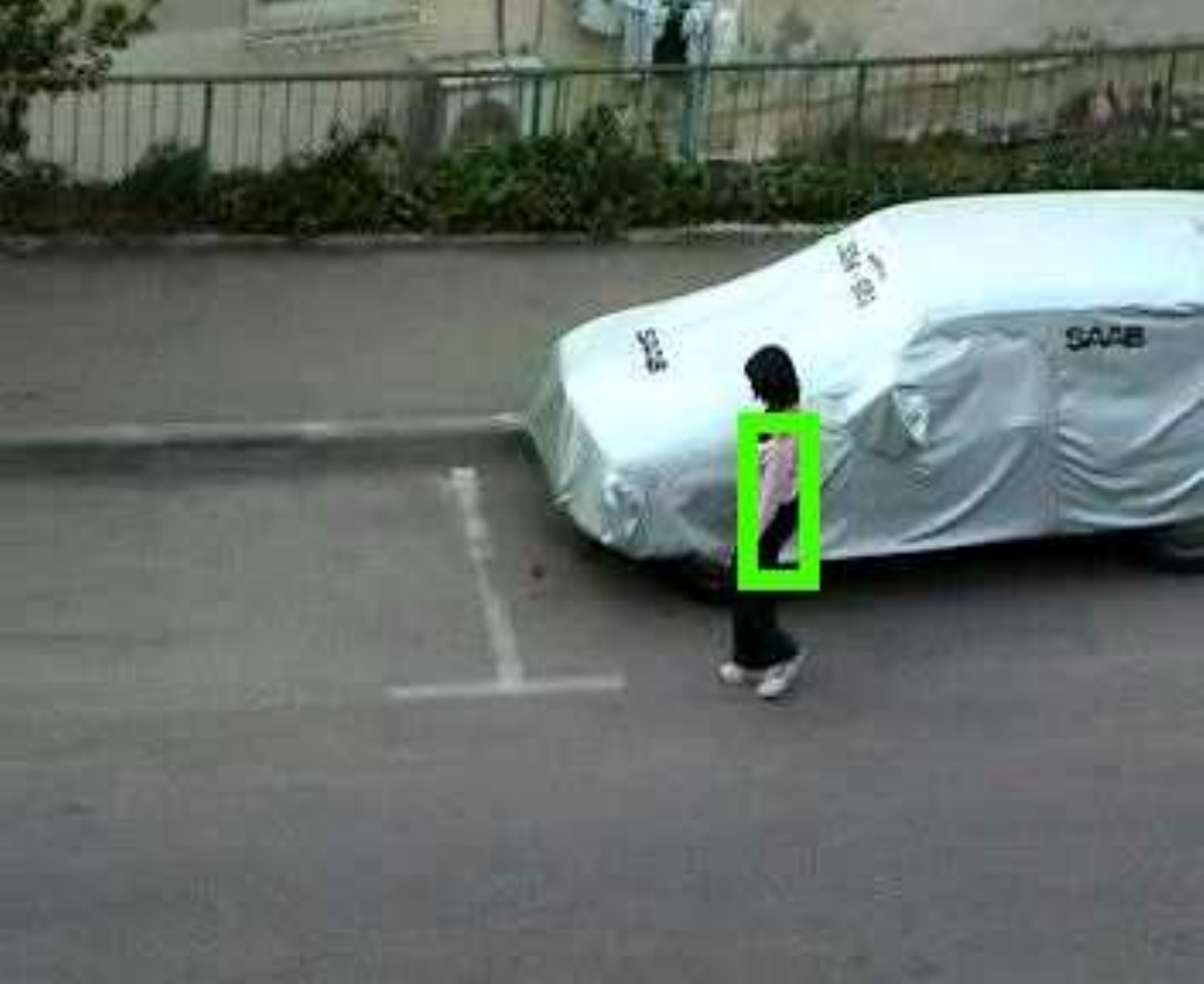}
\includegraphics[width=0.16\textwidth]{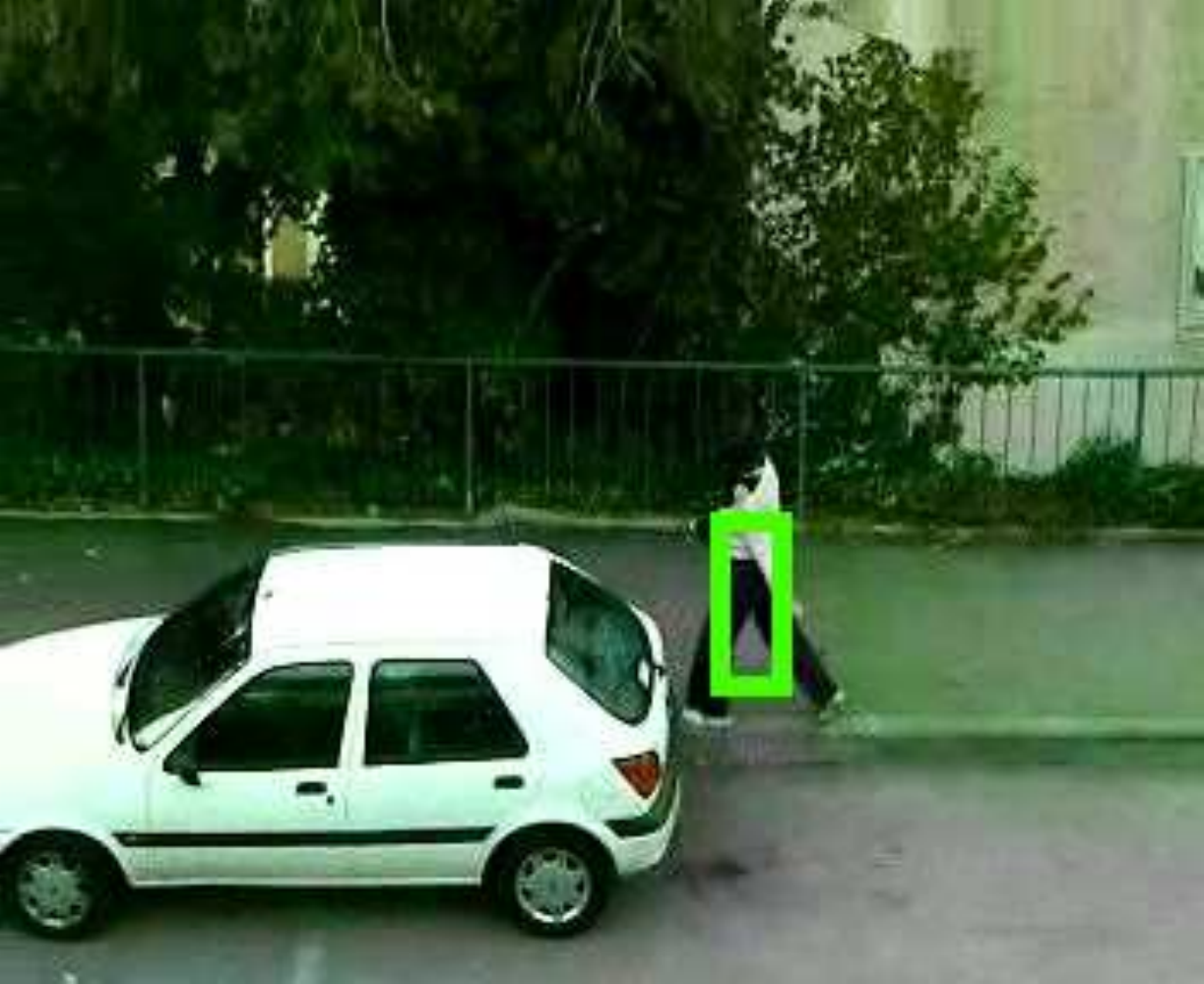}
\includegraphics[width=0.16\textwidth]{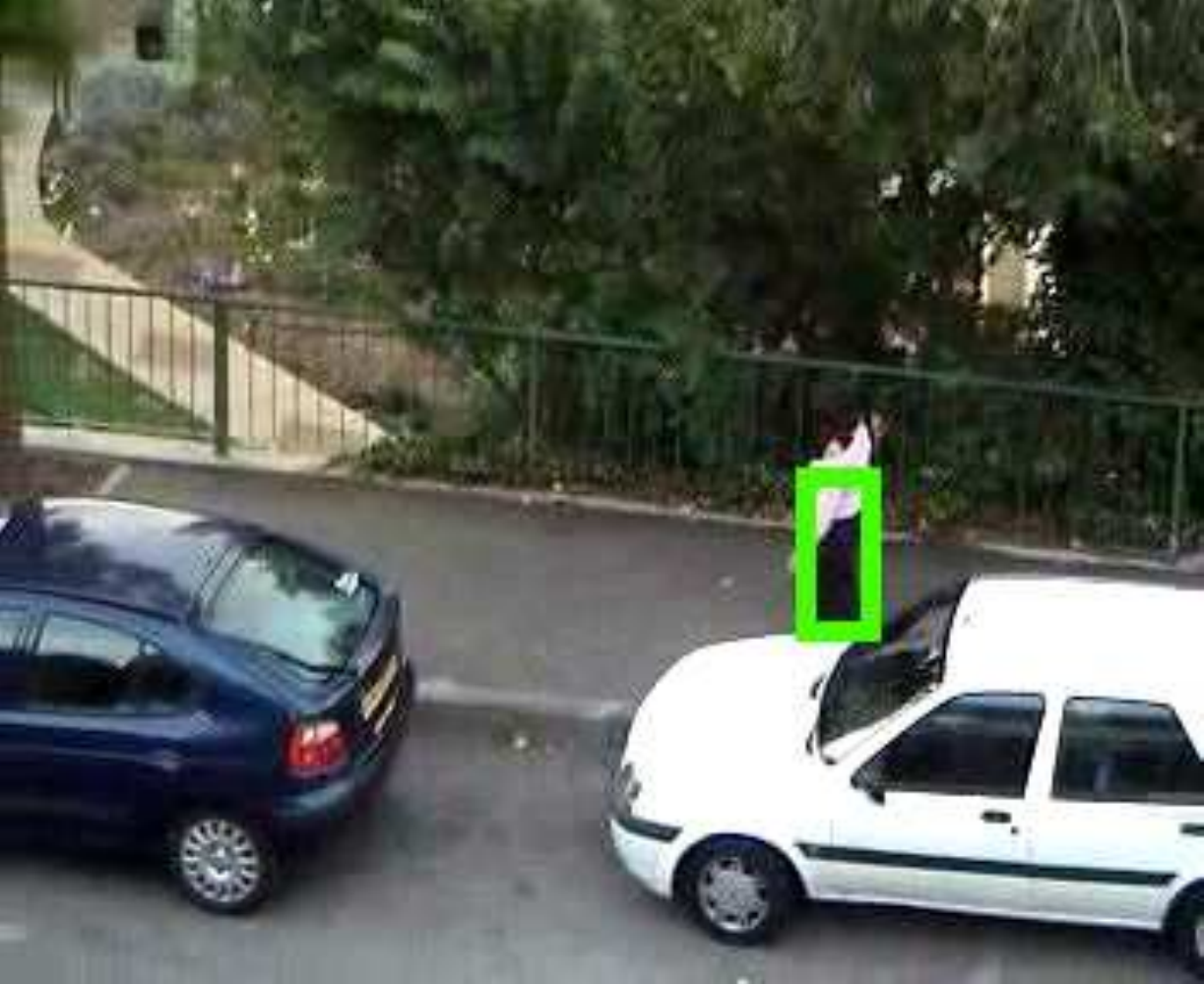}
\includegraphics[width=0.16\textwidth]{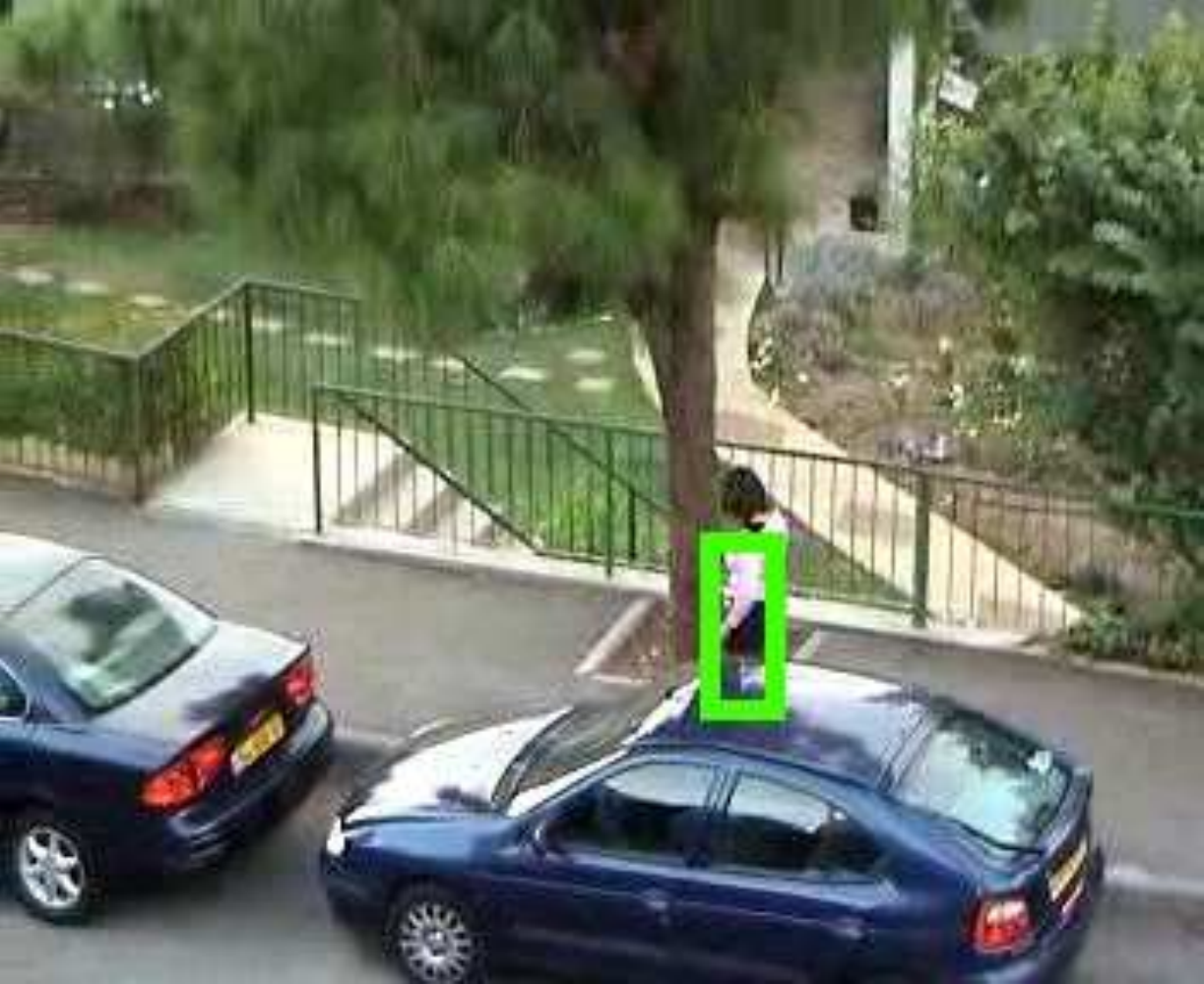}
\includegraphics[width=0.16\textwidth]{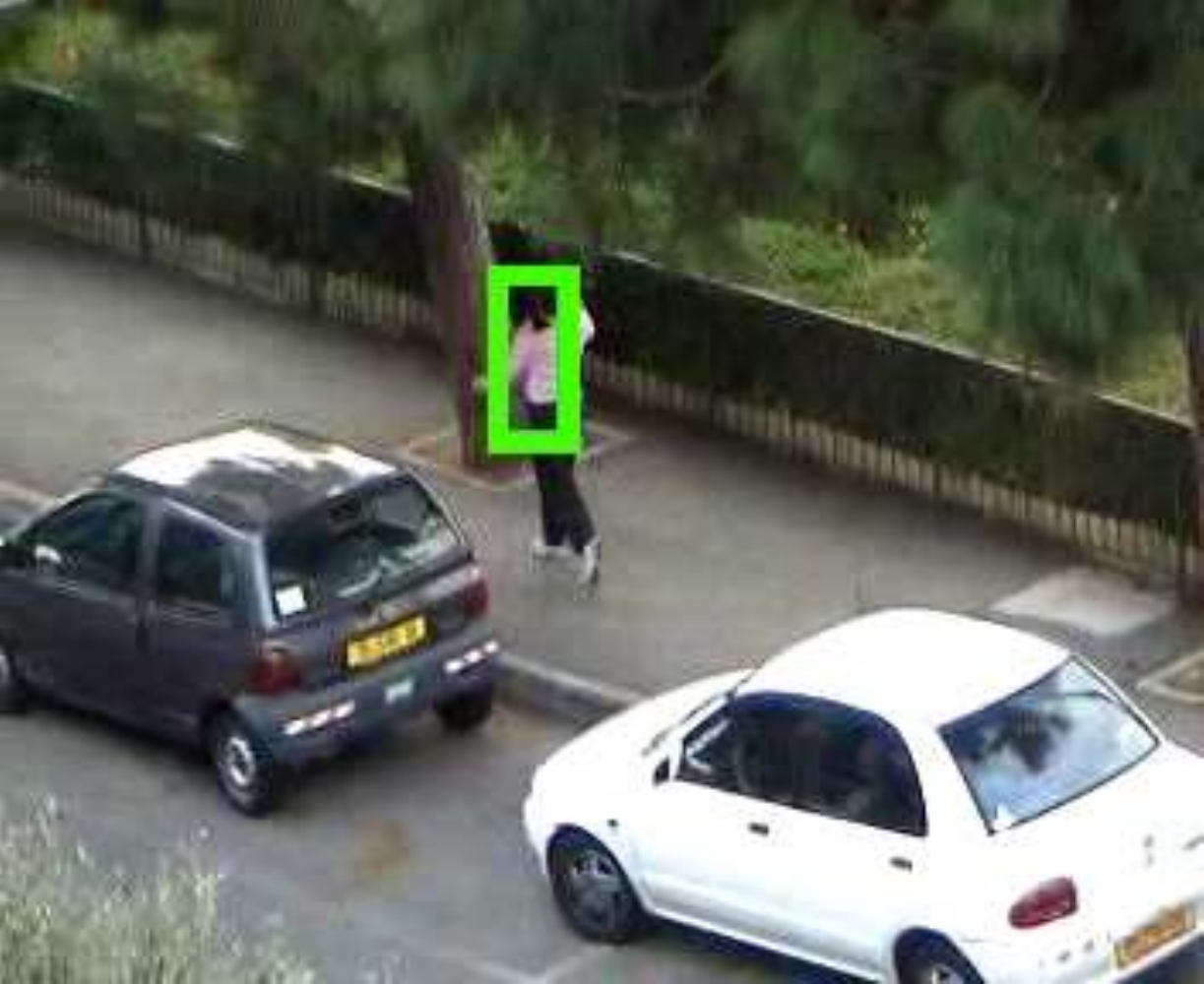}
\includegraphics[width=0.16\textwidth]{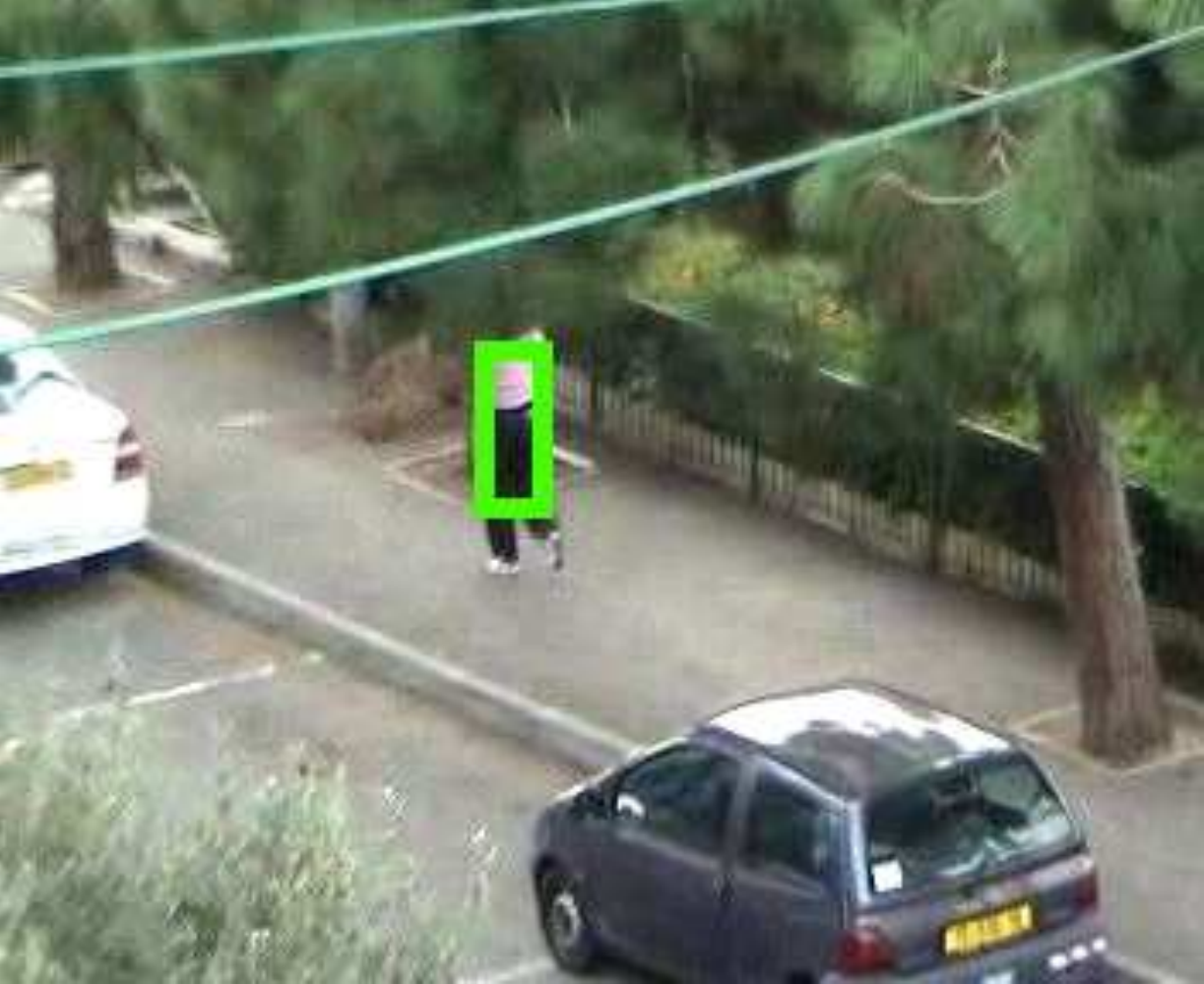}
\caption{{\tt Walker} sequence 3. Tracking results of the proposed generalized
  kernel tracker.
  Frames 20, 98, 152, 220, 444, 553 are shown. 
  The video is of size $352 \times 288$ and frame rate $ 30 $ FPS.
  }
  \label{fig:tracking5}
\end{figure*}

	In the first experiment, the tracked person moves quickly.
	Hence the displacement between neighboring frames is large.
    The illumination also changes. The background scene is cluttered 
    and contains materials with similar color as the target.
    The proposed algorithm tracks the whole
	sequence successfully. Fig.~\ref{fig:tracking1} summarizes
	the tracking results. The standard MS tracker fails at frame 
    \#57; recovers at frame \#74 and then fails again. The particle
    filter also loses the target due to motion blur and fast
    movement. Our on-line adaptive tracker achieves the most accurate
    results.
	
    Fig.~\ref{fig:tracking2} shows that the results on a more challenging
    video. The target turns around and 
    at some frames it even moves out
    of the view. At frame \#194, the target disappears. Generalized
    kernel tracker and particle filter recovers at the following
    frames while the MS tracker fails. Again we can see the proposed
    tracker performs best due to its learned template model and
    on-line adaptivity. 
    When the head turns around, all trackers can lock the target
    because compared with the background, the hair color is more
    similar to the face color.
    These two experiments show the proposed tracker's robustness to
    motion blur, large pose change and target's fast movement over
    the standard MS tracker and particle filter based tracker.
    In the experiments, to initialize the proposed tracker,
    we randomly
    pick up a few negative samples from the background. 
    We have found this simple treatment works well.

   We present more samples from three more sequences in 
   Figs.~\ref{fig:tracking3}, \ref{fig:tracking4} and
   \ref{fig:tracking5}. 
   We mark only our tracker in these frames. 
   From Figs.~\ref{fig:tracking3} and \ref{fig:tracking4}
   we see that
   despite the target moving into shadow at some frames, our tracker 
   successfully tracks the target through the whole sequences.
%
%
%
%
%
%

   We have shown promising tracking results of the proposed tracker
   on several video clips. We now present some quantitative
   comparisons of our algorithm with other trackers.   

\begin{figure}[th]
\centering
\includegraphics[width=0.117\textwidth]{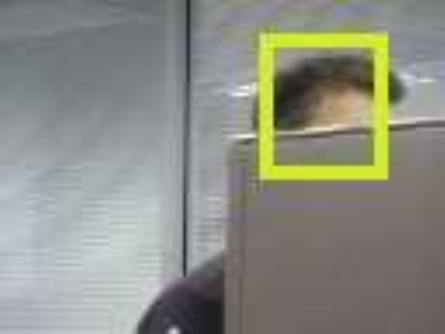}
\includegraphics[width=0.117\textwidth]{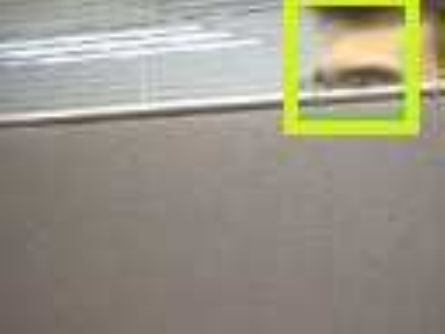}
\includegraphics[width=0.117\textwidth]{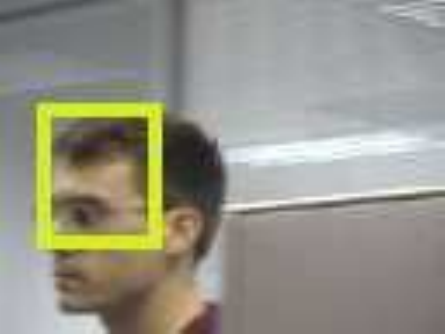}
\includegraphics[width=0.117\textwidth]{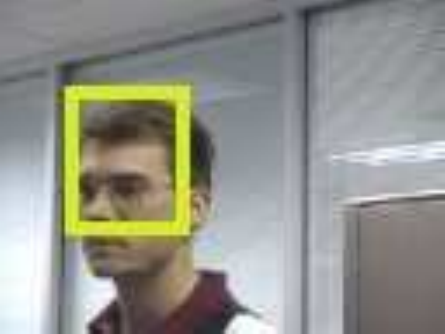}
\\
\includegraphics[width=0.117\textwidth]{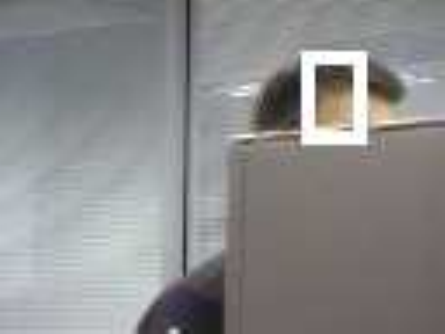}
\includegraphics[width=0.117\textwidth]{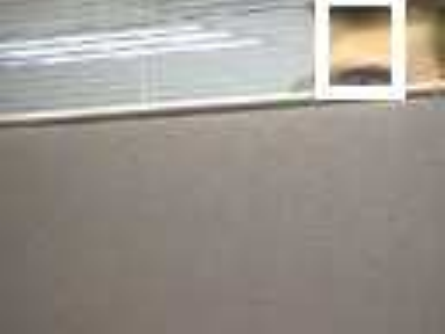}
\includegraphics[width=0.117\textwidth]{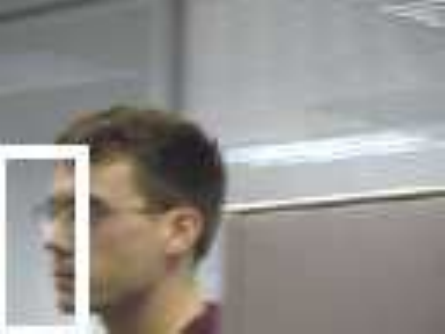}
\includegraphics[width=0.117\textwidth]{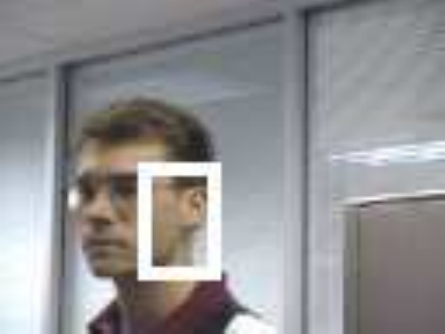}
\caption{ {\tt Cubicle} sequence 1. 
Tracking results of the proposed tracker (top) and particle filtering
(bottom). Frames 16, 30, 41, 45 are shown. 
The video is of size $352 \times 288$ and $ 30 $ FPS.
}
  \label{fig:cubicle1}
\end{figure}

   First, we run the proposed tracker, MS, and
   particle filter trackers on the {\tt cubicle} sequence 1.
   In Fig.~\ref{fig:cubicle1}, we show some tracking frames of our
   method and particle filtering.
   Compared with particle filtering, ours are much better in terms of
   accuracy and much faster in terms of the tracking speed. 
   Our results are also {\em slightly} 
   better than the standard MS tracker. But visually there is no
   significant difference, so we have not included MS results in
   Fig.~\ref{fig:cubicle1}.

\begin{figure}[th]
\centering
\includegraphics[width=0.4\textwidth]{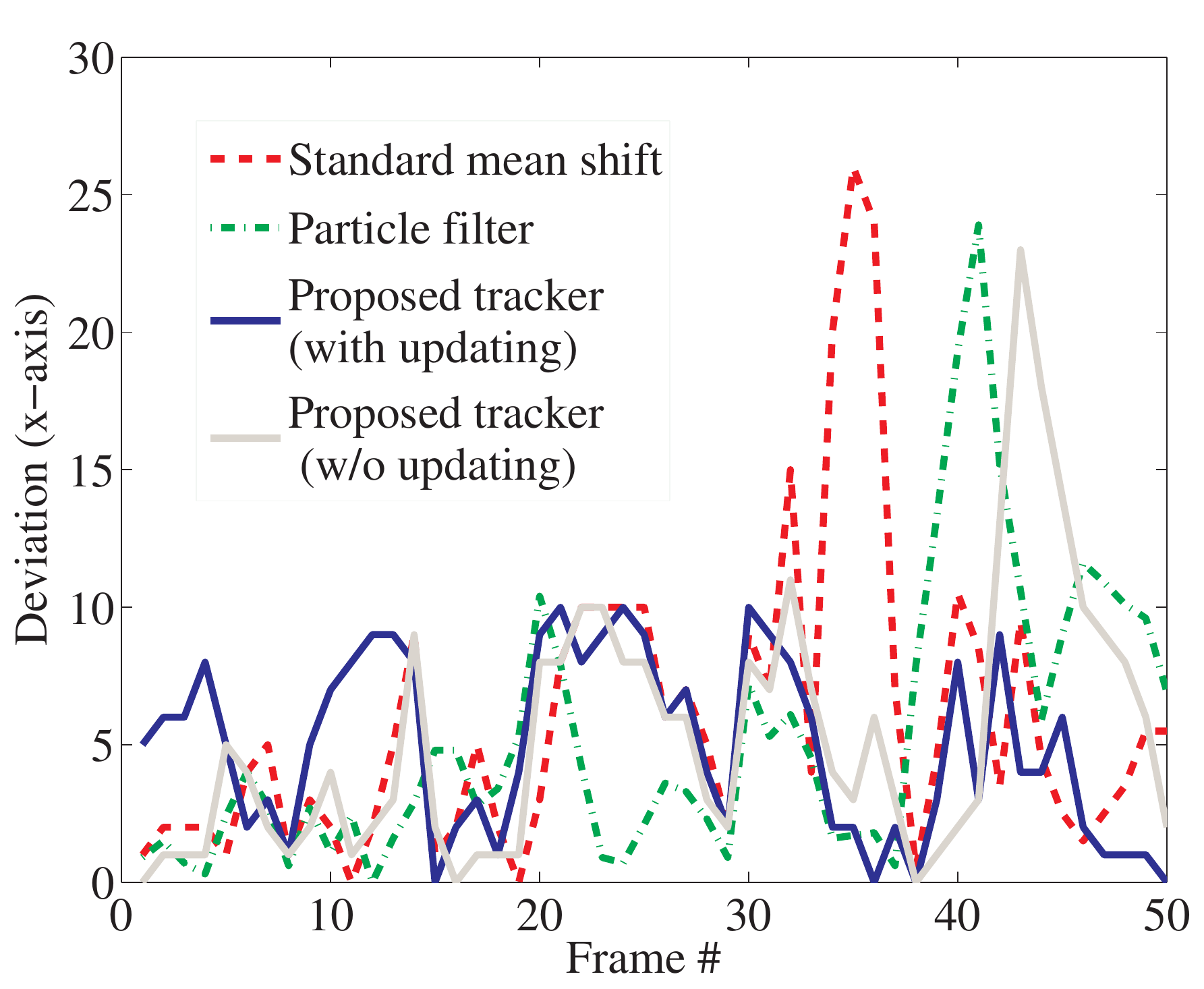}
\includegraphics[width=0.4\textwidth]{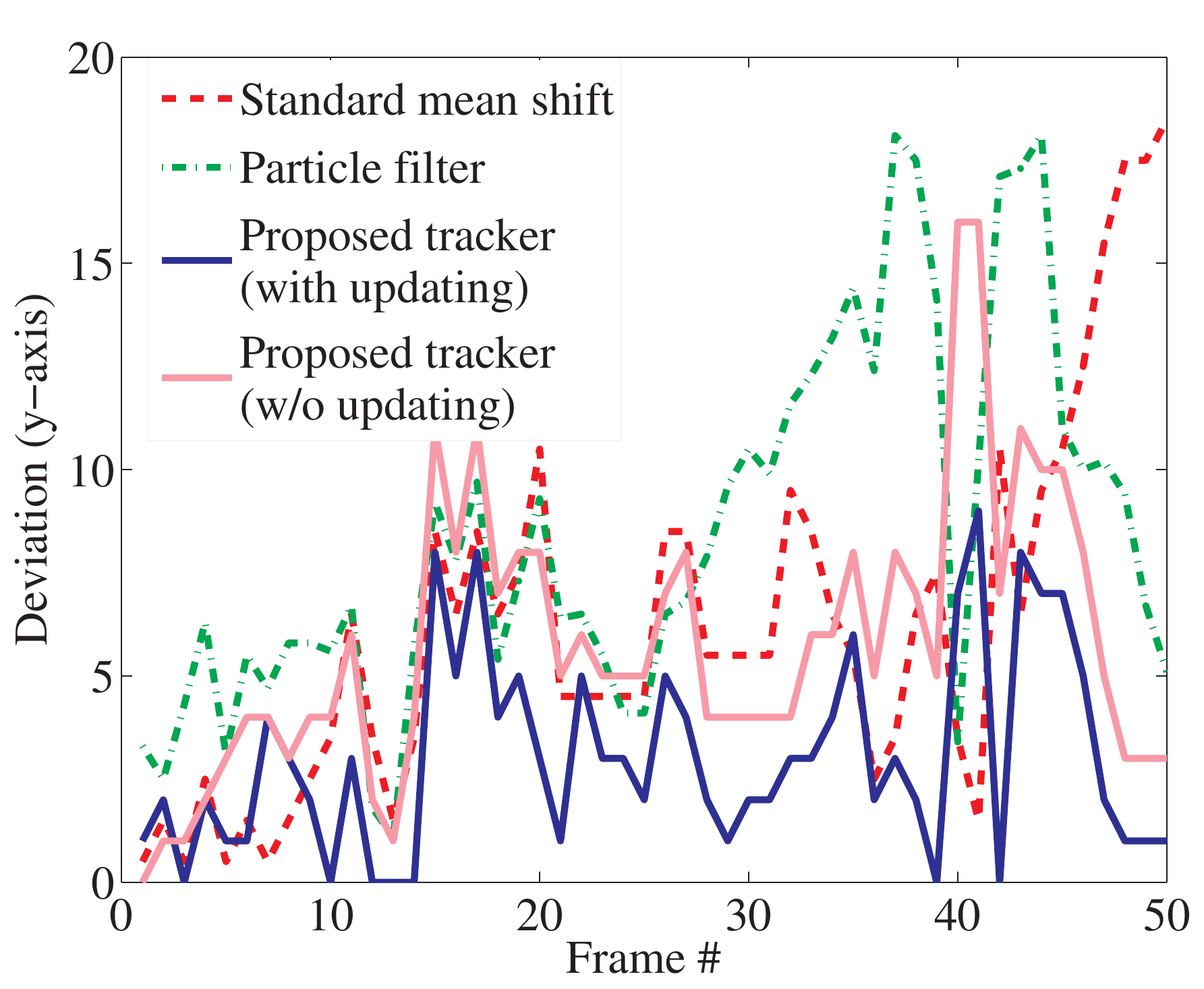}
  \caption{The $ \ell_1$-norm absolute error (pixels) 
  of the object's center against
  the ground truth on the {\tt cubicle} sequence 1. 
  The two figures correspond to $x$-, and $y$-axis, respectively. 
  The proposed tracker with on-line updating gives the best result. 
  As expected, the proposed tracker without updating shows a similar
  performance with the standard MS tracker. 
  }
  \label{fig:qant1}
\end{figure}

   Again, the particle filter tracker uses $1500$ particles.
   We have run the particle filter $ 5 $ times and the best result is
   reported. 
   Fig.~\ref{fig:qant1} shows the absolute deviation of the tracked
   object's center at each frame. Clearly the generalized kernel
   tracker demonstrates the best result. We have reported the average
   tracking error (the Euclidean distance of the object's center
   against the ground truth) in Table~\ref{tab:quat1}, which shows the proposed
   tracker outperforms MS and particle filter. 
   In Table~\ref{tab:quat1}, the error variance estimates are
   calculated from the tracking results of all frames regardless 
   the target is lost or not. 
   We have also proved the importance of on-line SVM update.
   As mentioned, when we
   switch off the on-line update, our proposed tracker would behave
   similarly to the standard MS tracker. We see from Table~\ref{tab:quat1}
   that even without updating,  the generalized kernel tracker is
   slightly better than the standard MS tracker. This might be because
   the initialization schemes are different: the generalized kernel
   tracker can take multiple positive as well as negative training examples
   to {\em learn} an appearance model, 
   while MS can only take a single image for
   initialization. Although we only use very few training examples
   (less than $ 10 $), it is already better than the standard MS
   tracker. 
   In this sequence, when the target object is occluded, the particle
   filter tracker only tracks the visible region such that the
   deviation becomes large. Our approach updates the learned
   appearance model using on-line SVM.  The region that partially
   contains the occlusion is added to the object class database
   gradually based on the  on-line update procedure. This way our tracker
   tracks the object position close to the ground truth.

   We also report the tracking failure rate (FR) for this video,
   which is the percentage of the number of failure frames in 
   the total number of frames. 
   If the distance between the tracked center and the ground truth's center
   is larger than a threshold, we mark it a failure.
   We have defined the threshold as $ 0.20 $ or $ 0.25 $ of the diagonal 
   length of the ground truth's bounding box, which results in
   two criteria:  \FRa and \FRb respectively.  The former is more strict than
   the latter. As shown in Table~\ref{tab:quat1}, our tracker with on-line 
   update produces lowest tracking failures under either criterion.


   \begin{table}[h!]
        \caption{The average tracking error against the ground truth (pixels) 
        on the {\tt cubicle}
        sequence 1. The mean and standard deviation are reported.
        We also report the tracking failure rates.}
        \footnotesize
        \begin{center}
        \resizebox{0.49\textwidth}{!}
        {
        \begin{tabular}{l|c|c|c|c}
        \hline
         & MS & Particle filter &  Ours (w/o update)  & Ours (update) \\
        \hline\hline
        error  & $ 9.6 \pm 5.7 $ & $10.5 \pm 5.8$  &  $8.5 \pm 4.9$ &
        $6.5 \pm 2.8$
        \\
        \FRa & 44.0\% & 44.0\% & 28.0\% & 6.0\%
        \\
        \FRb  & 16.0\% & 34.0\% & 14.0\% & 0.0\%
        \\
        \hline
        \end{tabular}
        }
        \label{tab:quat1}
        \end{center}
   \end{table}

    We also compare the running time of trackers, which is an important
    issue for real-time tracking applications. Table~\ref{tab:time1}
    reports the results on two sequences.\footnote{All algorithms are
    implemented in ANSI C++. 
    We have made the codes available at {\url{
    http: // code. google. com / p /detect/}}.
    A desktop with
    Intel Core$^{\rm TM}$ Duo 2.4-GHz CPU and $ 2 $-G RAM
    is used for running all the experiments.} 
   The generalized kernel
   tracker (around $ 65 $ fps)
   is comparable to the standard MS tracker, and much faster
   than the particle filter. This coincides with the theoretical
   analysis: our generalized kernel tracker's computational complexity
   is independent of the number of support vectors, so in the test
   phrase, the complexity is almost same as the standard MS. 
   One may argue that  
   the on-line update procedure introduces some overhead. 
   But the generalized kernel tracker employs the L-BFGS optimization
   algorithm which is about twice faster than MS, as shown in
   \cite{Fast07Shen}. Therefore, overall, the generalized kernel
   tracker runs as fast as the MS tracker. 
   Because the particle filter is stochastic, we have run it $ 5 $
   times and the average and standard deviation are reported. For
   our tracker and MS, they are deterministic and the standard 
   deviation is negligible.  
   Note that the computational complexity if the particle filter
   tracker is linearly proportional to the number of particles.


    \begin{table}[h!]
         \caption{Running time per frame (seconds).
         The stochastic particle filter tracker has run $ 5 $ times and
         the standard deviation is also reported. 
         }
         \small
         \begin{center}
         \begin{tabular}{c|c|c|c}
         \hline
         Sequence & MS & Particle filter &  Ours \\
         \hline\hline
         {\tt cubicle} 1 & $ 0.0156 $ &  $ 0.352 \pm 0.025  $  & $ 0.0155 $  
         \\
         {\tt walker}  3  & $ 0.0169 $ &  $ 0.331 \pm 0.038 $  & $ 0.0142 $ 
         \\
         \hline
         \end{tabular}
         \label{tab:time1}
         \end{center}
    \end{table}

\begin{figure}[th]
\centering
\includegraphics[width=0.117\textwidth]{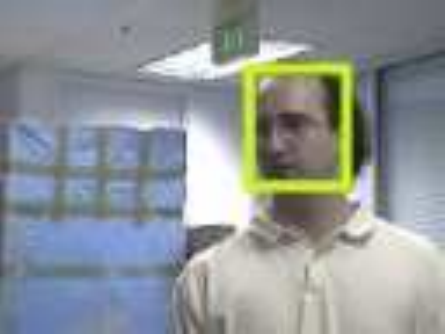}
\includegraphics[width=0.117\textwidth]{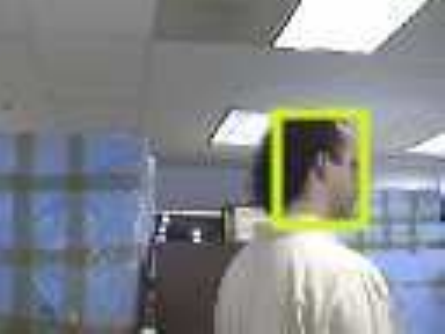}
\includegraphics[width=0.117\textwidth]{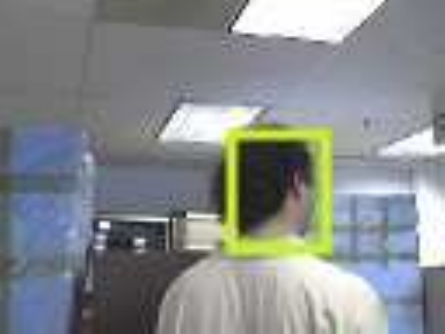}
\includegraphics[width=0.117\textwidth]{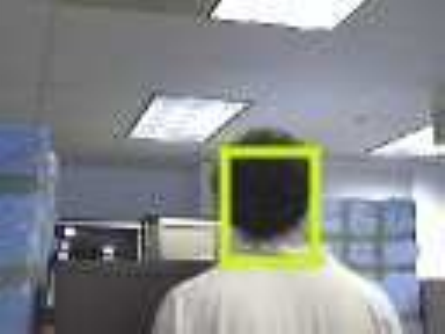}
\\
\includegraphics[width=0.117\textwidth]{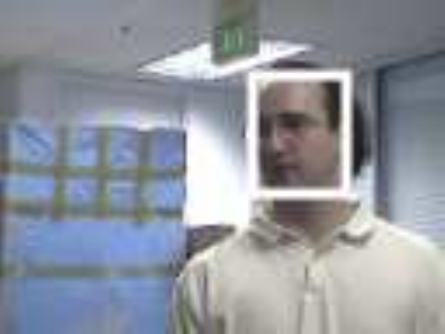}
\includegraphics[width=0.117\textwidth]{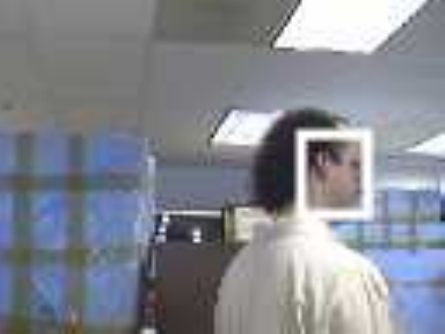}
\includegraphics[width=0.117\textwidth]{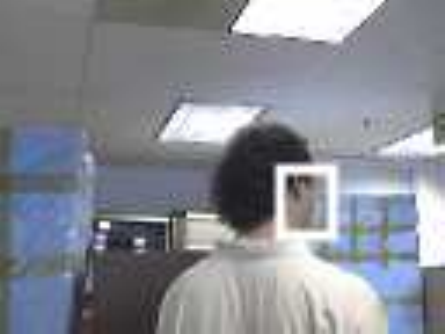}
\includegraphics[width=0.117\textwidth]{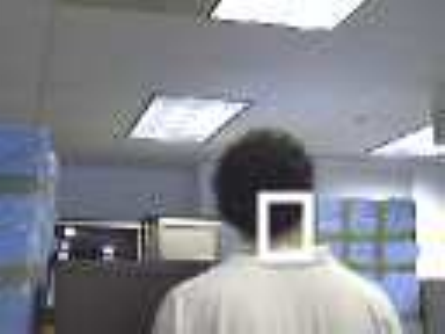}
\caption{ {\tt Cubicle} sequence 2. 
Tracking results of the proposed tracker (top) and particle filtering
(bottom). Frames 9, 55, 60, 64 are shown.  
The video is of size $352 \times 288$ and frame rate $ 30 $ FPS.
}
  \label{fig:cubicle2}
\end{figure}

    We have run another test on 
   {\tt cubicle}  sequence 2.
   We show some results of our method and particle filtering in
   Fig.~\ref{fig:cubicle2}. Although all the methods can track this
   sequence successfully, the proposed method achieves most accurate
   results.  We see that when the tracked object turns around, our
   algorithm is still able to track it accurately.  
   Table~\ref{tab:quat2} summarizes the quantitative performance. 
   Our method is also slightly better MS. Again we see that on-line update
   does indeed improve the accuracy. 
   We have also reported the tracking failure rates on this video.
   Our tracker with on-line update has the lowest tracking failures and 
   the one without on-line update is the second best. 
   These results are consistent with the previous experiments.

   \begin{table}[b!]
        \caption{The average tracking error against the ground truth (pixels) 
        on the {\tt cubicle}
        sequence 2. The mean and standard deviation are reported.
           We also report the tracking failure rates.
        }
        \footnotesize
        \begin{center}
              \resizebox{0.49\textwidth}{!}
        {
        \begin{tabular}{l|c|c|c|c}
        \hline
         & MS & Particle filter &  Ours (w/o update)  & Ours (update) \\
        \hline\hline
        error  & $ 5.7 \pm 3.5 $ & $8.4 \pm 3.4$  &  $5.5 \pm 3.2$ &
        $4.2 \pm 2.8$
        \\
         \FRa  & 4.6\% & 15.4\% & 3.1\% & 0.0\%
         \\
        \FRb & 0.0\% & 3.1\% & 0.0\% & 0.0\%
        \\
        \hline
        \end{tabular}
        }
        \label{tab:quat2}
        \end{center}
   \end{table}


\begin{figure}[th]
\centering
\includegraphics[width=0.4\textwidth]{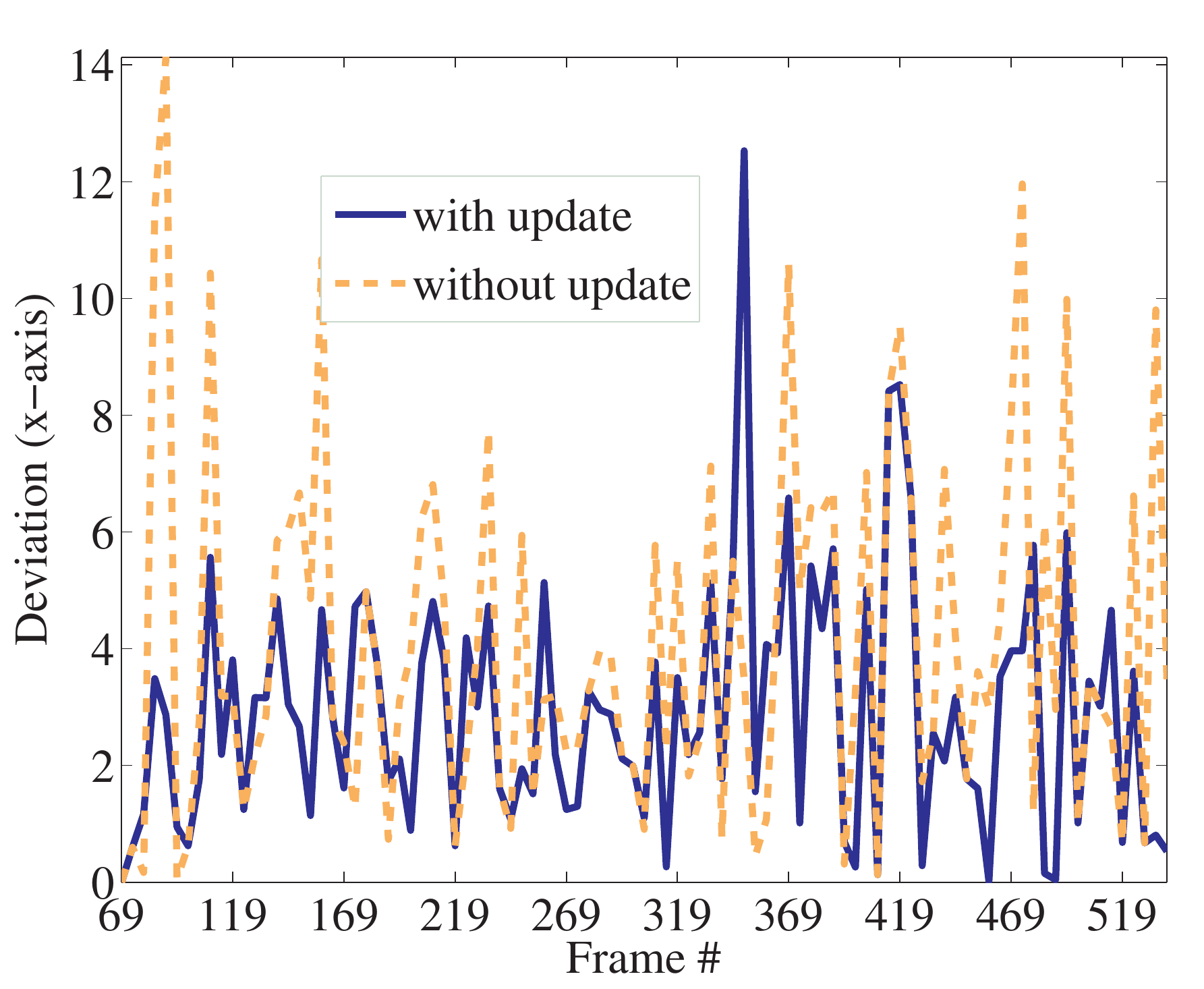}
\includegraphics[width=0.4\textwidth]{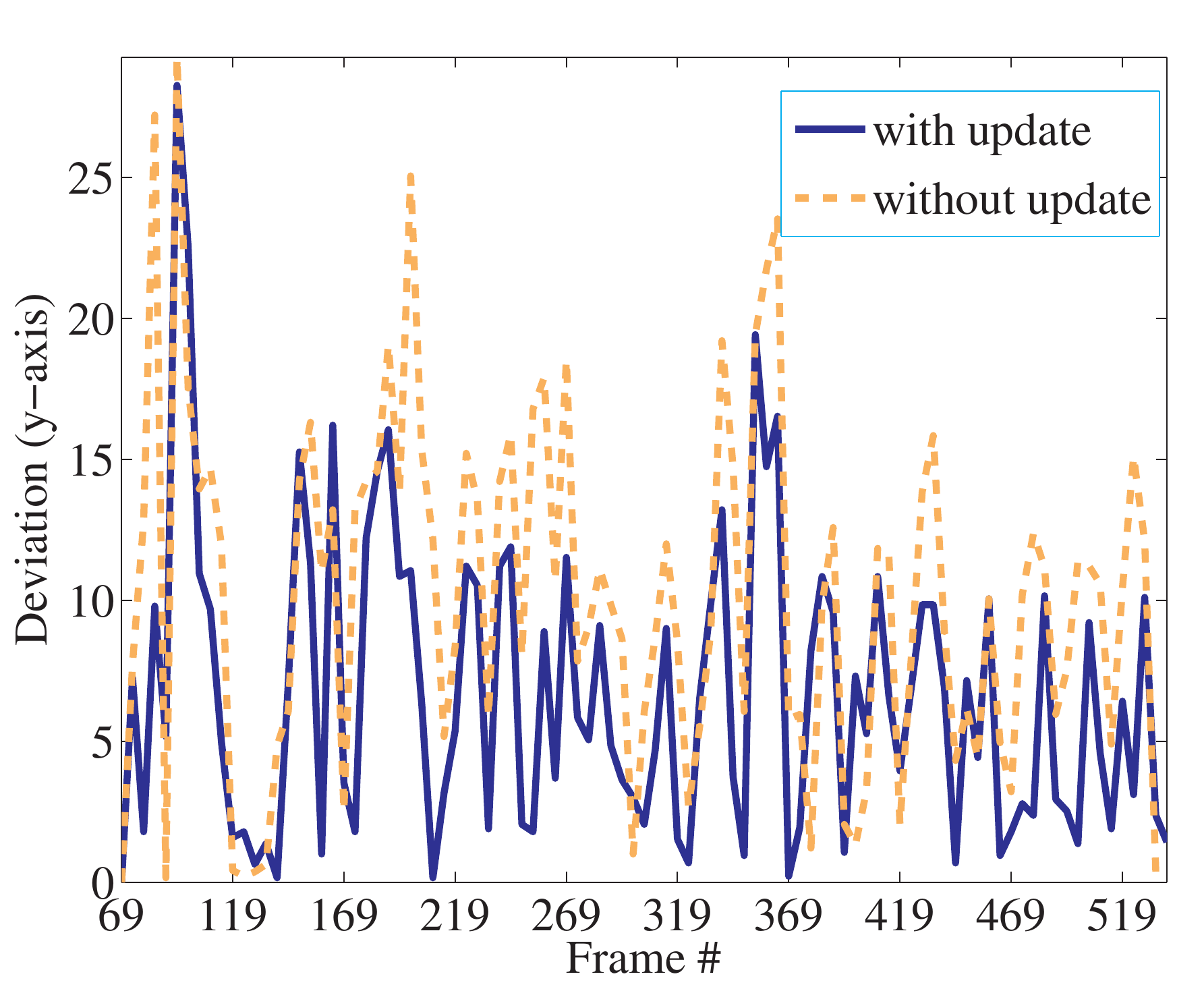}
  \caption{The $ \ell_1$-norm absolute error (pixels) 
  of the object's center against
  the ground truth on the {\tt walker} sequence 3. 
  The two figures correspond to $x$-, and $y$-axis, respectively. 
  It clearly shows that on-line update of the generalized 
  kernel tracker is beneficial: without on-line update, the error is
  larger. 
  }
  \label{fig:qant3}
\end{figure}

        To demonstrate the effectiveness of the on-line SVM learning,
        we switch off the on-line update and run the tracker on the
        {\tt walker} sequence 3. We plot the $ \ell_1$-norm absolute
        deviation of the tracked object's center in pixels at each
        frame in Fig.~\ref{fig:qant3}. Apparently, at most frames,
        on-line update produces more accurate tracking results.
        The average Euclidean tracking error is $ 8.0 \pm 4.9 $ pixels
        with on-line update and $ 12.7 \pm 5.8 $ pixels without
        on-line update.
        
        Conclusions that
        we can draw from these experiments are: (1) The
        proposed generalized kernel-based tracker performs better than
        the standard MS tracker on all the sequences that
        we have used; (2)
        On-line learning often improves tracking accuracy.


\section{Conclusion}
\label{sec:Conclusion}

	To summarize, we have proposed a novel approach to kernel
	based visual tracking, which performs better than 
	conventional single-view kernel trackers
	\cite{DVP2003KernelTrack,Elgammal2003Joint}.  
    Instead of minimizing the density distance between the candidate
    region and the template, the generalized MS tracker works by
    maximizing the SVM classification score. Experiments on
    localization and tracking show its efficiency and robustness.  In
    this way, we show the connection between standard MS tracking and
    SVM based tracking.  The proposed method provides a generalized
    framework to the previous methods.

    Future work will focus on the following possible avenues:
    \begin{itemize}
       \item                
          Other machine learning approaches such as relevance
          vector machines (RVM) \cite{Tipping01Sparse}, 
          might be employed to learn the
          representation model. Since in the test phrase, RVM and SVM  
          take the same form, RVM can be directly used here. 
          RVM achieves comparable recognition accuracy to
          the SVM, but requires substantially fewer kernel functions.
          %
          %
          %
          It would be interesting to compare different
          approaches' performances;

       \item
          The strategy in this paper can be easily plugged into a particle
          filter as an observation model. Improved tracking results are
          anticipated than for the simple color histogram particle filter
          tracker developed in \cite{P02Color}.

    \end{itemize}




\section*{Appendix}

Generally Collins' modified mean shift \cite{Collins03Blob}
(Equation \eqref{EQ:MSIteration}) 
cannot guarantee to converge to a
local maximum. It is obvious that a fixed point $ \x^* $ obtained by iteration
using Equation~\eqref{EQ:MSIteration} will not satisfy 
$$
      \grad f ( \x^* ) = 0. 
$$
$ f(\cdot) $ is the original cost function. Therefore, generally,
$ \x^* $ will not even be an extreme point of the original cost
function.  In the following example, $ \x^* $ obtained by Collins'
modified mean shift converges to a point which is close to a local
{\em minimum}, but not the exact minimum.

	\begin{figure}[h]
		\centering
			\includegraphics[width=0.4\textwidth]{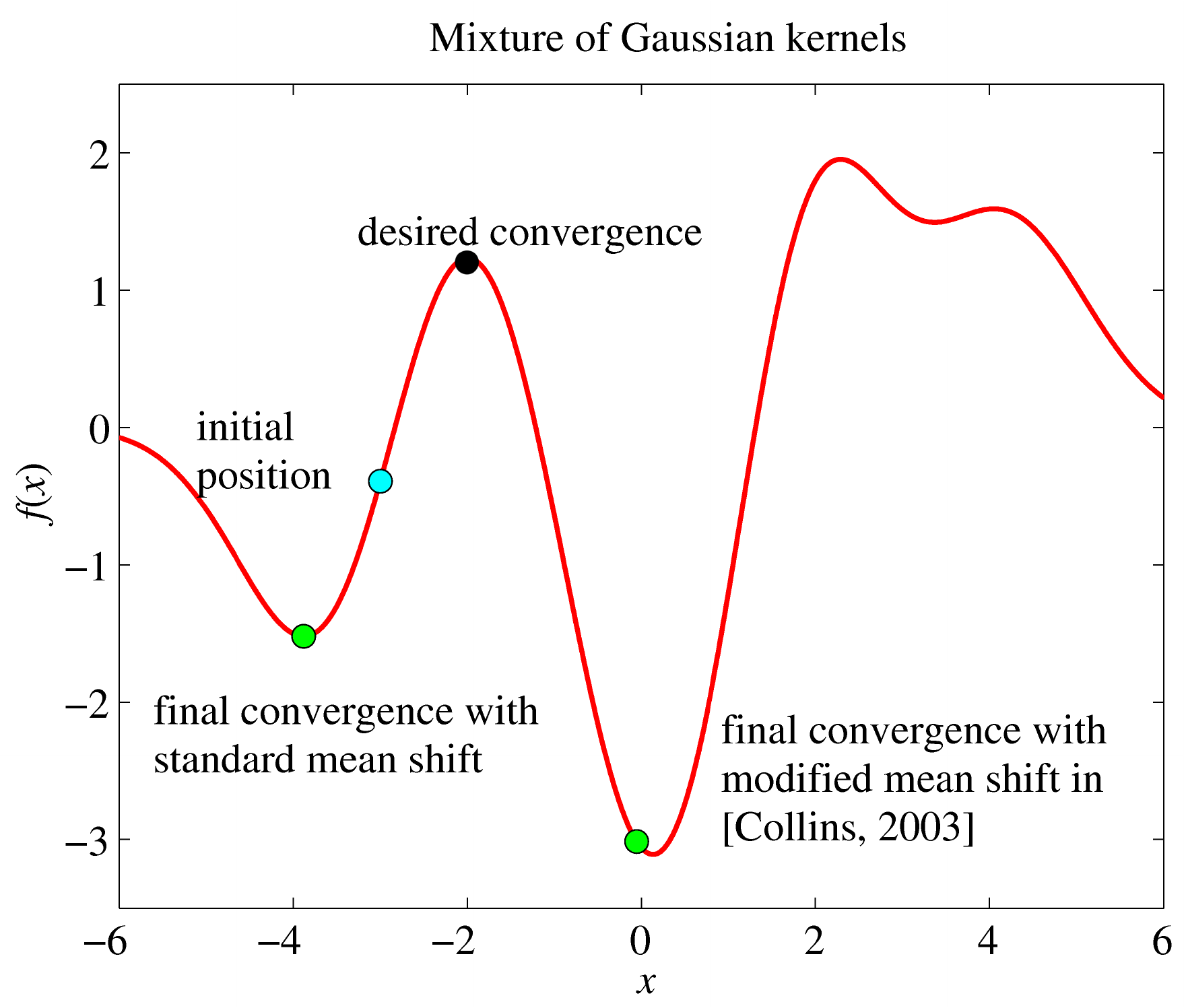}
            \caption{With negative weights, the modified mean shift proposed 
            in \cite{Collins03Blob} may not be able to converge to the local 
            maximum.  
            In this case, it converges to a position close to a local minimum         
            (not the exact minimum). The standard mean shift converges to the 
            nearest minimum.   
            }
		\label{fig:NegMSExp}
	\end{figure}

      In Fig.~\ref{fig:NegMSExp} we give an example 
      on a mixture of Gaussian kernel which contains 
      some negative weights. In this case  
      both the standard MS and Collins' 
      modified MS fail to converge to a maximum.

\footnotesize
\bibliographystyle{ieee}

\begin{IEEEbiography}
[{\includegraphics[width=1in,height=1.25in,clip,keepaspectratio]{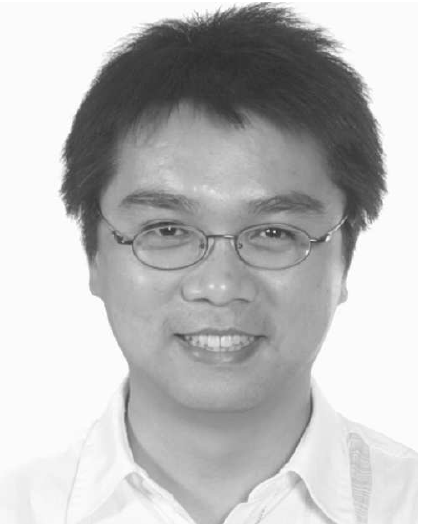}}]
{Chunhua Shen}
                   received the B.Sc. and M.Sc. degrees from Nanjing
                   University, China, and the Ph.D. degree from 
                   University of Adelaide,
                   Australia. 
                   He has been working as a research scientist 
                   in 
                   NICTA, Canberra Research Laboratory, Australia
                   since October 2005. 
                   He is also an adjunct research follow at Australian
                   National University and an adjunct lecturer at
                   University of Adelaide. 
                   His research interests include
                   statistical machine learning, convex optimization
                   and their application in computer vision.
\end{IEEEbiography}

\begin{IEEEbiography}
%
%
{Junae Kim} is a PhD student at the
        Research School of Information Sciences and Engineering, Australian National University.
        She is also attached to NICTA, Canberra Research Laboratory.
        She received the B.Sc. degree from Ewha Womans University, Korea in 2000,
        M.Sc. from Pohang University of Science and Technology, Korea in 2002, and
        M.Sc. from Australian National University in 2007.
        She was a researcher in Electronics and Telecommunications Research Institute (ETRI), Korea for 5 years
        before she moved to Australia.
        Her research interests include
        computer vision and machine learning.
\end{IEEEbiography}

\begin{IEEEbiography}
    [{\includegraphics[width=1in,height=1.25in,clip,keepaspectratio]{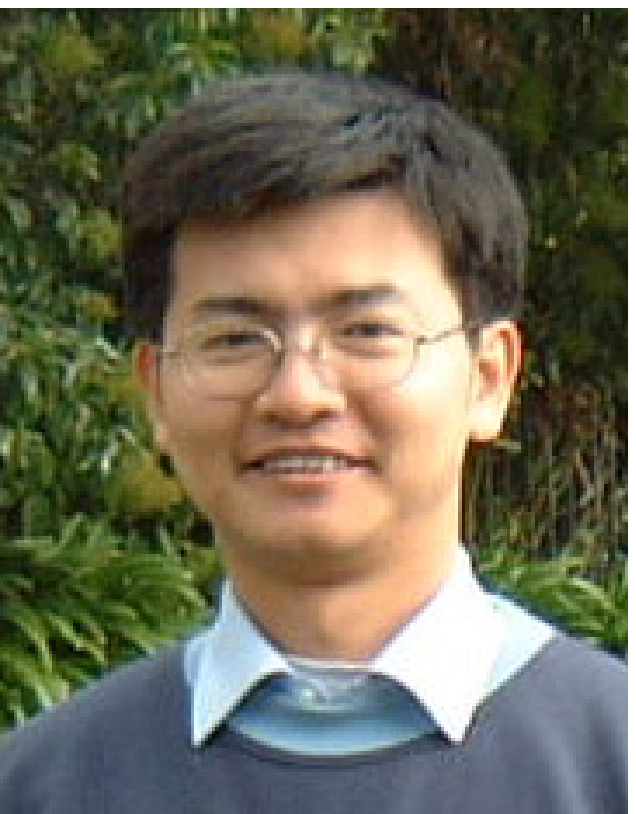}}]
    {Hanzi Wang}
                    received his B.Sc. degree in physics and M.Sc. degree in optics from Sichuan
                    University, China, in 1996 and 1999, respectively. He received his Ph.D. degree
                    in computer vision from Monash University, Australia, in 2004. He is a Senior
                    Research Fellow at the School of Computer Science, University of Adelaide,
                    Australia. His current research interest are mainly concentrated on computer
                    vision and pattern recognition including robust statistics, model fitting,
                    optical flow calculation, visual tracking, image segmentation, fundamental
                    matrix estimation and related fields. He has published more than 30 papers in
                    major international journals and conferences. He is a member of the IEEE
                    society.
\end{IEEEbiography}

\end{document}